\documentclass[10pt,twocolumn,letterpaper]{article}

\usepackage{cvpr}              % To produce the CAMERA-READY version
\usepackage[accsupp]{axessibility}

\definecolor{cvprblue}{rgb}{0.21,0.49,0.74}
\usepackage[pagebackref,breaklinks,colorlinks,allcolors=cvprblue]{hyperref}
\usepackage{multirow}
\usepackage{adjustbox}
\usepackage{colortbl}
\usepackage{mathtools}
\definecolor{cvprblue}{rgb}{0.21,0.49,0.74}
\usepackage[pagebackref,breaklinks,colorlinks,allcolors=cvprblue]{hyperref}

\def\paperID{15328} % *** Enter the Paper ID here
\def\confName{CVPR}
\def\confYear{2025}

\title{Comprehensive Information Bottleneck for Unveiling Universal Attribution to Interpret Vision Transformers}

\newcommand{\authorskip}{\hspace{7.5mm}}
\newcommand{\customfootnote}[1]{%
  \begingroup
  \renewcommand{\thefootnote}{}
  \footnote{#1}%
  \endgroup
}

\author{
Jung-Ho Hong \authorskip
Ho-Joong Kim \authorskip
Kyu-Sung Jeon
\authorskip
 Seong-Whan Lee$^*$ \\[2.5pt]
 \normalsize Dept. of Artificial Intelligence, Korea University, Seoul, Korea\\[-1pt]
 {\tt\small \{jungho-hong, hojoong\_kim, ksjeon, sw.lee\}@korea.ac.kr} \vspace{-1pt} \\
 {\hypersetup{urlcolor=blue}
\fontsize{9pt}{12pt}\selectfont \href{https://github.com/KU-HJH/CoIBA}{https://github.com/KU-HJH/CoIBA}}
}

\definecolor{LightCyan}{rgb}{0.88,1,1}
\colorlet{shadecolor}{gray!10}
\def \ours {CoIBA} 

\begin{document}

\maketitle
\begin{abstract}
The feature attribution method reveals the contribution of input variables to the decision-making process to provide an attribution map for explanation.
Existing methods grounded on the information bottleneck principle compute information in a specific layer to obtain attributions, compressing the features by injecting noise via a parametric damping ratio.
However, the attribution obtained in a specific layer neglects evidence of the decision-making process distributed across layers.
In this paper, we introduce a comprehensive information bottleneck (\ours{}), which discovers the relevant information in each targeted layer to explain the decision-making process.
Our core idea is applying information bottleneck in multiple targeted layers to estimate the comprehensive information by sharing a parametric damping ratio across the layers. 
Leveraging this shared ratio complements the over-compressed information to discover the omitted clues of the decision by sharing the relevant information across the targeted layers.
% We suggest the variational approach to fairly reflect the relevant information of each layer by upper bounding layer-wise information. %% Rebuttal 수정 전
We suggest the variational approach to fairly reflect the relevant information of each layer by upper bounding layer-wise information. %% Rebuttal 후 수정
Therefore, \ours{} guarantees that the discarded activation is unnecessary in every targeted layer to make a decision.
The extensive experimental results demonstrate the enhancement in faithfulness of the feature attributions provided by \ours{}.
\end{abstract}
\vspace{-10pt}
\customfootnote{*Corresponding author}
\section{Introduction}
\label{sec:intro}

Vision transformer (ViT) achieves remarkable performance in diverse fundamental computer vision tasks, such as multi-modal~\cite{clip,siglip} and self-supervised learning~\cite{dino,mae}.
Despite this achievement, the black box nature stemming from the complex structure of ViT restricts the testability of the end-to-end system, impeding the failure diagnosis~\cite{whyneed1}.
% is made from evidence desired by a user.
% This lack of transparency in ViT obscures the diagnosing of whether the decisions made by the model are derived from relevant image features, limiting its utilization in high-stake decisions.
% Why XAI need?
This limitation constrains the application of ViT in safety-critical areas where transparency and interpretability are considered important as well~\cite{whyneed2,whyneed3}.
% The limited transparency of ViT complicates verifying if decisions are grounded in relevant image features, thereby constraining its usage in high-stakes decision contexts.
% Feature attribution methods~\cite{generic,vit-cx,trans_attr,iia,beyond,iba} succeed in revealing the evidence of decisions made by the model.
Feature attribution methods~\cite{gradcam,integrated,lrp} are introduced to reveal the evidence supporting the decision-making process.
% Many of the existing attribution methods designed to interpret convolutional neural networks are not well-suited for ViT~\cite{trans_attr}.
Many of the existing attribution methods designed to interpret convolutional neural networks show limited explainability to interpret ViT~\cite{trans_attr}.
% Feature attribution methods~\cite{generic,vit-cx,trans_attr,iia,beyond,iba,inputiba}, specifically designed to interpret ViT, are introduced to address this issue by leveraging either attention weights computed in ViT or intermediate representations including the class token.
% Thus, feature attribution methods~\cite{generic,vit-cx,trans_attr,iia,beyond,iba,inputiba}, specifically designed to interpret ViT, are proposed to address this issue.
% These methods leverage either attention weights computed in ViT or intermediate representations including the class token to produce an attribution map.
To address this limitation, feature attribution methods~\cite{generic,vit-cx,trans_attr,iia,beyond} tailored for interpreting ViT have been proposed, leveraging either attention weights computed in ViT or intermediate representations, such as the class token, to generate attribution maps.
% explain the decision-making process, leveraging either attention weights computed in ViT or intermediate representations including the class token.
% These methods provide evidence with a visual explanation, reflecting the contribution of each input variable by leveraging the attention weights calculated in ViT.
% By calculating the attention weights of ViT, these methods leverage these weights to provide visual explanations that reflect the contribution of each input variable.
% These methods provide a visualization as an explanation, highlighting the contributions of each input variable.
% These methods explain the model decision by providing human-understandable visualizations represented as heatmap~\cite{}.

\begin{figure}[!t]
\centering
\subfloat[Qualitative visualizations and similarity comparison]{
\label{fig:intro_figure1_1}
\includegraphics[width=0.46\textwidth]{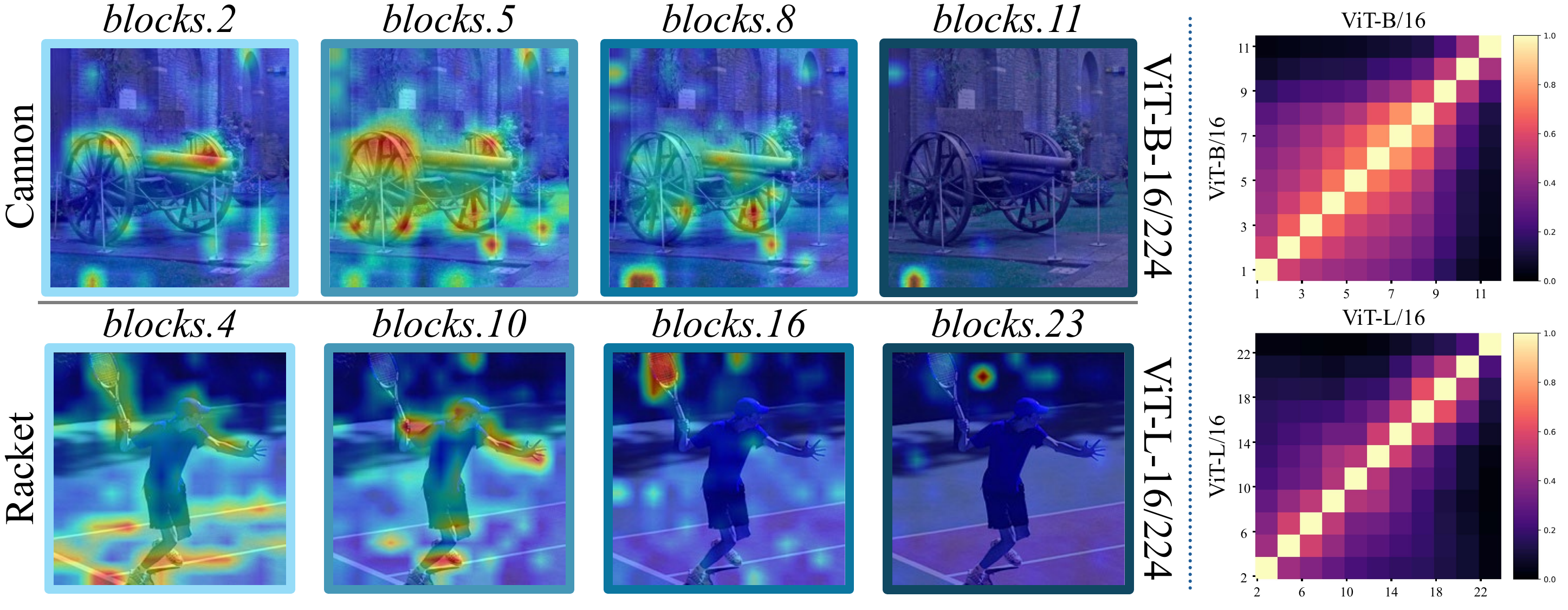}
\vspace{-2pt}
}

\subfloat[Cumulative number of each layer yielding optimal result per sample]{
\includegraphics[width=0.22\textwidth]{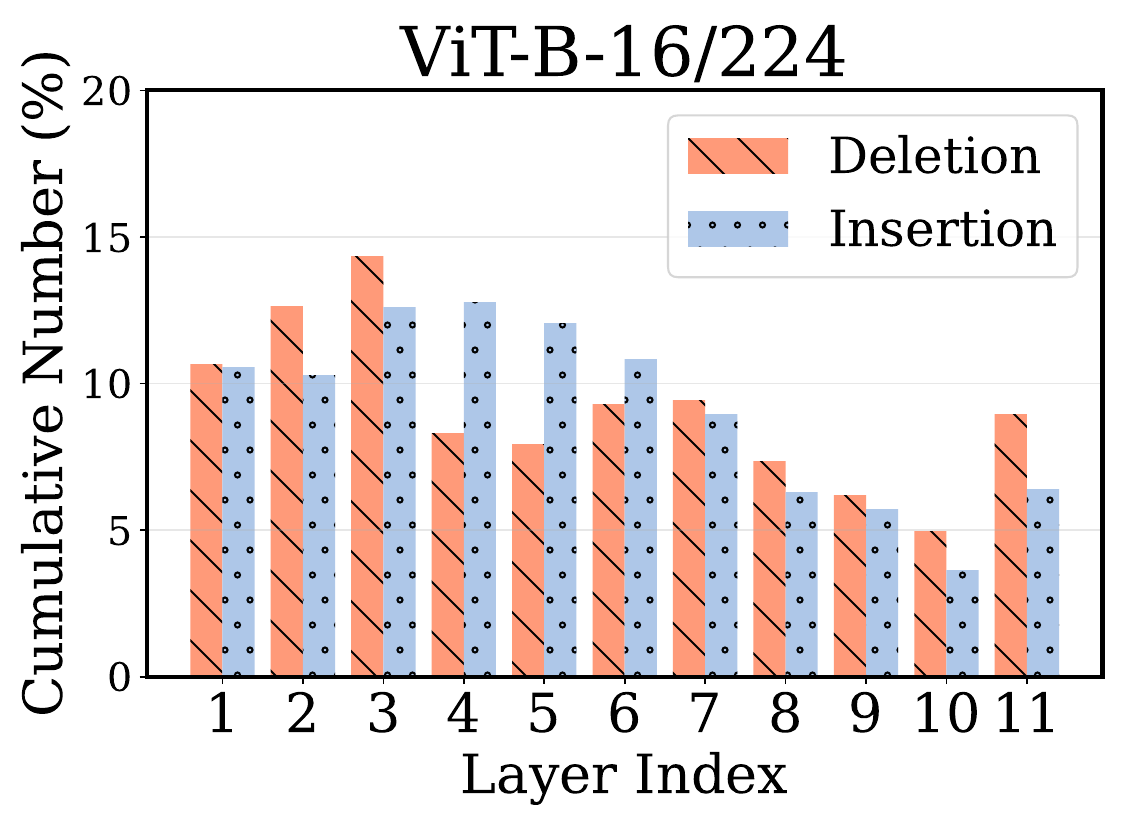}
\includegraphics[width=0.22\textwidth]{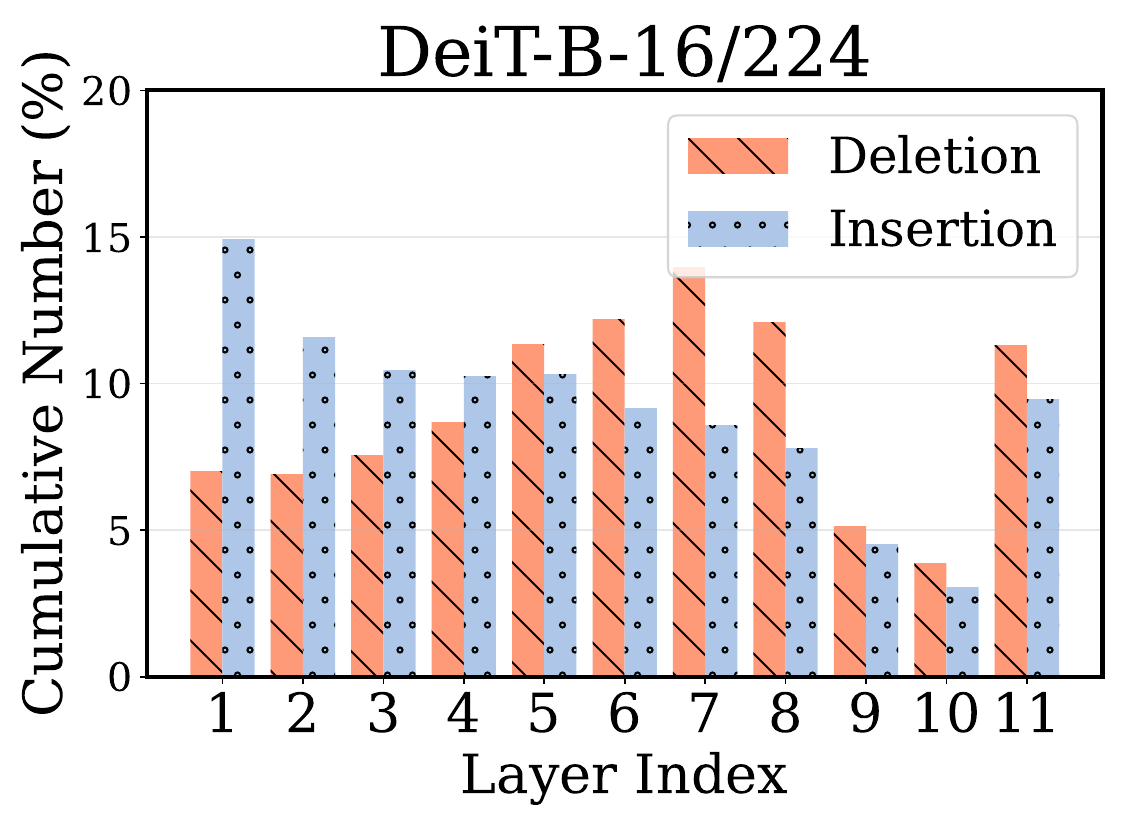}
\label{fig:intro_figure1_2}
\vspace{-2pt}
}
\vspace{-9pt}
\caption{\textbf{The quantitative and qualitative comparisons of attribution maps generated from different layers.}
    We illustrate the visualized attribution maps from the first to fourth columns in Fig.~\ref{fig:intro_figure1}\subref{fig:intro_figure1_1} utilizing IBA~\cite{iba}.
    The text on the top of each figure indicates the index of the layer.
    We select the layer index of the large model with interval 2.
    % We double the selected layer index of the results for the large model.
    The fifth column in Fig.~\ref{fig:intro_figure1}\subref{fig:intro_figure1_1} shows the similarity comparison between attributions of different layers.
    % The measurements of similarity between attributions are presented in the fifth column.
    Fig.~\ref{fig:intro_figure1}\subref{fig:intro_figure1_2} illustrates the cumulative number of each layer that provides the best explanation per sample.
    The uniformly distributed data indicates the absence of the layer optimal for the explanation.
    }

\vspace{-18pt}
\label{fig:intro_figure1}
\end{figure}

% \input{figs/intro_fig1}
% Existing feature attribution methods for ViT explanation cannot theoretically guarantee that low-scored attributions are unnecessary for the model decision~\cite{iba}.
Although feature attribution methods have advanced in interpreting ViT, they lack theoretical guarantees that low-scored attributions are not necessary to the decision-making process~\cite{iba}.
% Although existing feature attribution methods succeed in interpreting ViT utilizing propagation or gradient, they lack theoretical guarantees that low-scored attributions are irrelevant to the decision-making process~\cite{iba}.
The methods~\cite{iba,inputiba} grounded on the information bottleneck principle~\cite{ib_principle1,ib_principle2} address this limitation by guaranteeing the importance of attributions with a variational approximation. %are necessary for the prediction.
% IBA~\cite{iba} provides the feature attribution map by restricting the information flow of an intermediate layer, thereby obtaining the relevant information for the prediction.
% IBA~\cite{iba} leverages variational approximation to upper bound the estimation of information, alternating the unnecessary intermediate features with noise to remove information for subsequent layers.
% Information bottleneck for attribution (IBA)~\cite{iba} design the model-agnostic method to provide relevant information by leveraging variational approximation to upper bound the information.
Information bottleneck for attribution (IBA)~\cite{iba} proposes the model-agnostic method to compress the information unnecessary for the prediction by positioning the bottleneck layer into the target layer.
% estimation by inserting the bottleneck layer to the specifically targeted layer.
% By inserting the bottleneck layer into the targeted layer, IBA compresses the information unnecessary for the prediction.
% IBA reports the explainability of relevant information obtained with the information bottleneck principle and introduces the variational approximation approach to upper-bound information estimation.
% are not necessary for the decision-making process.
InputIBA~\cite{inputiba} succeeds in providing a high-resolution attribution map by directly measuring the information in the input domain, assuming the prior distribution with a generative model.
% restricting the information propagation in the input level.
% However, the existing IBA approach is designed to restrict the information to a specifically targeted layer, neglecting the role of different layers.
% ViT layers process the input samples with different attention distances~\cite{vit_vs_cnn}.
% Thus the relevant information of a specific layer to restrict information is incomplete to support the model decision as it reflects only partial information processed during the prediction.
% As a result, this incomplete attribution shows diminished faithfulness when explaining the difficult input samples.
% For the difficult datasets such as ImageNet-A~\cite{imagenet_a} and ImageNet-R~\cite{imagenet_r}, this limitation accelerates the diminish in faithfulness of the attribution map.
% However, the relevant information yielded by the aforementioned methods reflects the limited perspective on the decision-making process.
% However, existing methods grounded on the information bottleneck principle face two main challenges.
However, existing information bottleneck-based methods face two main challenges.
First, the resulting attribution map reflects the partial of the decision-making process as the restriction of the information flow is conducted in a specific targeted layer.
% the resulting attribution reflects the limited perspective on the decision-making process.
% This constraint results discrepancies in attribution maps obtained from different layers, emphasizing inconsistent evidence for the same prediction.
% Different layers capture different information.
This constraint provides inconsistent evidence for a single decision-making process, leading to unobvious interpretations.
% Lastly, none of the layers dominantly provides the most faithful attribution map of input variables to explain the decision-making process.
Lastly, no layer dominantly provides the most appropriate relevant information for explaining the decision-making process.
% To demonstrate the aforementioned issues, we provide quantitative and qualitative examples, illustrating this issue in Fig.~\ref{fig:intro_figure1}.

Fig.~\ref{fig:intro_figure1} shows the aforementioned issues with quantitative and qualitative examples.
In the first row in Fig.~\ref{fig:intro_figure1}\subref{fig:intro_figure1_1}, the visualization derived from the second layer (\textit{blocks.2}) highlights the \textit{barrel} whereas the fifth (\textit{blocks.5}) and eighth layers (\textit{blocks.8}) highlight the \textit{top} and \textit{bottom of the wheels}, respectively.
In the second row in Fig.~\ref{fig:intro_figure1}\subref{fig:intro_figure1_1}, the visualization in the earlier layer highlights the \textit{person holding the racket}, while the deeper layer progressively concentrates on the \textit{racket}.
To confirm the layer-wise discrepancy in attributions quantitatively, we measure SSIM~\cite{ssim} in the right row of Fig.~\ref{fig:intro_figure1}\subref{fig:intro_figure1_1}.
% As the layer providing attributions distance farther, the discrepancy in those attributions increases.
As the layer providing attributions distance farther, the dissimilarity in those attributions increases.
Thus, computing the relevant information from isolated layer-specific information bottlenecks highlights different attributions for the same decision.
% Identifying the layer consistently produces faithfulness attribution, which requires additional heuristics.
Among the layers producing the attribution maps, there is no dominant layer that provides the most appropriate relevant information to explain the decision-making process.
Fig.~\ref{fig:intro_figure1}\subref{fig:intro_figure1_2} compares the cumulative number of each layer yielding attributions with the optimal insertion and deletion scores per input sample.
% We illustrate this phenomenon in Fig.~\ref{fig:intro_figure1} (b).
% The results show that the layer producing the relevant information appropriate for explanation is not obvious.
The results show a lack of the layer that produces the most appropriate relevant information for explanation.
% Due to the lack of explicit criteria for distinguishing the optimal attribution, the existing method grounded on IBA requires a heuristic search to identify the most faithful attribution from the layers.
% Thus, identifying the most faithful attribution from the layers requires a heuristic search, due to the lack of criteria to distinguish the optimal layer. % 여기 확인.
Thus, identifying the most faithful attribution from the layers requires a heuristic search due to the lack of criteria to distinguish the optimal layer. % 여기 확인.

% In this paper, we introduce a comprehensive information bottleneck for attribution (\ours{}) to reflect information relevant to each bottleneck-inserted layer into the attribution map.

% In this paper, we propose a novel comprehensive information bottleneck for attribution (\ours{}), which provides an attribution map by reflecting information necessary in each targeted layer to make a decision, eliminating additional heuristics.
In this paper, we propose a novel comprehensive information bottleneck for attribution (\ours{}), which reveals the information relevant for each of the target layers to provide the attributions.
Our core idea is to reveal comprehensive relevant information by eliminating unnecessary information in every targeted layer of the prediction.
We estimate the comprehensive information by sharing a parameterized universal damping ratio across layers, thereby removing the requirements of additional heuristics.
This sharing strategy compensates for the over-compressed information of individual layers by sharing information in each layer necessary for the decision.
% Our core idea is to obtain a universal damping ratio that reveals comprehensive relevant information, eliminating the information unnecessary for the prediction in every layer.
% The universal damping ratio is shared across the targeted layers, restricting the information with the same ratio in each layer. %, guaranteeing the eliminated information is not necessary in the layers for the decision-making process.
% % This allows communication between the targeted layers by delivering relevant information between the layers via the parametric ratio, compensating for the overestimated information.
% This allows compensation for the overestimated information in individual layers by sharing information necessary in each layer for the decision-making process.
Since a channel element placed in the same location but different layers captures different features, we uniformly perturb all the channel elements in a single token representation, to handle the information of different layers with a shared parametric ratio.
% we design perturbation at the token level rather than at the channel level.
% In addition, we apply uniform channel perturbation, using an identical damping ratio across neurons that captures different concepts at each layer.
We suggest a variational upper bound to restrict information flowing in all the targeted layers.
Our approach eliminates the heuristic search for balancing inconsistent information along the layers by suggesting a variational upper bound to fairly reflect the layer-wise relevant information.
% Our variational upper bound induces the comprehensive relevant information to fairly reflect the layer-wise relevant information while eliminating the heuristic search for balancing inconsistent information along the layers.

We conduct experiments to show the correctness performance of \ours{} utilizing various assessments.
As the evaluations are solely approximations, \textit{i.e.} no ground truth exists, we include FunnyBirds~\cite{funnybirds}, which provide the ground truth and assess the comprehensive quality of attribution maps.
To assess the feature importance assessment, we leverage insertion/deletion~\cite{rise} and remove-and-debias (ROAD)~\cite{road}, which evaluate the faithfulness of attribution maps.
In addition, we scrutinize the faithfulness of \ours{} by analyzing the confident-aware assessment.
We discuss the effectiveness of our method with a rigorous discussion.

\section{Related Works}
\label{sec:related_works}
% \subsection{Fundamental Feature Attribution}
% To explain the decision-making process of the CNN, numerous fundamental feature attribution methods are introduced.
% GradCAM~\ref{gradcam} is one of the well-known approaches utilizing features from the internal representations.
% LayerCAM~\ref{layercam} enjoys abundant attributions by aggregating the features from the sequence of layers.
% RCAM~\ref{rcam} enhances the faithfulness of providing a significantly faithful attribution map from the internal layers by adhering to the layer-wise relevant propagation to obtain the visualization.
% LRP~\ref{lrp} introduces the fundamental approach of computing relevancy score to reveal the contribution of the input variables.

\noindent \textbf{Explanation Methods for Vision Transformer} 
% \subsection{Explanation Methods for Vision Transformer}
Existing explanatory methods designed to interpret ViT provide the attribution map utilizing the latent representation and the corresponding gradients.
% To shed light on the transparency of ViT, existing methods provide the attribution utilizing the latent representation and their gradients.
% Despite the success in interpreting the convolutional neural networks with gradient-based attribution maps~\cite{gradcam, integrated}, these methods show limited performance in explaining ViT.
However, these methods show limited explainability while being adapted to ViT~\cite{trans_attr}.
% direct utilization of the representation of intermediate layers is restrained to be adapted in the ViT~\cite{trans_attr}.
To address this limitation, Rollout~\cite{rollout} linearly combines the attention weights across the layers to compute the attribution map. 
Trans-attr~\cite{trans_attr} provides the class-discriminative attribution map by constructing the layer-wise relevance propagation rule for ViT.
Generic~\cite{generic} produces the attribution map with a generalized procedure to interpret diversified transformer models, leveraging the gradient of the attention map to produce the class-discriminative attribution maps.
IIA~\cite{iia} introduces the iterative integration across the input image, leveraging the internal representations processed by the model and their gradients.
ViT-CX~\cite{vit-cx} utilizes the patch embeddings and measures causal impacts to provide attributions.
Beyond~\cite{beyond} unfolds attention blocks with the chain rule between final prediction, CLS, and tokens to obtain token contribution.
Existing propagation-based approaches require a specific implementation to weigh the contributions of input variables.
In contrast to this constraint, \ours{} does not require implementing additional procedures as same as IBA.
Furthermore, aligning with the information bottleneck principle, \ours{} guarantees the highlighted attributions are important in all targeted layers for the decision.

\begin{figure}[!t]
\centering
\includegraphics[width=\linewidth]{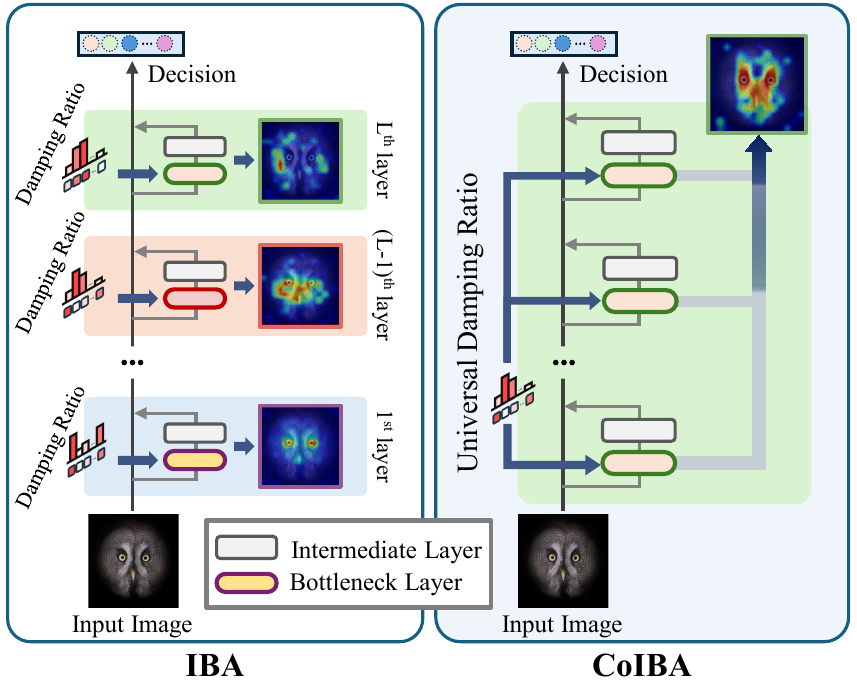}
\vspace{-18pt}
\caption{\textbf{Main difference between IBA and \ours{}.} IBA obtains the relevant information in the specific bottleneck-inserted layer. Thus, obtaining layer-wise attribution maps necessarily requires iterative running over the layers.
In contrast, \ours{} seeks the layer-wise relevant information from each bottleneck-inserted layer with a universal damping ratio.}
\vspace{-10pt}
\label{fig:intro_figure2}
\end{figure}
\noindent \textbf{Information Bottleneck Approach}
% Due to the theoretical solidity of the information bottleneck principle, various approaches adopt this principle.
The information bottleneck principle is broadly utilized to obtain information in the activation necessary for the specific objective.
To this end, the information bottleneck-based methods compress the information unnecessary for the objective while maintaining the relevant information.
Given the challenges of directly estimating the amount of information, a variational approach~\cite{vae} is used to approximate the intractable posteriors and marginals.
% The variational approximation approach~\cite{vae} introduces the way of restricting information with independent noise.
% Adding noise to the representation decreases the information~\cite{shannon}.
The Deep Variational Information Bottleneck~\cite{ib_principle2} method employs the information bottleneck principle in deep networks, using a variational lower bound alongside the reparameterization trick.
Referring to this strategy, IBA~\cite{iba} adapts the information bottleneck principles to restrict the information flow in an intermediate layer and provides the relevant information as an attribution map.
% utilizes the information bottleneck principle to provide attributions with the relevant information.
% IBA reports the relevant information is capable of revealing the contribution of the input image regions.
% reports that the relevancy revealed by the information bottleneck approach provides the explanation.
InputIBA~\cite{inputiba} provides high-resolution attributions by directly estimating relevant information in the input domain.
However, existing methods compute relevant information solely considering a specific layer, overlooking the information necessary in the different layers.
% In contrast to the aforementioned methods, \ours{} reflects comprehensive relevant information to the attribution, complementing the vanished relevant information in an isolated information restriction. 
% The enumerated methods are designed to estimate the information in a specific layer.
In contrast to this process, \ours{} obtains comprehensive relevant information from the multiple layers, highlighting the evidence important in every targeted layer to make a decision.

% \subsection{Layer-wise Information Bottleneck}

% \input{intro_fig2.t}

% \begin{itemize}

% \item LRP: Transformer attribution~\ref{trans_att} constructs the propagation rule to identify the contribution
% \item Generic: Generic transformer attribution~\ref{generic} 
% \item Unified: IIA~\ref{iia}
% \item \textbf{Perturbation-based}
% \item black-box: ViT-CX~\ref{vit-cx}
% \item Information Theory: IBA
% \end{itemize}

% \subsection{}

\section{Method}

% The main goal of \ours{} is a reflection of the relevancy of intermediate layers toward the resulting attribution map that correctly reveals the contributions of input variables.
The main goal of \ours{} is to produce an attribution map by reflecting the information relevant to the targeted layers, which is opposed to the IBA as shown in Fig.~\ref{fig:intro_figure2}.
To show this procedure, we first introduce IBA in Sec.~\ref{sec3-1} and discuss the limitations of the existing information bottleneck-based approach in Sec.~\ref{sec:motivation}.
Then, in Sec.~\ref{sec3-3}, we describe the overall procedure of \ours{}.
% generates an attribution map by considering the intermediate layers to amplify the faithfulness of the explanation.
Finally, we introduce a variational upper bound, which enhances the reflection of layer-wise relevant information in Sec.~\ref{sec3-4}.

\begin{figure}[t!]
\centering

% \hfill
% \subfloat[Feature importance measurement (ViT-B)]{
\includegraphics[width=.44\linewidth]{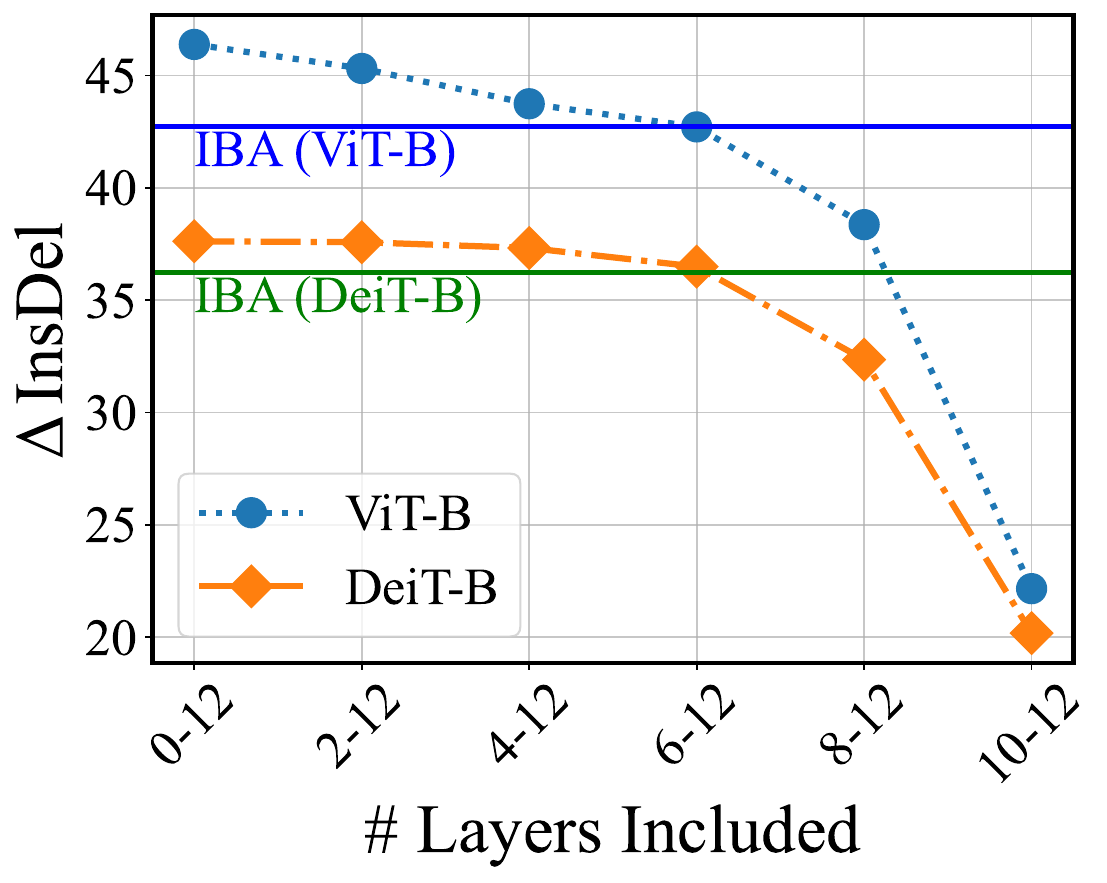}
\includegraphics[width=.45\linewidth]{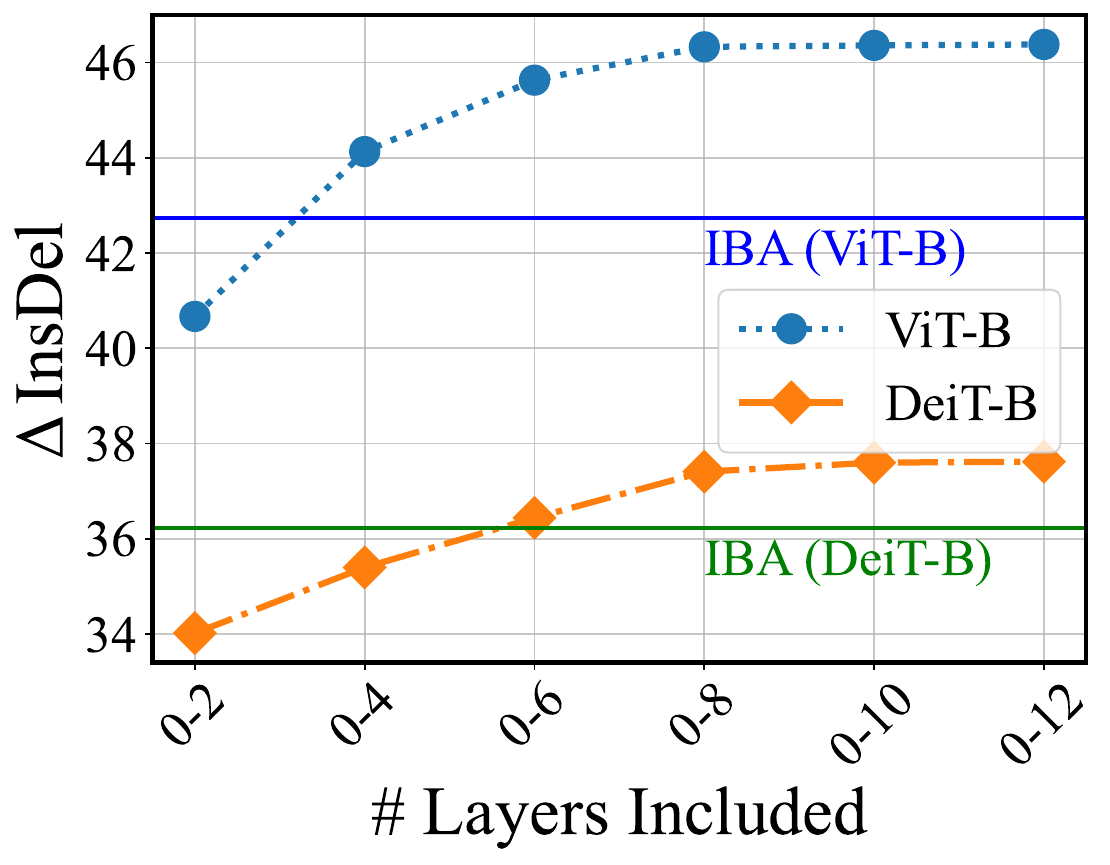} \\
\vspace{-2pt}
\includegraphics[width=.9\linewidth]{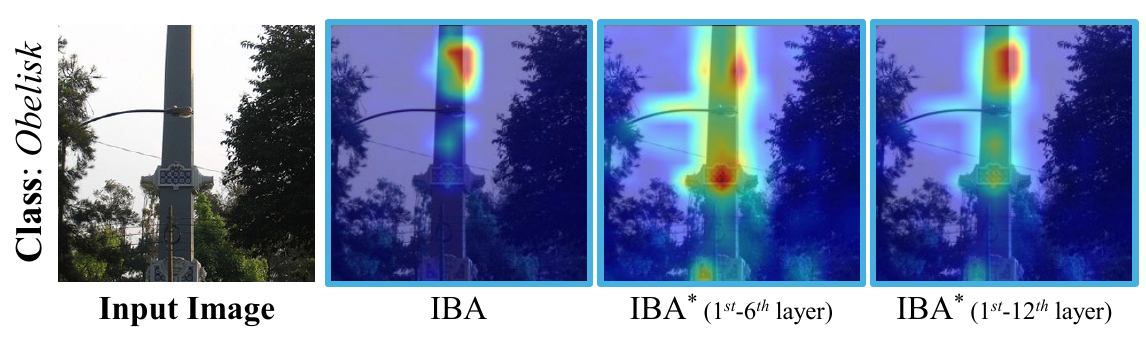}
% \includegraphics[width=.2\linewidth]{Assets/issue1_road_vit-b.pdf}
% \vspace{-2pt}
% }

\vspace{-11pt}
\caption{
\textbf{
Quantitative and qualitative results of IBA*.}
% We utilize IBA to obtain the set of attribution maps of each layer, iterating the method over the number of layers.
% IBA* indicates the results of linearly combined attribution maps obtained by iterating IBA over the layers.
In the first row, the horizontal line indicates the performance of layer-specific relevant information, \textit{i.e.}, IBA.
We report the difference in insertion and deletion scores ($\Delta$InsDel).
The higher score is better for this metric.
The visualized attribution maps are illustrated in the second row obtained from VIT-B-16/224~\cite{vit_orig}.
}
\label{fig:issue2}
\vspace{-15pt}

\end{figure}

\subsection{Background -- IBA}
\label{sec3-1}
IBA adopts an information bottleneck principle to produce relevant information to provide an attribution map.
% This relevant information is obtained from a specific bottleneck-inserted layer, restricting the information unnecessary for the model prediction within an intermediate representation.
% Since the direct estimation of mutual information is impractical, IBA optimizes variational approximation to upper-bound the mutual information.
% the formulated information bottleneck is optimized with variational approximation and minimizes the variational upper bound.
% The variational approximation ensures that the remained information after information restriction is relevant for the prediction.
% Here, minimizing the variational upper bound guarantees that the remained information is relevant for the target prediction.
To this end, in $l$-th layer, IBA computes the bottleneck representation $Z_l$ as follows:
\begin{equation}
    Z_l=\lambda_l R_l + (1 - \lambda_l)\epsilon_l \ \text{.}
\end{equation}
Here, the damping ratio $\lambda_l$ is a learnable parameter assigned at the bottleneck layer, manipulating the degree of damping signal between the activation $R_l$ and the independent noise $\epsilon_l$.
The independent noise $\epsilon_l$ is sampled from the Gaussian distribution.
To maintain the statistic of internal representation, the Gaussian distribution shares the mean $\mu_{R_l}$ and variance $\sigma_{R_l}$ with the representation $R_l$.

IBA minimizes the shared information between the activation $R_l$ and bottleneck variable $Z_l$ while maximizing the shared information between the bottleneck variable $Z_l$ and the label $Y$ as follows:
\begin{equation}
    \underset{\lambda{}_l}{\max} \ I[Z_l;Y]-\beta{}I[R_l, Z_l] \ \text{.}
\end{equation}
Here, $\beta{}$ is a hyperparameter manipulating the trade-off between compression and relevancy in $l$-th layer.
% The information restriction is conducted in the $l$th specific layer.
% IBA damps the signal by injecting the noise to diminish the mutual information.
% manipulate the shared information between the representation $X_l$ and the bottlenecked representation $Z_l$.

\subsection{Motivation}
\label{sec:motivation}
\begin{figure}[t!]
\centering

% \hfill
\includegraphics[width=.85\linewidth]{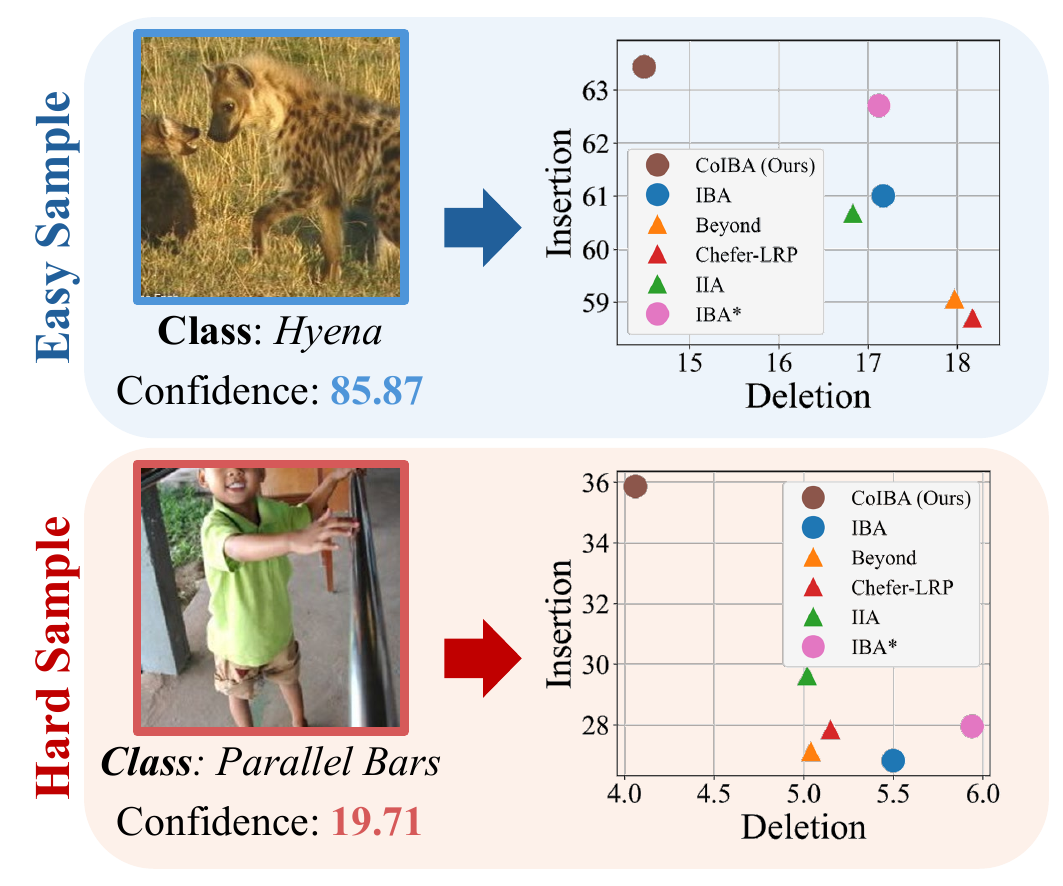}
\vspace{-10pt}
\caption{
\textbf{The quantitative comparisons of explanation methods in high-confident (easy) and low-confident (hard) input samples.}
The confidence score is the \textit{softmax} probability of the target class predicted by the model.
We compare the information-bottleneck-based methods including IBA~\cite{iba}, IBA$^*$, and \ours{} (marked with a circle), and the propagation-based explanation baselines including Beyond~\cite{beyond}, IIA~\cite{iia}, and Chefer-LRP~\cite{trans_attr} methods (marked with a triangle).
% The triangle and circle marks denote the results of propagation and IB-based approaches, respectively.
We utilize DeiT-B-16/224~\cite{deit} for the evaluation.
For insertion and deletion, higher and lower scores are better, respectively.
% Thus, achieving an upper-left direction is desirable for the methods.
}
\label{fig:issue1}
\vspace{-15pt}
\end{figure}

IBA selects a specific layer to obtain relevant information to provide an attribution map.
However, the relevant information of a specific single layer does not reflect the overall evidence required in the sequence of layers to make a decision.
To demonstrate our insistence, we illustrate the quantitative comparison of the attribution maps obtained in a specific layer and multiple layers in Fig.~\ref{fig:issue2}, leveraging insertion/deletion~\cite{rise}.
We iterate IBA multiple times along the layers to obtain the set of individual relevant information for each layer.
After that, we compute a linear combination over the set of individual relevant information.
We denote this procedure as IBA$^*$.
As shown in the results, the attribution maps produced by IBA$^*$ archive better scores compared to the IBA.
To scrutinize the quantitative results between IBA$^*$ and IBA, we split the insertion/deletion scores by the confidence scores of the model in Fig.~\ref{fig:issue1}.
%in terms of confidence scores computed by the model in Fig.~\ref{fig:issue1}.
% As shown in the results, the increase in scores computed by IBA$^*$ is only observed in easy samples.
As shown in the results, the increase in performance is only observed in high-confident samples, unlike in low-confident samples.
% , as the performance of low-confident samples is bounded.
Therefore, even iterating IBA, which requires high computational cost, shows limited performance enhancement according to the difficulty of each sample.

% In contrast to this observation, the performance of faithfulness in IBA$^*$ is bounded in low-confident samples.

\begin{figure*}[!th]
\centering
\includegraphics[width=0.92\textwidth]{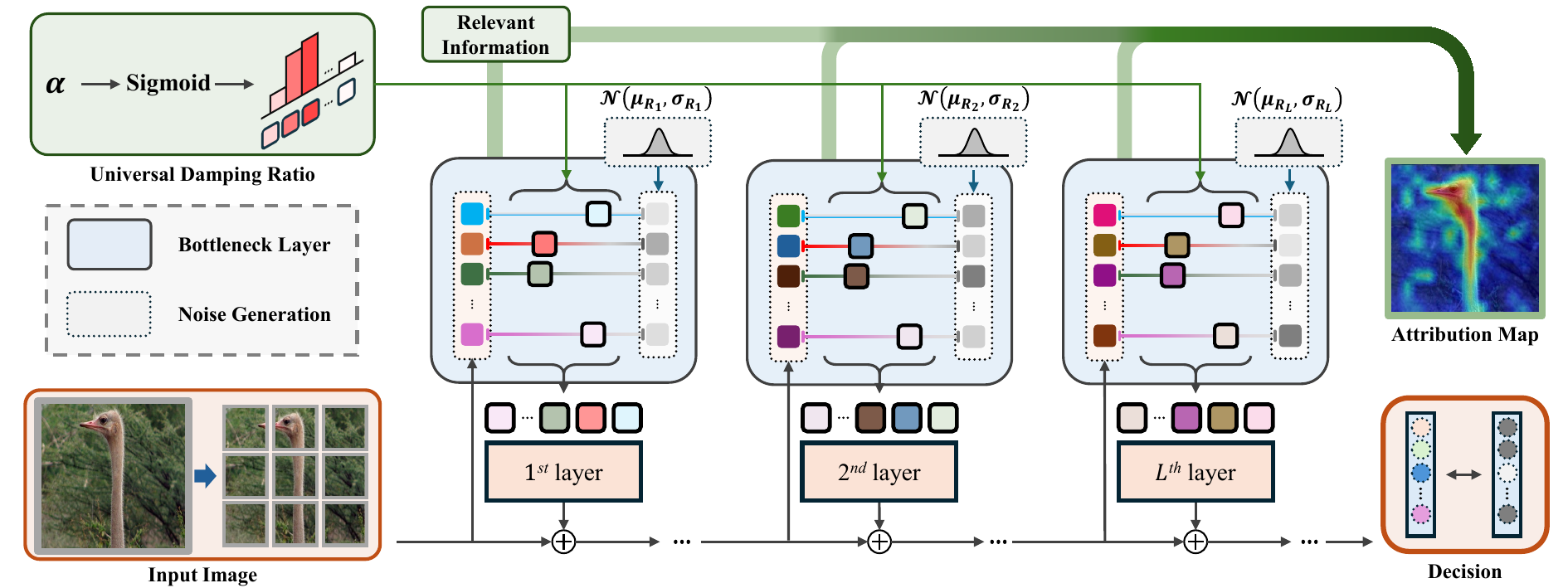}
\vspace{-10pt}
\caption{\textbf{An overview of \ours{}.} \ours{} restricts the information flow using a universal damping ratio while upper-bounding the information of each layer. After that, \ours{} aggregates the relevant information from each layer to produce an attribution map. 
% The optimization is conducted by minimizing the variational upper bound.
}
\vspace{-13pt}

\label{fig:main_figure}
\end{figure*}

\subsection{Comprehensive Information Bottleneck}
\label{sec3-3}
% \ours{} restricts the information unnecessary in entire bottleneck-inserted layers for the prediction to produce an attribution map.
% To this end, \ours{} inserts the bottleneck layer into the sequence of layers and restricts the information with the learnable universal damping ratio which is shared across the layers.
Our comprehensive information bottleneck (\ours{}) inserts the bottleneck layer into the sequence of targeted layers to restrict the information flowing in the intermediate layer.
Within the bottleneck layer, \ours{} restricts the information with the learnable universal damping ratio shared across the bottleneck-inserted layers.
Our goal is to optimize the universal damping ratio to obtain the layer-wise relevant information while fulfilling the information bottleneck objective.
% By leveraging a universal damping ratio, \ours{} restricts the layer-wise information flow while holding the information relevant to all bottleneck-inserted layers.
% Thereby, the damping ratio restricts the information of each bottleneck-inserted layer unnecessary for the prediction, discovering the omitted regions in layer-specific information restriction.
The universal damping ratio enables the attributions to identify distributed relevant information omitted by IBA, which restricts information in a layer-specific manner.

\noindent \textbf{Formulating Information Restriction} 
% Formally, in the circumstance of conducting the model prediction $\tilde{Y}=f(X)$, the output is computed by taking input $X$ to the model $f$.
The model is trained to make a decision $\tilde{Y}=f(X)$ to predict the ground truth $Y$ by taking input $X$.
To make a decision, the input sample $X$ is contextualized by passing through the sequence of intermediate layers.
Here, we insert the bottleneck layer into the targeted layer.
% $g_l=\{g_1,g_2,...,g_{L}\}$
% Among the intermediate layers, we select the bottleneck-inserted layers $f_l=\{f_1,f_2,...,f_{L}\}$ to restrict the information flow, where $L$ is the number of bottleneck-inserted layers.
The role of the bottleneck layer is to restrict the information by dampening the signal of the intermediate representation passing through the $L$ bottleneck-inserted layers.
We perturb the intermediate representations in the $l$-th layer to dampen the signal.
To dampen the signal, we inject the independent noise $\epsilon_l$ into the intermediate bottleneck representation $R'_l$, which is computed from the bottleneck variables $Z_{l-1}$ of the preceding layer.
Note that the first element of intermediate bottleneck representations is the non-perturbed intermediate representation, \textit{i.e.}, $R'_1=R_1$.
Thus, we obtain the bottleneck representation $Z_l$ as follows:
% Here, as the bottleneck layer damps the signal of the intermediate features, we define the bottleneck representation $Z_l$ at the $l$-th layer as follows:
\begin{equation}
    % Z_l= \lambda{}f_{l-1:l}(Z_{l-1}) + (1 - \lambda)\epsilon{}_l \ \text{,}
    Z_l= \lambda{}R'_l + (1 - \lambda)\epsilon{}_l \ \text{,}
\end{equation}
% where $\lambda \in \mathbb{R}^{P\times 1}$ and $\epsilon_l  \in \mathbb{R}^{P\times D}$ are a universal damping ratio and the independent noise, respectively.
Here, a universal damping ratio $\lambda \in \mathbb{R}^{P\times 1}$ manipulates the degree of perturbation for each token, denoting $P$ as the number of patches.
Thus, $\lambda$ is ranged from 0 to 1, allowing the propagation of the signal when $\lambda=1$ otherwise blocking it when $\lambda=0$.
As the universal damping ratio $\lambda$ is consistently adapted to the layers, we omit the layer index in the notation.
We compute $\lambda = \text{sigmoid}(\alpha)$ by passing the trainable parameter $\alpha$, which is initialized with 5, to the sigmoid function.
% We denote the number of patches and the size of the channel dimension as $P$ and $D$, respectively.
% In particular, since the specific element in the channel is not ensured to play the same role as the corresponding element of different layers, we singularize the learnable coefficient to enhance the independence against the token representations.
% As the universal damping ratio $\lambda$ is repeatedly adapted to the layers, it manipulates the information propagation with $\lambda=1$ whereas $\lambda=0$ blocks it.
% Note that we set the universal damping ratio to uniformly impute channel dimension to enhance the independence against the token representations.
The universal damping ratio uniformly perturbs the signal along the channel dimension.
% As our universal damping ratio manipulates the token, 
This setting avoids handling the neurons that capture different features with the same coefficient and induces concentration on the token's importance.
We empirically show this uniform perturbation across the channels enhances the faithfulness of resulting relevant information.
% This addresses the issue that a specific element in the channel is not ensured to play the same role as the corresponding element of different layers.
% As the damping ratio $\lambda$ shares the dimension with the intermediate representation except for the channel, according to the channel dimension, the noise is uniformly injected.
% The independent noise $\epsilon_l$ shares the dimension with the representation and is injected into the representation $f_{l-1:l}(Z_l)$ by linear combination.
The independent noise $\epsilon_l$ shares the dimension and is sampled from the Gaussian distribution with the mean $\mu_{R_l}$ and variance $\sigma_{R_l}^2$, such that $\epsilon_l \sim \mathcal{N}(\mu_{R_l}, \sigma_{R_l}^2)$.
To minimize statistical differences between the bottleneck variable and non-perturbed activations, we leverage the mean $\mu_{R_l}$ and variance $\sigma_{R_l}^2$ from those of the non-perturbed activations.
% Denoting $f_{l-1:l}$ as the sequence of operations positioned between two consecutive bottleneck layers, 
% The intermediate representation $R'_l$ is computed by forwarding the bottleneck variables toward the operations.
% Thus the first element of the sequence of intermediate representations is the original intermediate representation, \textit{i.e.} $R'_1=R_1$.
% The intermediate representation $R'_l$ is features computed by encoding the bottleneck variable of previous layers.
% The first element in the sequence of bottleneck variables is a non-perturbed intermediate representation, \textit{i.e.} $Z_0 = R_0$.
% The operations positioned between bottleneck layers are denoted as $f_{l-1:l}$.
% To maintain the statistics of the internal representations, we synchronize the mean and variance of the noise with the intermediate representation, i.e. $\epsilon_l \sim \mathcal{N}(\mu_{R_l}, \sigma_{R_l}^2)$.

\noindent \textbf{Information Bottleneck Objective} 
% Based on the formulation of the above bottleneck representation, we can increase or decrease the mutual information between bottleneck and intermediate representations.
The information bottleneck principle desires to obtain information necessary for the objective while compressing irrelevant information from the signal.
Compressing irrelevant information is performed by minimizing the shared information between the activation of a bottleneck-inserted network and the bottleneck representation.
Building upon this, we compress the amount of information by minimizing the mutual information between internal sequences of bottleneck representations while maximizing the mutual information between the ground truth and bottleneck representations.
% Combining two aspects, with the ground truth $Y$, we maximize the mutual information between the bottleneck variable and the ground truth $I[Z_{L}; Y]$ while minimizing the mutual information between bottleneck representations $I[Z_{l-1}; Z_{l}]$ as follows:
Combining two aspects with the ground truth $Y$, we maximize the mutual information between the bottleneck variable and the ground truth while minimizing the mutual information between bottleneck representations as follows:
\begin{equation} \label{eq:main1}
    \underset{\lambda}{\text{max}} \ I[Z_{L}; Y] - \frac{1}{L} \left( \sum{}_{l=1}^{L} \ \beta{}_l I[Z_{l-1}; Z_{l}] \right) \ \text{.}
\end{equation}
Here, the trade-off parameter $\beta_l$ governs the degree of compression in each layer.
Since the bottleneck layers are inserted into the sequence of layers, the compression term includes the accumulated sequence of mutual information of bottleneck variables $I[Z_{l-1}; Z_{l}]$.
% As \ours{} restricts the mutual information of all the bottleneck-inserted layers, we accumulate the sequence of mutual information of internal representations as $\sum_{l} I[Z_{l-1}; Z_{l}]$.
Here, the $Z_0$ indicates the original intermediate representation as the first element, such that $Z_0 = R_1$.
% The intermediate representation $R_1$ is not perturbed in the first bottleneck-inserted layer.
We maximize the relevancy term $I[Z_{L}; Y]$ while compression to maintain the shared information between bottleneck variables and the ground truth.
We maximize this term by minimizing the cross-entropy loss $\mathcal{L}_{ce}$.
% To maintain the predictive information in the internal representations, we maximize the $I[Y; Z_{<L}]$.
% Here, we minimize the cross-entropy loss $L_{ce}$ to maximize the predictive information.

Utilizing KL divergence, the mutual information between internal representations $I[Z_{l-1}; Z_l]$ is formulated by:
\begin{equation}
    I[Z_{l-1}; Z_{l}] = \mathbb{E}_{Z_{l-1}}[D_{KL}[P(Z_l|Z_{l-1})||P(Z_l)]] 
\end{equation}
However, the direct estimation of the prior distribution $P(Z_l)$ is intractable as it requires the integration of bottleneck representations $P(Z_l)=\int{}_{Z_{l-1}} P(Z_l | Z_{l-1})  P(Z_{l-1}) d Z_{l-1}$ of the corresponding $l$-th layer.
Thus, we assume the prior distribution $Q(Z_l)$ as Gaussian distribution $\mathcal{N}(\mu{}_{R_l}, \sigma{}_{R_l}^2)$, reconstituting the mutual information computation as follows:
\begin{equation} \label{eq:upper_bound1}
    I[Z_l; Z_{l-1}] 
\leq \mathbb{E}_{Z_{l-1}}[D_{KL}[P(Z_l|Z_{l-1})||Q(Z_l)]] \text{.}
\end{equation}
% As we obtain the over-estimated mutual information by assuming the prior distribution, minimizing the term compresses the mutual information.
Since assuming the prior distribution with Gaussian distribution only overestimates the mutual information, minimizing this term suppresses the mutual information of the compression term.

\begin{figure*}[!ht]
\centering
\includegraphics[width=0.98\linewidth]{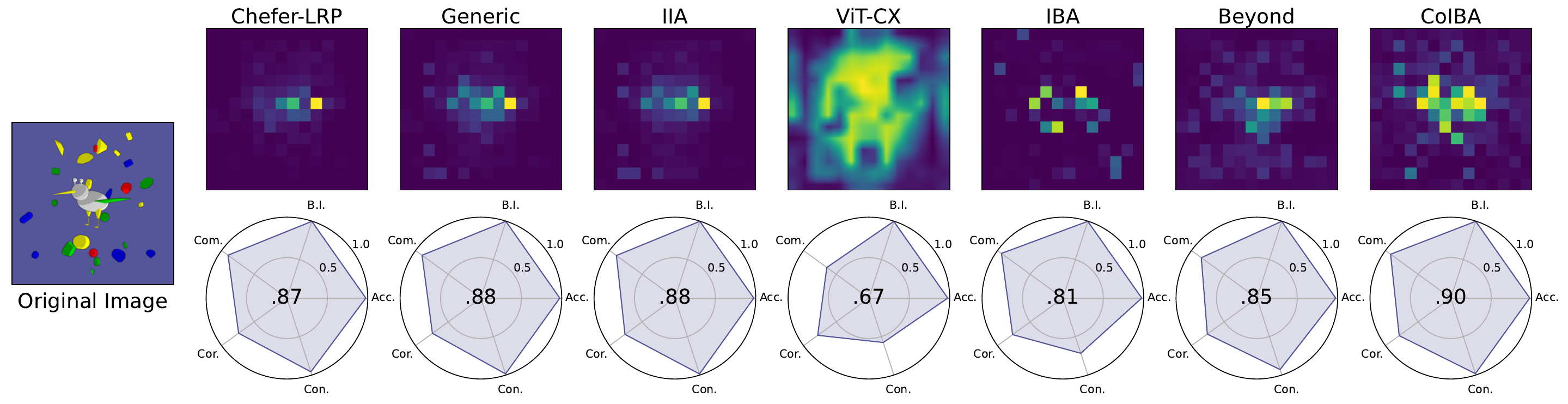}
\vspace{-6pt}
\caption{\textbf{Quantitative comprehensive assessment on FunnyBirds experiment:} The abbreviations Com. Cor. and Con. indicate \textit{Completeness}, \textit{correctness}, and \textit{contrastivity}. The center score indicates the comprehensive result of enumerated aspects. Acc. and B.I. indicate accuracy and background independence, respectively. We provide detailed numeric in the supplementary material.}
% The center score denotes the mean of the completeness (Com.), correctness (Cor.), and contrastivity (Con.) dimensions. Additionally, we report the accuracy (Acc.) and background independence (B.I.)
\vspace{-8pt}
\label{fig:funnybirds}
\end{figure*}

\begin{table*}[!ht]
\centering
\begin{centering}

\resizebox{0.9\linewidth}{!}{
\begin{tabular}{lcccccccc}
\specialrule{2.0pt}{1pt}{1pt}
Variant                   & Model                                                        & Chefer-LRP~\cite{trans_attr} & Generic~\cite{generic} & IIA~\cite{iia} & ViT-CX~\cite{vit-cx} & IBA~\cite{iba} & Beyond~\cite{beyond} & \ours{} \\ \hline
                        % & ViT-B-8/224$^*$ & - & 19.57/49.86 & 18.43/51.32 & 20.97/46.75 & 16.68/51.91 & 17.10/51.82 & \textbf{13.44}/\textbf{54.18} \\
                        % & ViT-B-32/224$^*$ & - & 17.54/55.25 & 16.11/56.35 & 16.04/58.09 & 15.51/58.84 & 15.15/58.51 & \textbf{13.73}/\textbf{62.02} \\ 
\multirow{4}{*}{ViT}    & ViT$^*$-S-16/224 & - & 14.56/56.08 & 13.69/57.31 & 14.98/56.65 & 13.56/\underline{58.64} & \underline{13.50}/58.47 & \textbf{11.08}/\textbf{63.35} \\ 
                        & ViT$^*$-T-16/224 & - & 9.00/48.95 & 8.50/\underline{49.75} & 12.08/45.09 & \underline{8.02}/49.22 & 11.26/44.07 & \textbf{6.97}/\textbf{54.23} \\
                        & ViT-B-16/224 & 17.67/56.69 & 17.92/57.03 & 16.56/58.7 & 16.94/58.17 & 17.23/59.06 & \underline{15.84}/\underline{60.37} & \textbf{13.01}/\textbf{62.58} \\
                        & ViT-L-16/224 & 20.97/53.81 & 20.17/55.54 & 20.14/55.22 & 19.76/\underline{57.38} & 20.88/54.85 & \underline{18.99}/56.55 & \textbf{15.40}/\textbf{61.14} \\  \hline
% \multirow{5}{*}{DeiT}    & DeiT-T-16/224 & 7.70/35.67 & 7.36/36.88 & 6.62/37.72 &                               11.45/32.20 & 6.80/36.44 & 7.13/37.22 &                                                   \textbf{6.02}/\textbf{41.54} \\
\multirow{4}{*}{DeiT}    & DeiT-S-16/224 & 12.76/47.51 & 12.10/48.64 & 10.89/\underline{49.76} & 
                         17.57/44.43 & 11.43/47.87 & \underline{11.13}/48.88 & \textbf{9.55}/\textbf{53.38} \\ 
                         & DeiT-B-16/224 & 14.63/48.79 & 15.10/49.41 & 13.69/\underline{50.51} & 18.21/46.97 & \underline{13.95}/50.18 & 14.45/48.88 & \textbf{11.79}/\textbf{53.96} \\ 
                         & DeiT3-B-16/224 & - & 15.70/52.42 & 15.38/52.70 & 18.88/50.40 & 15.07/53.32 & \underline{14.71}/\underline{54.07} & \textbf{12.97}/\textbf{56.54} \\ 
                         & DeiT3-L-16/224 & - & 22.10/62.73 & 17.75/55.93 & 26.11/59.60 & 21.01/\underline{65.60} & \underline{19.34}/64.27 & \textbf{17.64}/\textbf{67.20} \\  \hline
% \multirow{3}{*}{IN-A}    & Dino-B-16/224  & - & 7.62/50.42 & 7.46/50.52 & 14.09/45.55 & 
%                          8.91/49.45 & 8.12/50.23 & \textbf{6.83}/\textbf{53.68} \\ 
%                          & MAE-B-16/224 & - & 24.44/42.8 & 24.53/43.63 & 20.01/45.80 & 15.66/48.90 & 16.23/48.09 & \textbf{13.77}/\textbf{53.42} \\  
%                          & BeiT-B-16/224 & - & 24.82/47.04 & 25.34/47.50 & 21.58/54.06 & 18.80/53.68 & 19.75/51.82 & \textbf{13.51}/\textbf{62.45} \\ \hline
% \multirow{2}{*}{IN-A}    & ViT-B-16/224 & - & 1.82/18.36 & 16.56/58.7 & 16.94/58.17 & 17.23/59.06 & 15.84/60.37 & \textbf{13.01}/\textbf{62.58} \\
%                         & ViT-B-16/224 & 17.67/56.69 & 17.92/57.03 & 16.56/58.7 & 16.94/58.17 & 17.23/59.06 & 15.84/60.37 & \textbf{13.01}/\textbf{62.58} \\
\multirow{2}{*}{Swin} & Swin-B & - & 33.76/42.94 & - & - & \underline{18.03}/\underline{52.81} & 23.73/50.39  & \textbf{17.09}/\textbf{54.80} \\ 
                         & Swin2-B & - & 34.71/47.16 & - & - & \underline{20.94}/\underline{53.85} & 31.32/49.44 & \textbf{18.88}/\textbf{55.77} \\

\specialrule{2.0pt}{1pt}{1pt}
\end{tabular}
}
\vspace{-8pt}
\caption{\textbf{Quantitative feature importance assessment on insertion $\uparrow$ / deletion $\downarrow$.} $(^*)$ denotes the ViT trained with strong regularization techniques~\cite{vit}. We underline the state-of-the-art performance among the baselines.}
\vspace{-15pt}
% \end{adjustbox}
\label{tab:insdel}
\end{centering}
\end{table*}

\subsection{Variational Upper Bound}
\label{sec3-4}
As discussed in Sec.~\ref{sec:discussion_upperbound}, the relevant information obtained by suppressing layer-wise mutual information suffers from reflecting layer-wise relevant information due to the inconsistent amount of information along the layers.  %% Rebuttal 수정 후
Thus, obtaining the sequence of trade-off parameters $\{\beta_l\}_{l=1}^L$ requires an extensive heuristic search to balance the degree of compression for each layer.  %% Rebuttal 수정 전
% As our objective function in Eq.~\eqref{eq:main1} requires iteration over the layers, the computational overhead significantly increases as the depth of the network increases. %% Rebuttal 후 수정
To overcome this issue, we suggest the variational upper bound for \ours{}, referring to the two following points.
First, since there is no additional information propagated during the forwarding pass, the mutual information of intermediate layers would not be greater than the mutual information between the model input and output.
Second, the iterated noise-injection procedure diminishes the mutual information among the internal and bottleneck representations, \textit{i.e.}, $I[Z_l; Z_{l+1}] \leq I[Z_{l-1}; Z_{l}]$.
Building upon these bases, we establish an upper bound that encompasses the combined mutual information from all subsequent layers as follows:
\begin{equation}\label{eq:upper_bound2}
    \begin{split}
    I[R_1; Z_1] \geq \frac{1}{L} \left( \sum{}_{l=1}^{L} \ I[Z_{l-1}; Z_{l}] \right) \ \text{.}  
    \end{split}
\end{equation}
Accordingly, we reconstitute the objective to be a simplified formula, leaving only a single hyperparameter $\beta$ as follows:
\begin{equation}\label{eq:main2}
\begin{split}
    \underset{\lambda}{\text{max}} \ I[Z_L; Y] - \beta{}I[R_1; Z_1] \ \text{.}
\end{split}
\end{equation}
To compress the subsequent layers, the simplified objective necessitates calculating only the mutual information of the first layer $I[R_1; Z_1]$.
This term solely utilizes the non-perturbed intermediate and bottleneck representations of the first layer, simplifying the objective calculation.
In terms of relevant information computation, referring to the data processing inequality, the inequality $I[Y; Z_l] \leq I[Y; Z_{l-1}]$ holds.
Thereby, in contrast to the layer-specific information bottleneck, which overestimates mutual information in the earlier layers during compression~\cite{inputiba}, \ours{} relieves the overestimation.
This is because, as discussed in Sec.~\ref{sec:discussion_universal}, the relevant information is compensated as the sequence of bottleneck variables joins the objective computation.
% To maximize the shared information against the label $Y$, we minimize the cross entropy loss $\mathcal{L}_{ce}$.
% The derivative of this approach is provided in the Appendix.

\section{Experiments}

% \subsection{Experimental Setup}
% \subsubsection{Architectures}
% We measure the quality of attribution maps in two main perspectives: causality, and localization.
% To assess the causality, we utilize Funnybird~\cite{funnybirds}, insertion/deletion~\cite{rise}, and remove-and-debias (ROAD)~\cite{road} assessments.
% We thoroughly discuss the validity of our approach via confidential analysis with causality assessments.
% In addition to the causality, we provide sensitivity-N and sanity check assessments in the supplementary material.
\subsection{Setup}

% \paragraph{Architectures}
\noindent \textbf{Architectures}
We select variants of ViT for our experiment, including original ViT~\cite{vit_orig} and DeiT models~\cite{deit3,deit}, trained with ImageNet-21k (IN-21k) and ImageNet-1k (IN-1k), respectively.
We include the variants in depth interpolated from ViT-T, ViT-S, ViT-B, ViT-L, and ViT-H models.
For the ViT-T and ViT-S models, we utilize the model pre-trained with massive regularization techniques~\cite{vit}.
The ViT-H model is pre-trained with CLIP~\cite{clip}.
We present the settings of the ViT-B model with patch size 16 and image resolution 224 as ViT-B-16/224.
% We utilize a fine-tuned model pre-trained with CLIP to include the ViT-H model.
% Furthermore, We include (8, 16, 32) as variants of patch size and (224, 384) as input image resolutions.
% Especially, the family of ViT architectures is trained with ImageNet-22k~\cite{imagenet} and fine-tuned with ImageNet-1k.
% Different from this setting, we select DeiT and DeiT3 architectures trained with only ImageNet-1k to demonstrate the generability of \ours{} against the capacity of the training dataset.
% As for the variants of training strategy, the self-supervised learned models, we include BeiT~\cite{beit}, Dino~\cite{dino}, and MAE~\cite{mae}.
We include Swin transformers~\cite{swin1,swin2}, demonstrating the generalized ability of \ours{} in multi-scale features processing.
The details of settings and further results about convolutional neural networks are included in the supplementary material.
% Since the family of Swin architectures has differently shaped intermediate representations, we leverage bilinear interpolation to match the shape during noise injection.
% We utilize CLIP-ViT-base-32/224~\cite{clip} architecture as a multi-modal model.

% \paragraph{Baselines}
\noindent \textbf{Feature Attribution Methods}
We compare the various feature attribution methods including Chefer-LRP~\cite{trans_attr}, Generic~\cite{generic}, IIA~\cite{iia}, ViT-CX~\cite{vit-cx}, Beyond~\cite{beyond}, and IBA~\cite{iba} as baselines.
% For all the baselines, we follow the settings recommended by the authors.
We select $\beta=10$ for IBA and $6$-th layer to insert bottleneck.
For \ours{}, we select trade-off hyper-parameter $\beta$ as 1.
We insert the bottleneck layer from $s$-th departure to $e$-th arrival layers, which are chosen, $4$, and $12$, respectively, for ViT-B.
For the optimization, we set the learning rate, batch size, and optimizer as 1, 10, and Adam~\cite{adam}.
The additional hyperparameter settings are included in the supplementary material.
% For three hyper-parameters required in \ours{}, $\beta$, and the indices of departure layer $s$ and arrival layers $e$, we set $\beta=1$ and restrict the information flow in 75\% of layers in the architecture.
% We provide the quantitative comparisons of different hyper-parameters in the supplementary material.
For Swin transformers, we only report the explanation methods that do not require methodology modification, including Generic, ViT-CX, IBA, Beyond, and \ours{}.
We utilize RTX A6000 GPU for all the experiments.

% \paragraph{Datasets}
\noindent \textbf{Datasets}
We utilize ImageNet-1k~\cite{imagenet} (IN-1k), ImageNet-A (IN-A)~\cite{imagenet_a} and ImageNet-R (IN-R)~\cite{imagenet_r} validation datasets for the experiments.
% We leverage IN-A and IN-R, which contain the difficult samples yielding low-confidence scores from the model. % to conduct the difficult-aware analysis.
We leverage IN-A and IN-R to conduct the difficult-aware analysis.
% In addition to the IN-1k dataset, we report the difficulty-aware experimental results leveraging ImageNet-A (IN-A)~\cite{imagenet_a} and ImageNet-R (IN-R)~\cite{imagenet_r} datasets, which include difficult data.
% We utilize these datasets to analyze the overall quality of attribution maps in various confidence scores computed by a model.

\subsection{FunnyBirds Assessment}
% \noindent \textbf{Funnybirds Assessment}
% In contrast to existing evaluation metrics for XAI, which assess the quality of an attribution map with only approximation, 
The FunnyBirds~\cite{funnybirds} experiment provides the ground truth for the evaluation.
With the help of ground truth, the Funnybird framework assesses the quality of the attribution map from three perspectives: \textit{completeness}, \textit{correctness}, and \textit{contrastivity}.
% The \textit{completeness} (Com.) assesses whether the attribution map (not) highlights the (ir)relevant parts of the birds.
% \textit{Correctness} (Cor.) measures whether the attribution correlates with the actual importance of each part.
% Finally, \textit{contrastivity} (Con.) measures the class discriminative ability of an attribution method by directly comparing the class-specific parts.
% With the success of constructing a ground truth-included dataset, the Funnybird framework assesses the quality of the attribution map from three perspectives: completeness, correctness, and contrastivity.
Fig.~\ref{fig:funnybirds} illustrates the qualitative and quantitative results of the Funnybirds experiment.
% We illustrate the qualitative and quantitative results of the FunnyBirds experiment in Fig.~\ref{fig:funnybirds}.
As shown in the results, \ours{} outperforms all the baselines, including the propagation or gradient-based approaches.
Furthermore, in contrast to IBA, \ours{} obtains an attribution map with significantly enhanced \textit{contrastivity}.
% We detail the individual scores including Com., Cor., and Con. dimensions in the supplementary material.

\begin{table*}[!ht]
\centering
\begin{centering}
\resizebox{0.87\linewidth}{!}{
\begin{tabular}{cccccccc}
\specialrule{2.0pt}{1pt}{1pt}
Model  & Chefer-LRP~\cite{trans_attr} & Generic~\cite{generic} & IIA~\cite{iia} & ViT-CX~\cite{vit-cx} & IBA~\cite{iba} & Beyond~\cite{beyond} & \ours{} \\ \hline
 ViT-B-16/224 & 20.87/64.37 & 22.15/65.31 & 19.58/66.19 & 19.04/65.36 & 20.99/\underline{68.80} & \underline{17.93}/68.44 & \textbf{16.63}/\textbf{73.68} \\ 
 DeiT-B-16/224 & 15.31/59.55 & 16.76/59.78 & \underline{14.74}/61.21 & 18.37/57.44 & 15.62/\underline{61.53} & 15.37/60.06 & \textbf{11.59}/\textbf{67.12} \\ 
 ViT-L-16/224 & 29.92/64.87 & 28.31/66.51 & 28.39/65.92 & \underline{24.30}/\underline{67.27} & 30.55/66.97 & 25.47/66.68 & \textbf{22.03}/\textbf{74.88} \\ 
 DeiT3-L-16/224 & - & 20.28/69.87 & 19.72/70.02 & 24.43/66.60 & 19.56/\underline{71.23} & \underline{19.41}/70.52 & \textbf{15.02}/\textbf{76.13} \\ 
 ViT$^\dagger$-H-16/224 & - & 30.06/61.34 & 32.27/60.98 & 30.34/57.70 & 28.30/62.29 & \underline{28.11}/\underline{62.96} & \textbf{23.38}/\textbf{66.95} \\ 

\specialrule{2.0pt}{1pt}{1pt}
\end{tabular}
}
\vspace{-7pt}
\caption{\textbf{Quantitative feature importance evaluation of ROAD (MoRF $\downarrow$ / LeRF $\uparrow$).} The higher and lower scores indicate better attribution map quality for MoRF and LeRF, respectively. The underlined scores indicate the highest performance among the baselines. $(^\dagger)$ denotes the ViT trained with CLIP~\cite{clip}.
% We provide further results in the supplementary material.
}
\vspace{-12pt}
% \end{adjustbox}
\label{tab:road_results1}
\end{centering}
\end{table*}

\begin{figure*}[th!]
\centering

% \subfloat[]{
% \includegraphics[width=.232\linewidth]{Assets/road_result2_vit-b.pdf}}
% \subfloat[]{
% \includegraphics[width=.18\linewidth]{Assets/road_result2_deit-b.pdf}}
% \subfloat[]{
% \includegraphics[width=.18\linewidth]{Assets/road_result2_vit-l.pdf}}
% \subfloat[]{
% \includegraphics[width=.18\linewidth]{Assets/road_result2_deit3-l.pdf}}
% \subfloat[]{
% \includegraphics[width=.18\linewidth]{Assets/road_result2_vit-h.pdf}}

\subfloat[IN-1k]{
\includegraphics[width=.25\linewidth]{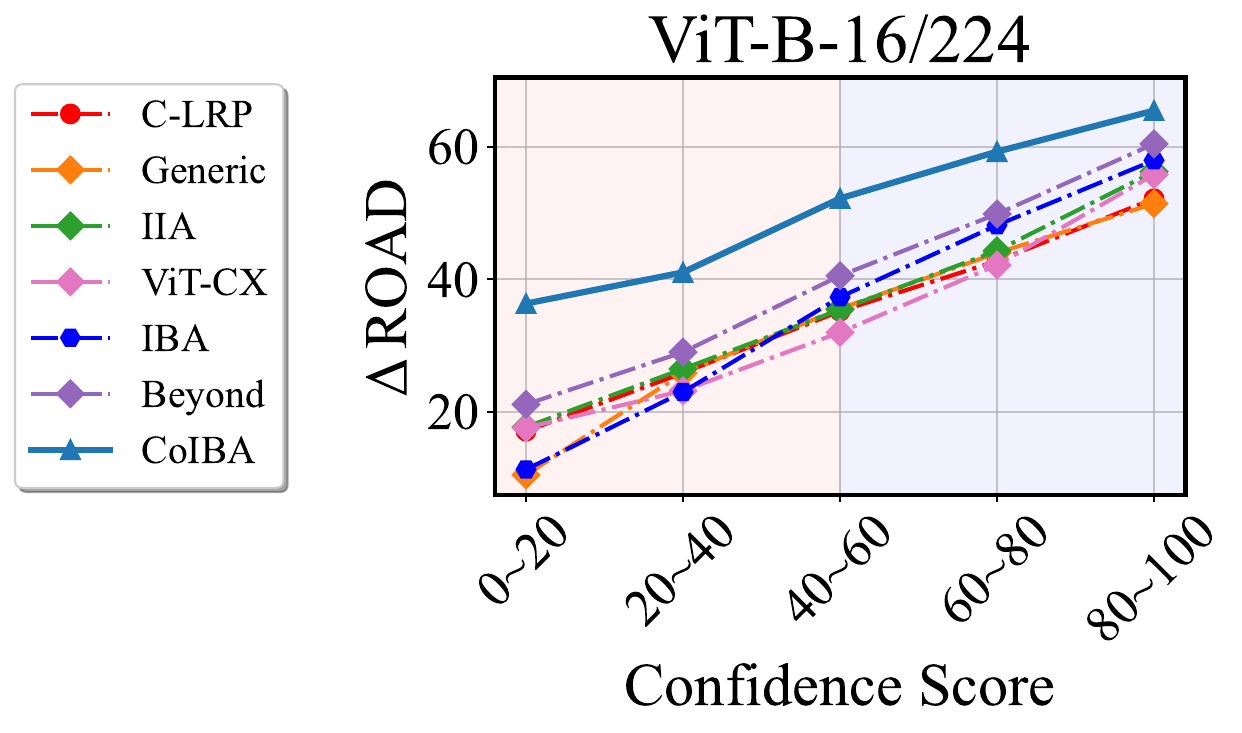}
\includegraphics[width=.18\linewidth]{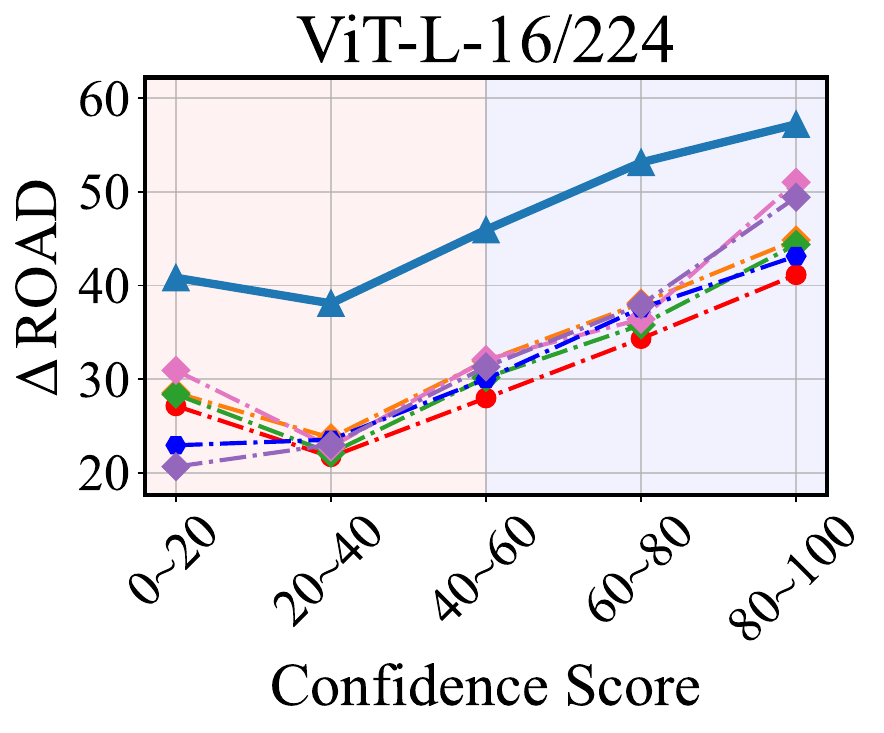}
\includegraphics[width=.18\linewidth]{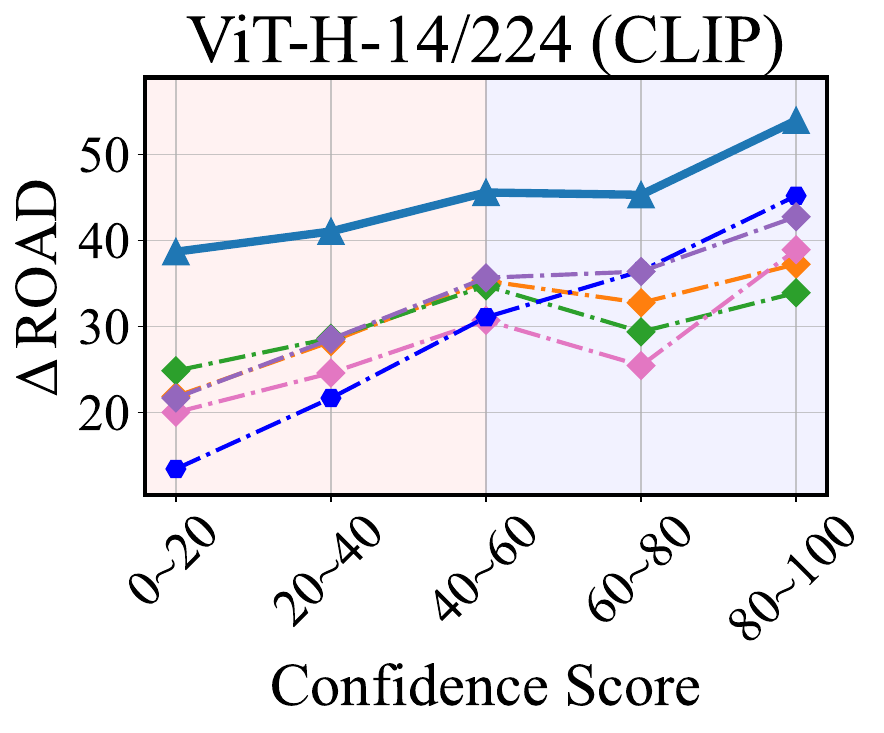}
\vspace{-2pt}
}
% \hspace{0.06\linewidth}
\subfloat[IN-A]{
\includegraphics[width=.18\linewidth]{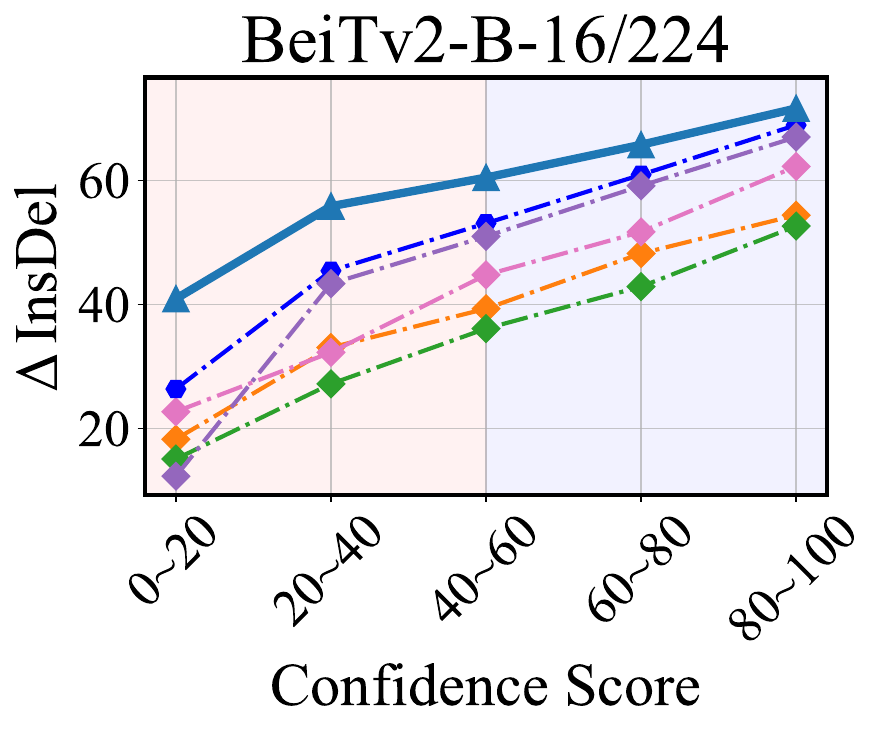}
\vspace{-2pt}
}
\subfloat[IN-R]{
\includegraphics[width=.18\linewidth]{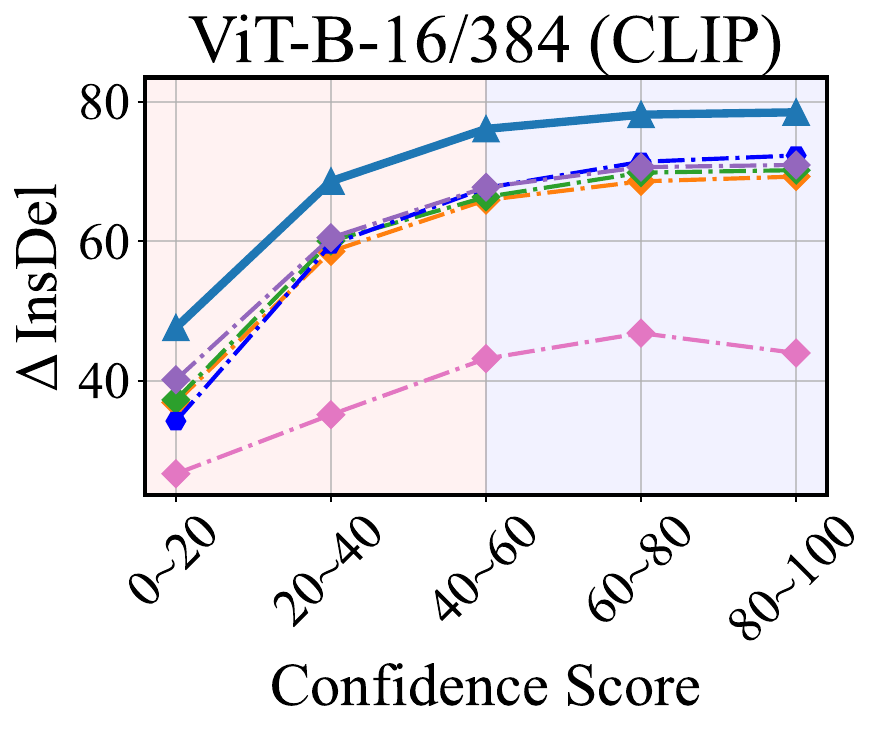}
\vspace{-2pt}
}
% \vspace{-7pt}
% \vspace{-7pt}
\vspace{-10pt}

\caption{\textbf{Quantitative difficulty-aware correctness assessment on insertion/deletion and ROAD.}
We measure differences in insertion/deletion ($\Delta$InsDel$\uparrow$) and ROAD ($\Delta$ROAD$\uparrow$) scores. We fill the regions including low-confident samples and high-confident samples with red and blue, respectively, based on the prediction made by the model. The first to third columns illustrate the quantitative results of the IN-1k. The fourth and fifth columns include the results of IN-A and IN-R datasets, respectively.
% We include additional results in the supplementary material.
}
% 가로축이 뭔지 확실히 얘끼할 것.
\vspace{-18pt}
\label{fig:difficulty_aware}
\end{figure*}

\subsection{Insertion/Deletion}
Insertion/deletion~\cite{rise} measures the correctness of feature importance highlighted by an attribution map.
% by deleting or inserting the input variables according to their contribution plotted in the attribution.
The insertion/deletion method gradually inserts or deletes informative pixels in ascending order of attribution to compute the area under the curve scores.
% Thus methods that achieve higher and lower scores are considered as better methods in insertion/deletion, respectively.
In insertion/deletion, methods providing better correctness performance achieve higher scores for insertion and lower scores for deletion.
We gradually deleted or inserted around every 3.5\% pixels of the input image and randomly sampled 6,000 samples from the IN-1k validation dataset for the assessment.
As shown in Table~\ref{tab:insdel}, the attributions provided by \ours{} yield predominant correctness performance compared to baselines.
% Especially, \ours{} produces the qualified attribution maps with differently designed architecture, Swin, demonstrating the general ability of \ours{}.
% Aligning with this result, \ours{} succeeds in providing the high correct performance regardless of the depth of the model.
Especially, the consistent increase in correctness from the Swin transformer demonstrates the generalizability of \ours{}.

\subsection{Remove and Debias}
The remove-and-debias (ROAD)~\cite{road} experiments measure the correctness of the attribution maps, addressing the class information leakage problem (\textit{i.e.}, the shape of the mask) observed in the remove-and-retrain (ROAR)~\cite{roar} experiment.
% To address this limitation, ROAD relieves the class leakage problem and omits the re-training process in evaluation.
% ROAD imputes the image regions corresponding to the relevancy in terms of most relevant first (MoRF), and least relevant first (LeRF).
% Following the recommended settings provided by the authors, we employ the noisy linear imputation method for perturbation.
We average the quantitative scores over the 6,000 randomly selected images when 20, 40, 60, and 80\% of pixels are imputed in two perspectives: most relevant first (MoRF) and least relevant first (LeRF).
As shown in Table~\ref{tab:road_results1}, \ours{} outperforms existing methods in MoRF and LeRF experiments with a sizable gap.
These results demonstrate that regardless of the imputation type, \ours{} outperforms all the baselines regardless of the model setting (patch size and depth) and pre-trained dataset.
% Importantly, \ours{} surpasses performance in both MoRF and LeRF, outperforming the various explanation methods and models.

% \input{tables/ehr}

\subsection{Difficulty-aware Correctness Assessment}
\label{exp:difficulty-aware}
% We demonstrate the performance of \ours{} in providing attribution maps with the input samples of diversified difficulties.
We compare the quantitative quality of attribution maps yielded by \ours{} and baseline methods in terms of difficulty-aware assessment, in addition to correctness assessments of overall samples.
Leveraging the confidence scores, and the prediction probability of a target class, we divide the confidence scores with 20 intervals as in Sec.~\ref{sec:motivation}.
As shown in Fig.~\ref{fig:difficulty_aware}, \ours{} consistently outperforms all the baselines regardless of the confidence scores predicted with each sample, predicted by the model.
Along with IN-1k, we include the quantitative results of IN-A and IN-R.
As both IN-A and IN-R datasets include the low-confident (difficult) samples compared to IN-1k, these quantitative results support the demonstration that \ours{} outperforms the correct performance with low-confident samples.

\begin{figure}[t!]
\centering
\vspace{-3pt}
\subfloat[Ablation on departure layer]{
\includegraphics[width=.22\textwidth]{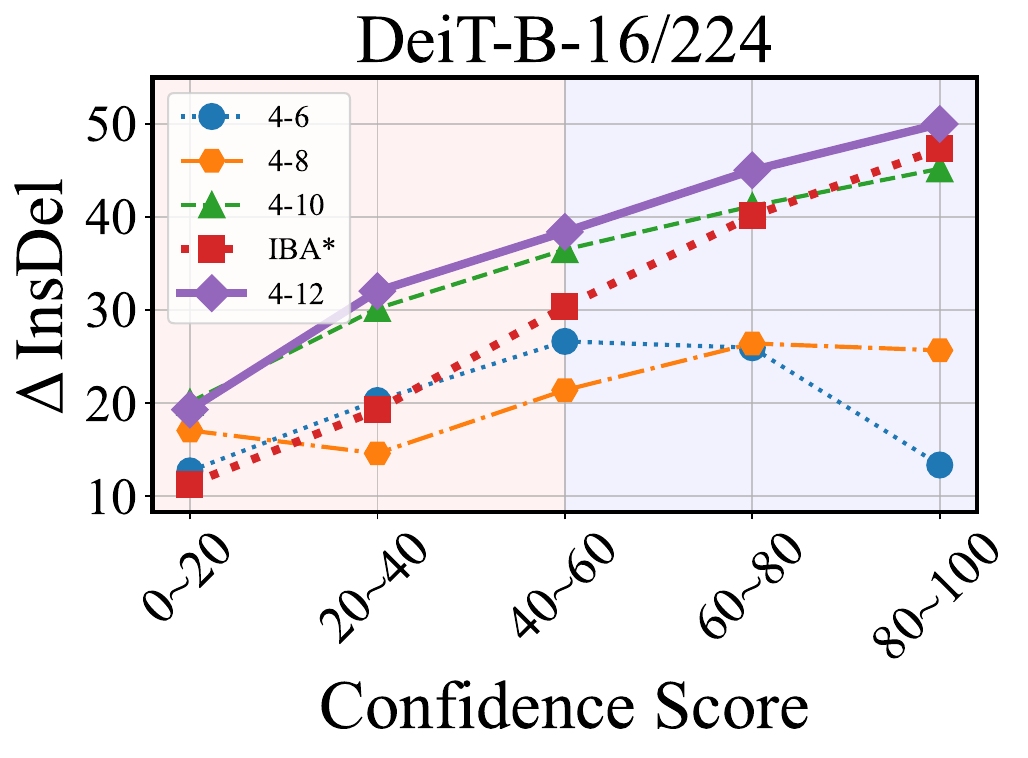}
\vspace{-3pt}
}
\subfloat[Ablation on arrival layer]{
\includegraphics[width=.22\textwidth]{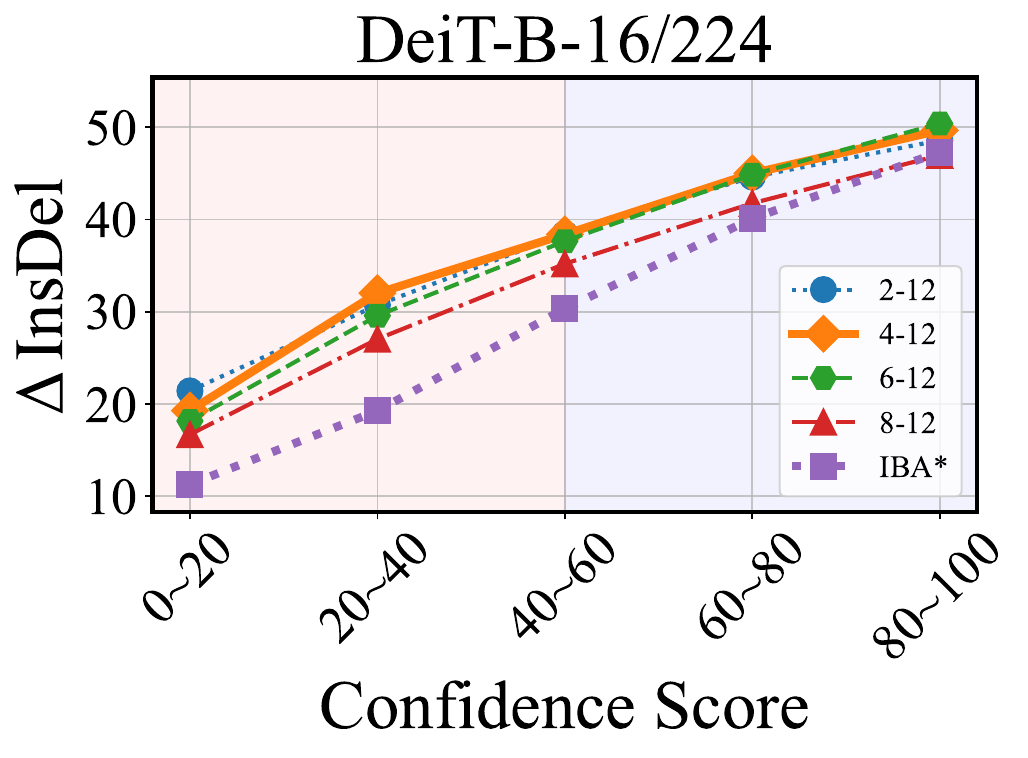}
\vspace{-3pt}
}

\vspace{-10pt}
\caption{\textbf{Comparisons of different departure $s$ and arrival $e$ layer settings.}
% The difference in insertion and deletion scores ($\Delta$InsDel) for different hyperparameter settings $s$ and $e$, meaning the interpolating depart and arrival layers, respectively. 
We plot the hyper-parameters selected for \ours{} as a solid line.
We compare the quantitative results of the discrepancy between AUC scores of insertion/deletion ($\Delta$InsDel) with different intervals of confidence scores outputted by the model.
The better attribution method achieves a higher score.
We provide further results in the supplementary material.}
\vspace{-15pt}
\label{fig:hyperparam1}

\end{figure}

\subsection{Discussion}
\label{sec:discussion}
\begin{figure}[t!]
\centering
\vspace{2pt}
\subfloat[Discrepancy in mutual info.]{
\label{fig:discussion1_1_mi}
\includegraphics[width=.45\linewidth]{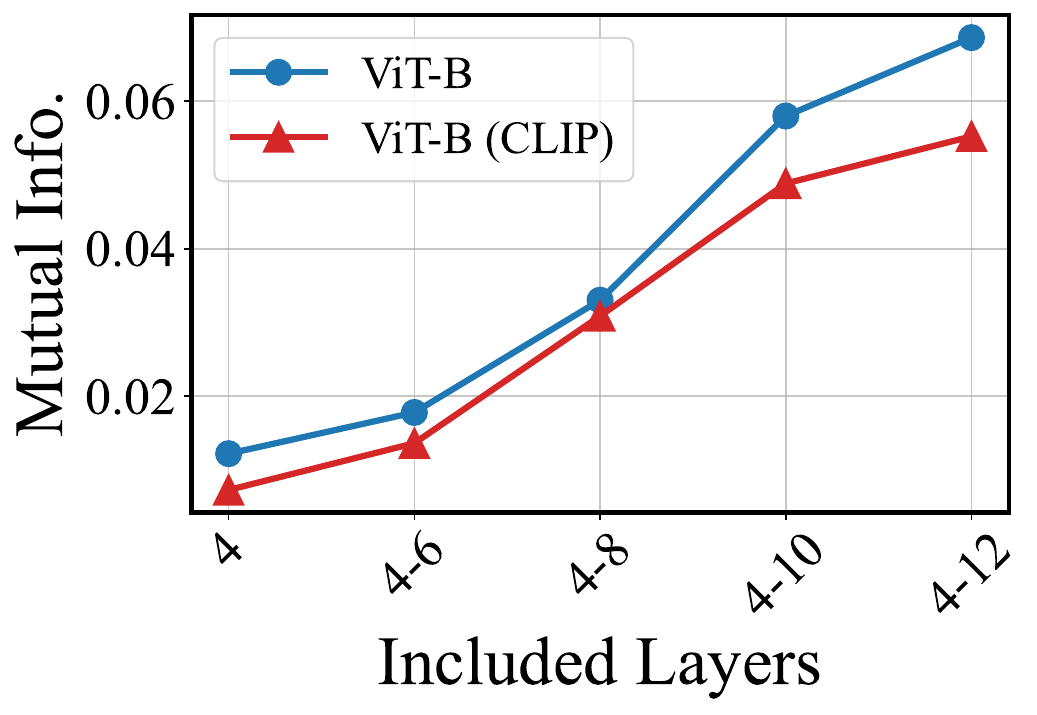}
\vspace{-3pt}
}
\subfloat[Discrepancy in accuracy]{
\label{fig:discussion1_2_acc}
\includegraphics[width=.45\linewidth]{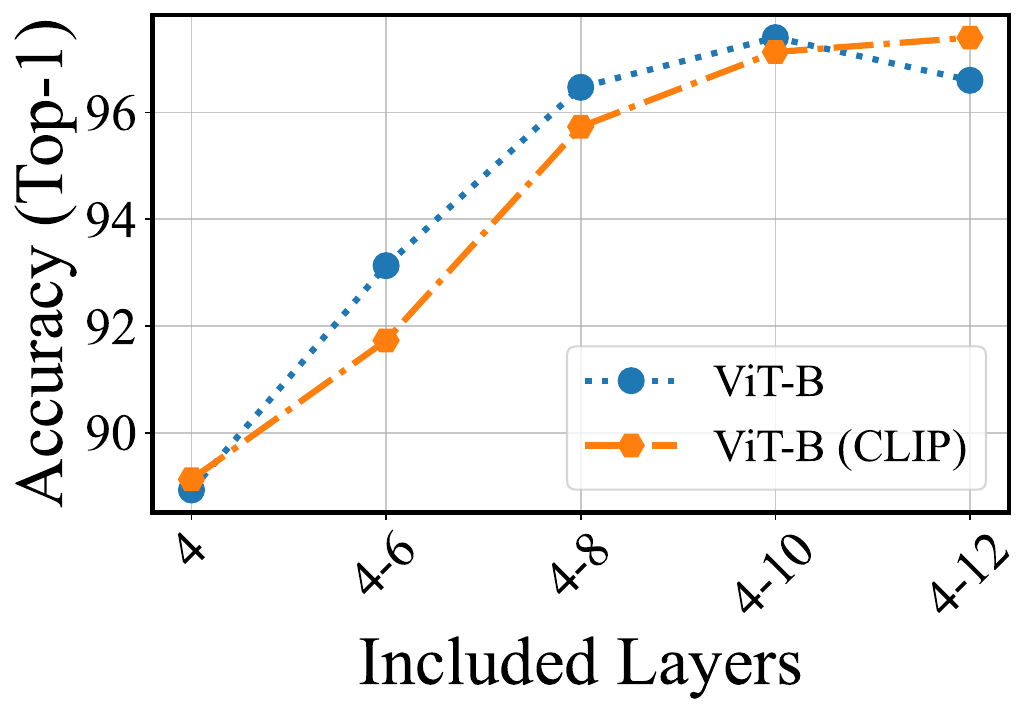}
\vspace{-3pt}
}

\subfloat[Ablation on universal damping ratio]{
\label{fig:discussion1_3_correctness}
\includegraphics[width=.46\linewidth]{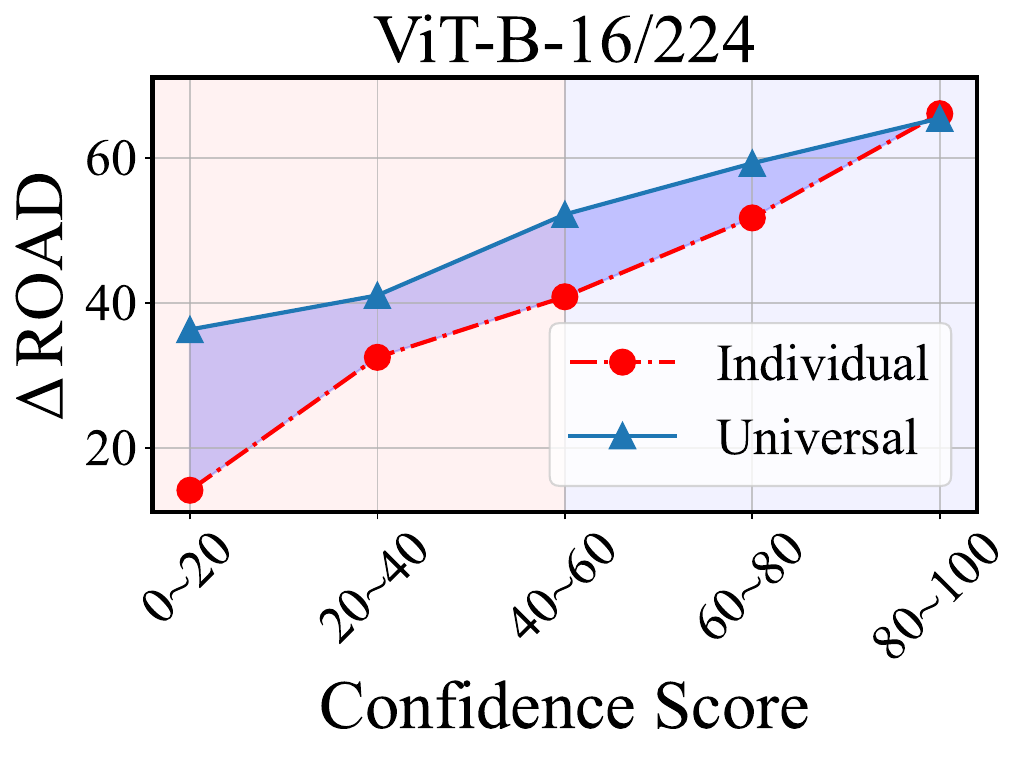}
\vspace{-1pt}
\includegraphics[width=.46\linewidth]{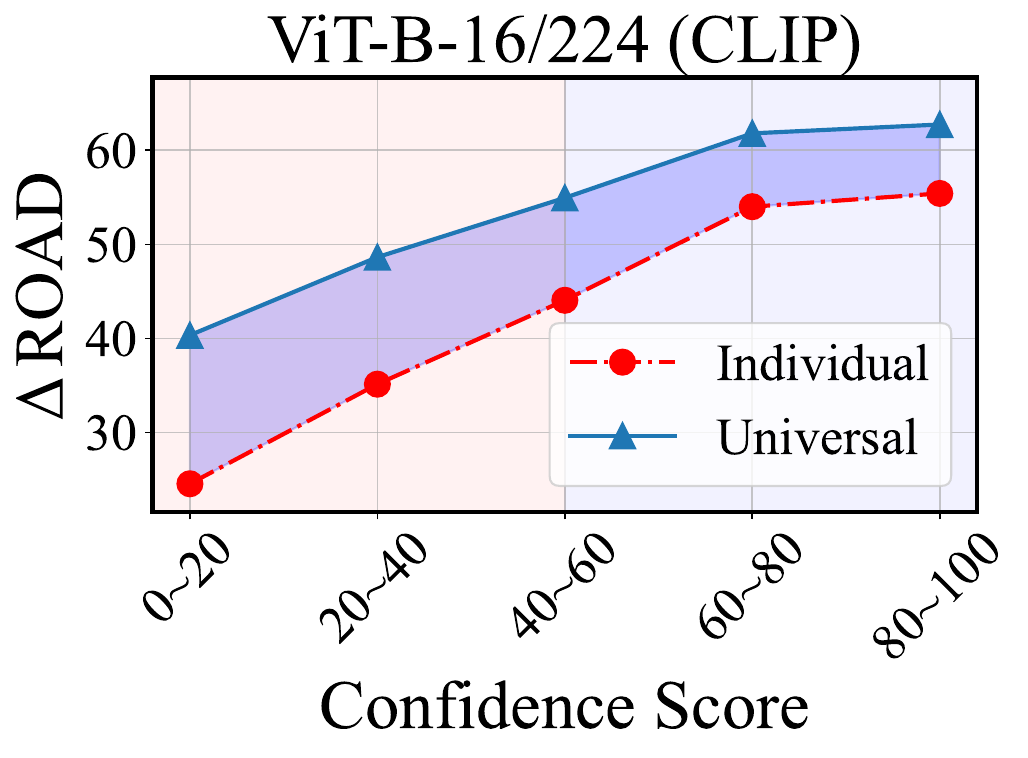}
\vspace{-1pt}
}

\vspace{-8pt}
\caption{
\textbf{Ablation study on layers to analyze universal damping ratio.}
% We provide the change in mutual information in Fig.~\ref{fig:discussion1}\subref{fig:discussion1_1_mi} and top-1 accuracy in Fig.~\ref{fig:discussion1}\subref{fig:discussion1_2_acc} by ablating the number of bottleneck-inserted layers.
Fig.~\ref{fig:discussion1}\subref{fig:discussion1_1_mi} illustrates the change in mutual information for compression terms $I[Z_1; R_1]$ in Eq.~\eqref{eq:main2} and the Fig.~\ref{fig:discussion1}\subref{fig:discussion1_2_acc} shows the change in accuracy.
We utilize the ViT-B-16/224 model for these experiments.
The mutual info. denotes mutual information.
% We report the results by dividing the scores by the confidence scores computed by the model.
% For the comparison, we utilize ViT-B architecture pre-trained with IN-22k and CLIP datasets.
% We filled the discrepancy in performances between individual and universal with blue color in Fig.~\ref{fig:discussion2_2} and ~\ref{fig:discussion2_3}.
% The solid line indicates the results of \ours{}.
% We compare the difference in ROAD score ($\Delta$).
}
\vspace{-15pt}
\label{fig:discussion1}
\end{figure}

\noindent \textbf{Ablation on Bottleneck-inserted Layers}
% In this section, we introduce the empirical experiment to explain the selection of hyperparameters in \ours{}.
In this section, we confirm whether including multiple layers enhances the correctness of the resulting attribution map.
% We measure whether including the layers while computing the attribution maps enhances the quality of an attribution map.
Fig.~\ref{fig:discussion1} shows the comparison of the quantitative results by interpolating departure ($s$) and arrival ($e$) layers from earlier to deeper layers.
% As \ours{} includes two types of hyper-parameters: departure ($s$) and arrival ($e$) layers, we compare the quantitative results by interpolating $s$ and $e$ from earlier to deeper layers.
The results demonstrate that increasing the number of layers to compute relevant information consistently enhances the correctness quality of an attribution map.
% both hyper-parameters to compare the quantitative results among the different confidence scores.
% For example, if $s$ and $e$ are defined as $2$ and $10$, \ours{} restricts the information flow from +$2$nd to $10$th layers.
% We compare the quantitative results of the discrepancy between insertion/deletion game scores with different intervals of confidence scores yielded by the ViT model.
% The better-qualified attribution map yields higher discrepancy in 6insertion/deletion games.
% The difference in scores of each confidence score indicates the quality of the attribution map for different difficulty levels.
% As shown in Fig.~\ref{fig:hyperparam1}, increasing the number of included layers while computing relevant information enhances the correctness of the attribution map regardless of the confidence scores.
% Regardless of departure or arrival layers, the quality of attribution is consistently amplified as the number of bottleneck-inserted layers is increased.
% These results demonstrate that considering the increasing number of layers reveals the omitted relevant information.
% Furthermore, \ours{} consistently outperforms IBA* which requires the iteration over the layers with large margins, requiring only a single iteration.
Furthermore, \ours{} consistently outperforms IBA* with large margins, requiring only a single iteration.
Thus, in \ours{}, various layers gladly reveal the omitted relevant information obtained from a specific layer.
% \input{figs/qualitative}

% As easy samples (confidence scores within 80 to 100) are highly included in the dataset, we include the $4$ and $12$ as $s$ and $e$.

% \input{figs/discussion2_3}

% \begin{figure}[t!]
% \centering

% \includegraphics[width=.48\linewidth]{Assets/discussion2_2_vit-b.pdf}
% \vspace{-2pt}
% \includegraphics[width=.48\linewidth]{Assets/discussion2_2_vit-b-clip.pdf}
% \vspace{-2pt}

% \vspace{-8pt}
% \caption{
% \textbf{Quantitative Comparison  --  Effectiveness of Channel Uniform Perturbation:}
% We report the results by dividing the scores by the confidence scores computed by the model.
% For the comparison, we utilize ViT-B architecture pre-trained with IN-22k and CLIP datasets.
% We compare the difference in ROAD score ($\Delta$).
% The solid line indicates the results yielded by \ours{}.
% }
% \vspace{-13pt}
% \label{fig:discussion2_2}
% \end{figure}

% \paragraph{Synchronized vs Non-synchronized \ours{}}
\noindent \textbf{Effectiveness of Universal Damping Ratio}
\label{sec:discussion_universal}
% To analyze the effectiveness of our synchronized setting, we compare the $\Delta$ROAD scores of information restrictions utilizing universal or individual damping ratio.
% We report the results by dividing the scores by the confidence scores made by the model.
% Similar to the investigation in Sec.~\ref{sec3-2}, we divide the cases in terms of the difficulty of input samples.
% In this section, we analyze the effectiveness of leveraging the individual and universal damping ratios for the information restriction.
We investigate whether leveraging the universal damping ratio compensates for over-compression in Fig.~\ref{fig:discussion1}\subref{fig:discussion1_1_mi} while amplifying the relevant information in Fig.~\ref{fig:discussion1}\subref{fig:discussion1_2_acc}.
% Fig.~\ref{fig:discussion1}\subref{fig:discussion1_1_mi} shows the mutual information estimated from different numbers of targeted layers.
We compare the different numbers of targeted layers to compare the quantitative results.
The results illustrated in Fig.~\ref{fig:discussion1}\subref{fig:discussion1_1_mi} show that leveraging the universal damping ratio complements the over-compressed information that occurred in earlier layers, due to the delivered relevant information from deeper layers.
Concurrently, as shown in Fig.~\ref{fig:discussion1}\subref{fig:discussion1_3_correctness}, leveraging the universal damping ratio amplifies the relevant information term in Eq~\eqref{eq:main2}.
% As shown in Fig.~\ref{fig:discussion1}\subref{fig:discussion1_2_acc}, as the accuracy is consistently increased, considering multiple layers amplifies the relevant information term in Eq~\eqref{eq:main2}.
% Therefore, as \ours{} broadcasts the relevant information between targeted layers through the universal damping ratio, the universal damping ratio addresses the over-estimation issue while compensating the relevant term in Eq.~\eqref{eq:main2}.
% Therefore, \ours{} enhances the correctness of the attribution map by delivering the relevant information from deeper to earlier layers through the universal damping ratio, addressing the over-estimation issue in earlier layers.
Aligning with these results, compared by assigning individual damping ratios to each layer, our universal damping ratio enhances the correctness performance of \ours{}.
Therefore, leveraging the universal damping ratio to targeted multiple layers significantly enhances the correctness performance of \ours{} by compensating for the over-compression and amplifying the relevant information.

% For the individual damping ratio setting, we assign the individual damping ratio which is not shared between layers.
% We illustrate the quantitative results in Fig.~\ref{fig:discussion1}.
% We confirm the aforementioned compensation enhances the correctness of the resulting attribution map.
% In addition to the aforementioned effectiveness of \ours{}, we compare whether targeting multiple layers concurrently enhances the correctness performance of the resulting attribution map.
% As shown in Fig.~\ref{fig:discussion1}\subref{fig:discussion1_3_correctness}, compared to individual damping ratio, computing the relevant information with universal damping ratio enlarges the correctness scores with a sizable gap.
% % Despite the correctness performance in the ViT-B model being similar in high-confident samples, the performance of the individual damping ratio is sharply diminished for low-confident samples.
% In particular, in contrast to the high-confident samples, the performance of the individual damping ratio on the ViT-B model is sharply diminished for low-confident samples.
% Therefore, targeting multiple layers significantly enhances the correctness performance of \ours{} by compensating the over-estimation and amplifying the relevant information.
% In earlier layers, the mutual information tends to be over-estimated, losing the relevant information due to the over-compression~\cite{inputiba}.
\begin{figure}[t!]
\centering

% \subfloat[Ablation on channel uniform perturbation]{
% \label{fig:discussion2_1}
% \includegraphics[width=.46\linewidth]{Assets/discussion2_2_vit-b.pdf}
% \vspace{-2pt}
% \includegraphics[width=.46\linewidth]{Assets/discussion2_2_vit-b-clip.pdf}
% \vspace{-2pt}
% }

% \subfloat[Ablation on universal damping ratio]{
% \label{fig:discussion2_2}
% \includegraphics[width=.46\linewidth]{Assets/discussion2_1_vit-b.pdf}
% \vspace{-2pt}
% \includegraphics[width=.46\linewidth]{Assets/discussion2_1_vit-b-clip.pdf}
% \vspace{-2pt}
% }

% \subfloat[Ablation on universal damping ratio]{
% \label{fig:discussion2_2}
% \includegraphics[width=.4\linewidth]{Assets/discussion2_1_vit-b.pdf}
% \vspace{-2pt}
% \includegraphics[width=.4\linewidth]{Assets/discussion2_1_vit-b-clip.pdf}
% \vspace{-2pt}
% }

% \subfloat[Effectiveness of Variational Upper Bound]{
% \label{fig:discussion2_3}
% \includegraphics[width=.48\linewidth]{Assets/discussion2_3_vit-b.pdf}
% \vspace{-2pt}
% \includegraphics[width=.48\linewidth]{Assets/discussion2_3_vit-b-clip.pdf}
% \vspace{-2pt}
% }
% \subfloat[Faithfulness assessment]{
% \label{fig:discussion2_3}
% \includegraphics[width=.46\linewidth]{Assets/discussion2_3_vit-b.pdf}
% \vspace{-2pt}
% }
% \subfloat[Similarity measurement]{
% \label{fig:discussion2_3_cka}
% \includegraphics[width=.46\linewidth]{Assets/discussion_cka1.pdf}
% \vspace{8.5pt}
% } 

\subfloat[Faithfulness assessment]{
\label{fig:discussion2_3}
\includegraphics[width=.455\linewidth]{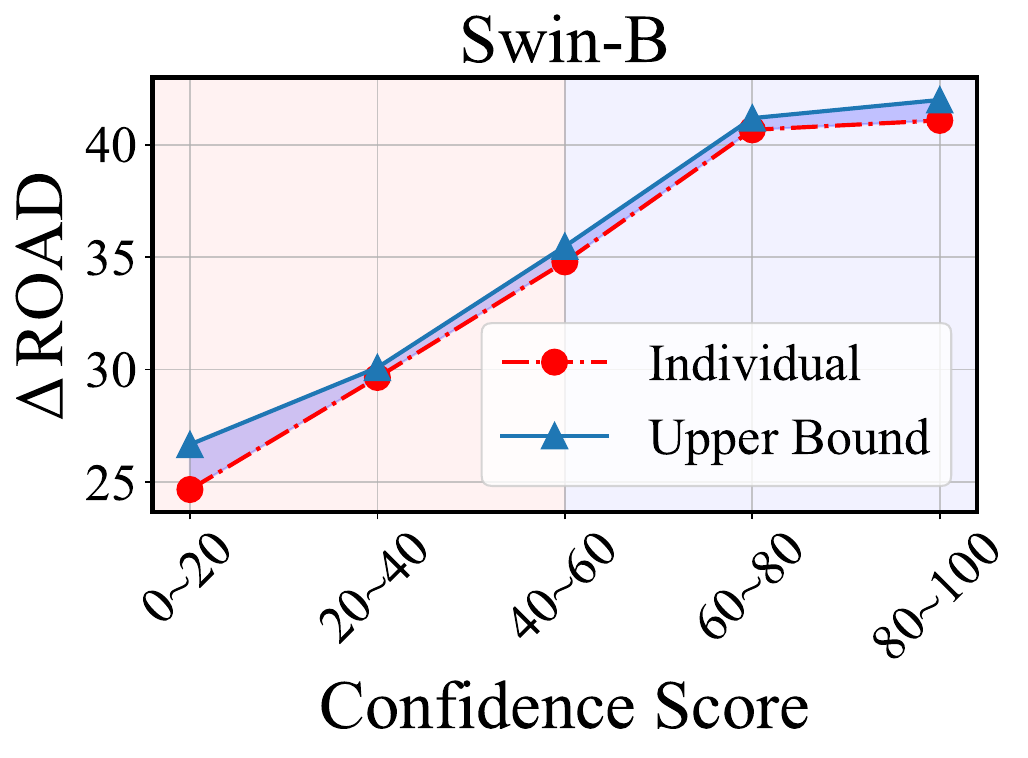}
\vspace{-3pt}
}
\hspace{-10pt}
\subfloat[Similarity measurement]{
\label{fig:discussion2_3_cka}
\includegraphics[width=.455\linewidth]{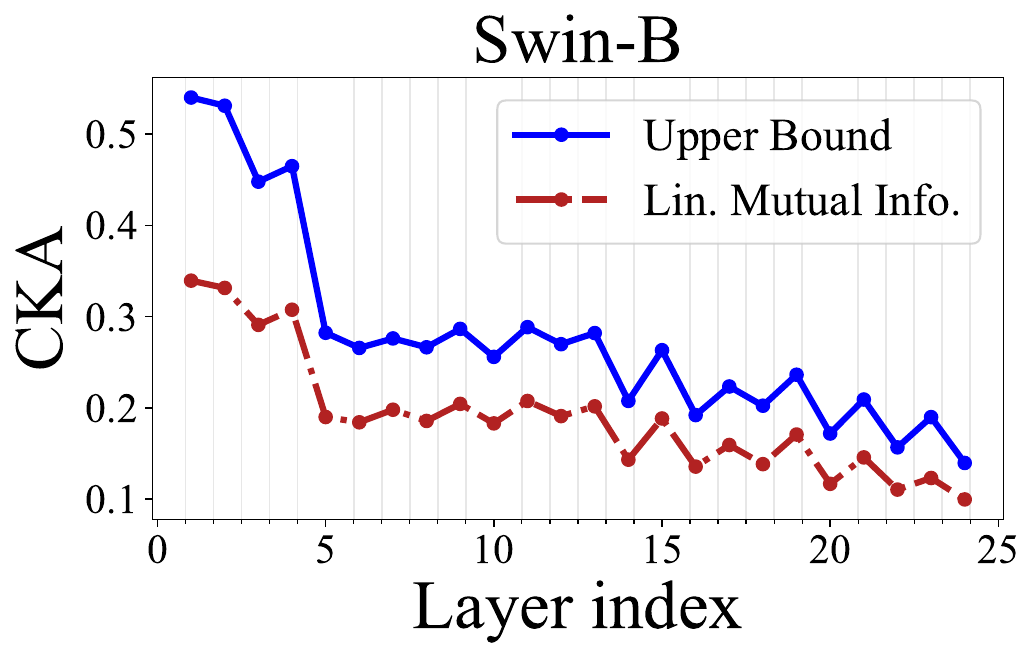}
\vspace{7.5pt}
}

\subfloat[Ablation on channel uniform perturbation]{
\label{fig:discussion2_1}
\includegraphics[width=.455\linewidth]{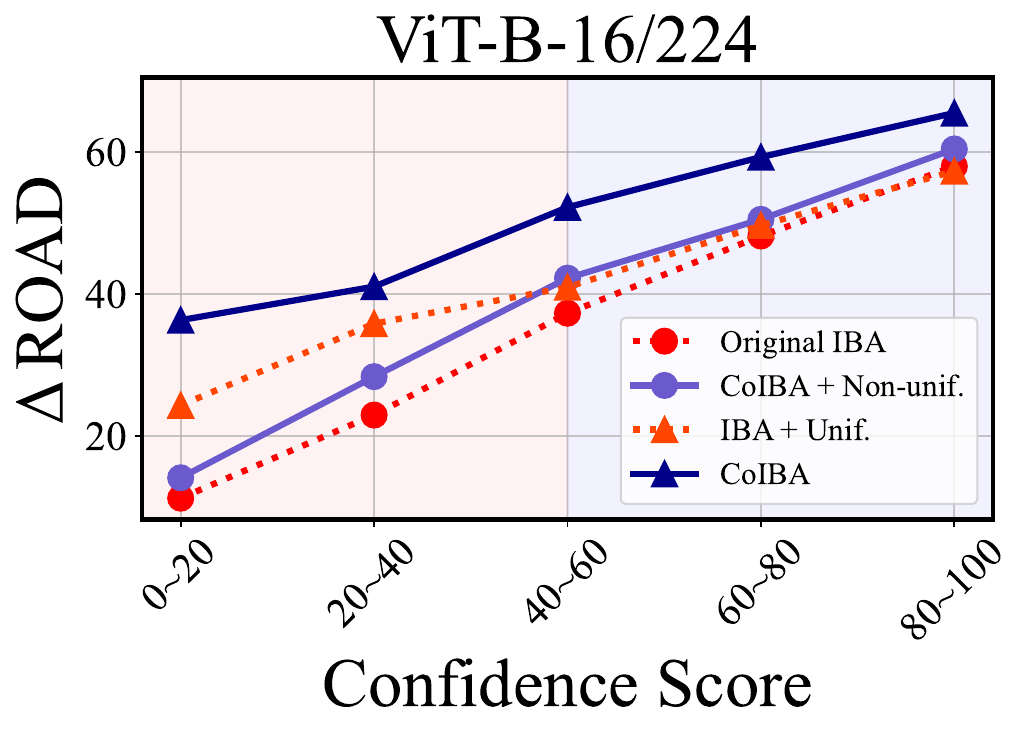}
\vspace{-1pt}
\includegraphics[width=.455\linewidth]{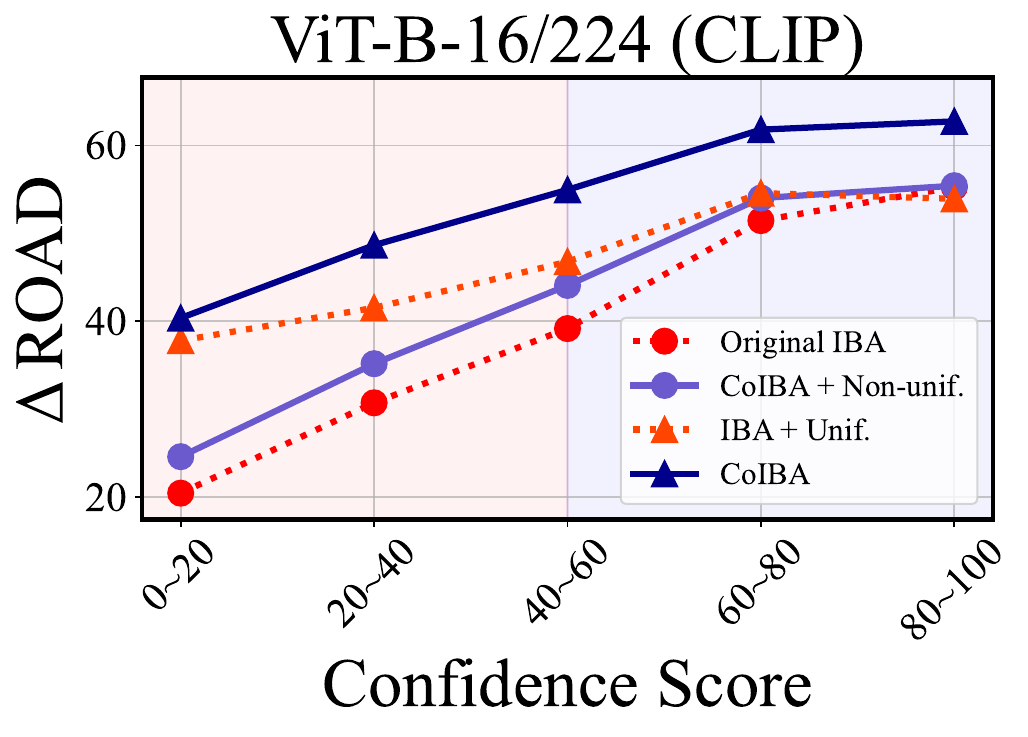}
\vspace{-1pt}
}

\vspace{-7pt}
\caption{
\textbf{Quantitative analysis on \ours{}.}
We report the results divided by the confidence score computed by the model.
For the comparison, we include the ViT-B model pre-trained with CLIP in Fig.~\ref{fig:discussion2}\subref{fig:discussion2_1}.
We filled the discrepancy in performances with blue color in Fig.~\ref{fig:discussion2}\subref{fig:discussion2_1}.
The solid line indicates the results of \ours{}.
% We compare the difference in ROAD score ($\Delta$).
}
\vspace{-14pt}
\label{fig:discussion2}
\end{figure}

\noindent \textbf{Effectiveness of Variational Upper Bound}
\label{sec:discussion_upperbound}
% We demonstrate the effectiveness of our variational upper bound by measuring correctness performance and whether the relevancy of different layers is fairly reflected in attributions.  %% Rebuttal 수정 전
We demonstrate the effectiveness of our variational upper bound by measuring correctness performance and whether the relevancy of different layers is fairly reflected in attributions.
%compared to our objective in Eq.~\ref{eq:main1}.  %% Rebuttal 후 수정
We compare quantitative results obtained by leveraging a variational upper bound (Eq.~\eqref{eq:main2}) versus linearly combining layer-wise mutual information (Eq.~\eqref{eq:main1}) in Fig.~\ref{fig:discussion2}\subref{fig:discussion2_3}.
% Fig.~\ref{fig:discussion2}\subref{fig:discussion2_3} is better at leveraging our variational upper bound approximates identifying feature importance, regardless of confidence scores, as shown in the results.
As shown in the results, leveraging our variational upper bound is better at correctly identifying feature importance, regardless of confidence scores.
% We compare quantitative results obtained by leveraging a variational upper bound (Eq.~\eqref{eq:main2}) versus linearly combining layer-wise mutual information (Eq.~\eqref{eq:main1}) in Fig.~\ref{fig:discussion2}\subref{fig:discussion2_3}.
% As shown in the results, leveraging our variational upper bound successfully approximates identifying feature importance, regardless of confident scores.
% In addition to this result, we confirm whether the relevant information of each layer is fairly reflected in the attribution map.  %% Rebuttal 수정 전
In addition to this result, we confirm whether the relevant information of each layer is fairly reflected in the attribution map by comparing similarities.  %% Rebuttal 수정 전
Concretely, we compare the similarity of both settings (Eq.~\eqref{eq:main1} and~\eqref{eq:main2}) with layer-specific relevant information computed by iterating IBA with channel uniform perturbation, denoted as IBA$^*$.
We employ CKA~\cite{cka} as a similarity metric.
% As shown in Fig.~\ref{fig:discussion2}\subref{fig:discussion2_3_cka}, leveraging our upper bound yields the relevant information less biased to the layers.  %% Rebuttal 수정 전
As shown in Fig.~\ref{fig:discussion2}\subref{fig:discussion2_3_cka}, leveraging our upper bound yields the comprehensive relevant information of all the layers.  %% Rebuttal 수정 전
% As shown in Fig.~\ref{fig:discussion2}\subref{fig:discussion2_3_cka}, leveraging our upper bound shows similar relevant information compared to the linear combination.  %% Rebuttal 후 수정
% We compare the computational cost in the Supplementary.
This result demonstrates that our variational upper bound encourages the resulting comprehensive relevant information to fairly reflect the relevancy of each layer.

\noindent \textbf{Effectiveness of Uniform Channel Perturbation}
% In this section, we compare the correctness performance of uniform (unif.) and non-uniform (non-unif.) channel perturbation strategies to demonstrate the effectiveness of uniform channel perturbation.
We demonstrate the effectiveness of uniform channel perturbation by comparing the correctness performance of uniform (unif.) and non-uniform (non-unif.) channel perturbation strategies.
% For analysis, we utilize ViT-B architecture pre-trained with IN-22k and CLIP datasets to compare the difference in ROAD scores.
As shown in Fig.~\ref{fig:discussion2}\subref{fig:discussion2_1},  the uniform channel perturbation consistently enhances the correctness performance of IBA for low-confident samples.
% In particular, compared to \ours{} leveraging uniform and non-uniform channel perturbation strategies, uniform channel perturbation significantly enlarges the correctness of resulting attribution maps, across various confident samples.
In addition to this result, uniform channel perturbation significantly enlarges the correctness of resulting attribution maps, regardless of the confidence scores.
% Therefore, our uniform channel perturbation enhances the correctness performance regardless of the method.
These enhanced correctness performances demonstrate the effectiveness of our uniform channel perturbation, regardless of the method.
% In particular, attribution maps of low-confident input samples are consistently enhanced.

% \subsection{Qualitative Results}
% In this section, we compare qualitative visualizations against the different baselines.
% Blue and red colors indicate negative and positive attributions, respectively.
% Fig.~\ref{fig:qualitatative} shows the results for comparison of five state-of-the-art explanatory methods with various architectures.
% As shown in the results, produced attributions of the \textit{Hammer Head} and \textit{Junco} classes correctly highlight the class-discriminative features of the facial region.
% Furthermore, \ours{} provides the visualization for \textit{Prairie chicken} and \textit{Hammerhead} classes, not highlighting the background and focusing on the class-discriminative object.

% Localization
% \vspace{-3pt}
\section{Conclusion}
% \vspace{-3pt}

In this paper, we introduce \ours{}, which reveals the comprehensive relevant information to produce an attribution map, revealing the omitted relevancy in IBA with a theoretical guarantee.
\ours{} shares the universal damping ratio to compensate for the over-compressed information, delivering relevancy among the bottleneck-inserted layers.
\ours{} clearly judges the importance of each token by leveraging uniform channel perturbation.
We leverage variational approximation to upper bound the information to ensure the eliminated activations are not necessary for the bottleneck-inserted layers to make a decision.
% Thereby, \ours{} obviates the need for heuristic-based identification of the layer that provides faithful attributions, thereby reducing the unreliability inherent in layer-specific relevant information.
\ours{} demonstrates a substantial improvement over existing methods in numerous experiments and discussions.
% In particular, \ours{} outperforms in highlighting important features regardless of the difficulty of each input sample.
% highlighting the effectiveness of \ours{} in generating explanations step-by-step.

% \vspace{7pt}
\newpage
\section*{Acknowledgement}
This work was supported by the Institute of Information \& Communications Technology Planning \& Evaluation (IITP) grant, funded by the Korea government (MSIT) (No. RS-2019-II190079, Artificial Intelligence Graduate School Program (Korea University), IITP-2025-RS-2024-00436857, ITRC (Information Technology Research Center), and No. RS-2022-II220984, Development of Artificial Intelligence Technology for Personalized Plug-and-Play Explanation and Verification of Explanation).
{
    \small
    \bibliographystyle{ieeenat_fullname}
    \bibliography{main}

\begin{thebibliography}{42}
\providecommand{\natexlab}[1]{#1}
\providecommand{\url}[1]{\texttt{#1}}
\expandafter\ifx\csname urlstyle\endcsname\relax
  \providecommand{\doi}[1]{doi: #1}\else
  \providecommand{\doi}{doi: \begingroup \urlstyle{rm}\Url}\fi

\bibitem[Abnar and Zuidema(2020)]{rollout}
Samira Abnar and Willem Zuidema.
\newblock Quantifying attention flow in transformers.
\newblock In \emph{Proceedings of the 58th Annual Meeting of the Association for Computational Linguistics}, pages 4190--4197, 2020.

\bibitem[Adebayo et~al.(2018)Adebayo, Gilmer, Muelly, Goodfellow, Hardt, and Kim]{sanity_check}
Julius Adebayo, Justin Gilmer, Michael Muelly, Ian Goodfellow, Moritz Hardt, and Been Kim.
\newblock Sanity checks for saliency maps.
\newblock \emph{Advances in Neural Information Processing Systems}, 31, 2018.

\bibitem[Alemi et~al.(2017)Alemi, Fischer, Dillon, and Murphy]{ib_principle2}
Alexander~A Alemi, Ian Fischer, Joshua~V Dillon, and Kevin Murphy.
\newblock Deep variational information bottleneck.
\newblock In \emph{International Conference on Learning Representations}, 2017.

\bibitem[Ancona et~al.(2018)Ancona, Ceolini, {\"O}ztireli, and Gross]{sensitivity_n}
Marco Ancona, Enea Ceolini, Cengiz {\"O}ztireli, and Markus Gross.
\newblock Towards better understanding of gradient-based attribution methods for deep neural networks.
\newblock In \emph{International Conference on Learning Representations}, 2018.

\bibitem[Bao et~al.(2022)Bao, Dong, Piao, and Wei]{beit}
Hangbo Bao, Li Dong, Songhao Piao, and Furu Wei.
\newblock {BE}it: {BERT} pre-training of image transformers.
\newblock In \emph{International Conference on Learning Representations}, 2022.

\bibitem[Barkan et~al.(2023)Barkan, Asher, Eshel, Koenigstein, et~al.]{iia}
Oren Barkan, Yuval Asher, Amit Eshel, Noam Koenigstein, et~al.
\newblock Visual explanations via iterated integrated attributions.
\newblock In \emph{Proceedings of the IEEE/CVF International Conference on Computer Vision}, pages 2073--2084, 2023.

\bibitem[Binder et~al.(2016)Binder, Montavon, Lapuschkin, M{\"u}ller, and Samek]{lrp}
Alexander Binder, Gr{\'e}goire Montavon, Sebastian Lapuschkin, Klaus-Robert M{\"u}ller, and Wojciech Samek.
\newblock Layer-wise relevance propagation for neural networks with local renormalization layers.
\newblock In \emph{Artificial Neural Networks and Machine Learning--ICANN 2016: 25th International Conference on Artificial Neural Networks, Barcelona, Spain, September 6-9, 2016, Proceedings, Part II 25}, pages 63--71. Springer, 2016.

\bibitem[Caron et~al.(2021)Caron, Touvron, Misra, J{\'e}gou, Mairal, Bojanowski, and Joulin]{dino}
Mathilde Caron, Hugo Touvron, Ishan Misra, Herv{\'e} J{\'e}gou, Julien Mairal, Piotr Bojanowski, and Armand Joulin.
\newblock Emerging properties in self-supervised vision transformers.
\newblock In \emph{Proceedings of the IEEE/CVF International Conference on Computer Vision}, pages 9650--9660, 2021.

\bibitem[Chefer et~al.(2021{\natexlab{a}})Chefer, Gur, and Wolf]{generic}
Hila Chefer, Shir Gur, and Lior Wolf.
\newblock Generic attention-model explainability for interpreting bi-modal and encoder-decoder transformers.
\newblock In \emph{Proceedings of the IEEE/CVF International Conference on Computer Vision}, pages 397--406, 2021{\natexlab{a}}.

\bibitem[Chefer et~al.(2021{\natexlab{b}})Chefer, Gur, and Wolf]{trans_attr}
Hila Chefer, Shir Gur, and Lior Wolf.
\newblock Transformer interpretability beyond attention visualization.
\newblock In \emph{Proceedings of the IEEE/CVF Conference on Computer Vision and Pattern Recognition}, pages 782--791, 2021{\natexlab{b}}.

\bibitem[Chen et~al.(2022)Chen, Li, Yu, Dou, and Xiong]{beyond}
Jiamin Chen, Xuhong Li, Lei Yu, Dejing Dou, and Haoyi Xiong.
\newblock Beyond intuition: Rethinking token attributions inside transformers.
\newblock \emph{Transactions on Machine Learning Research}, 2022.

\bibitem[Deng et~al.(2009)Deng, Dong, Socher, Li, Li, and Fei-Fei]{imagenet}
Jia Deng, Wei Dong, Richard Socher, Li-Jia Li, Kai Li, and Li Fei-Fei.
\newblock Imagenet: A large-scale hierarchical image database.
\newblock In \emph{CVPR}, pages 248--255. IEEE, 2009.

\bibitem[Doshi-Velez and Kim(2017)]{whyneed1}
Finale Doshi-Velez and Been Kim.
\newblock Towards a rigorous science of interpretable machine learning.
\newblock \emph{arXiv preprint arXiv:1702.08608}, 2017.

\bibitem[Dosovitskiy et~al.(2020)Dosovitskiy, Beyer, Kolesnikov, Weissenborn, Zhai, Unterthiner, Dehghani, Minderer, Heigold, Gelly, et~al.]{vit_orig}
Alexey Dosovitskiy, Lucas Beyer, Alexander Kolesnikov, Dirk Weissenborn, Xiaohua Zhai, Thomas Unterthiner, Mostafa Dehghani, Matthias Minderer, G Heigold, S Gelly, et~al.
\newblock An image is worth 16x16 words: Transformers for image recognition at scale.
\newblock In \emph{International Conference on Learning Representations}, 2020.

\bibitem[Fang et~al.(2023)Fang, Wang, Xie, Sun, Wu, Wang, Huang, Wang, and Cao]{eva}
Yuxin Fang, Wen Wang, Binhui Xie, Quan Sun, Ledell Wu, Xinggang Wang, Tiejun Huang, Xinlong Wang, and Yue Cao.
\newblock Eva: Exploring the limits of masked visual representation learning at scale.
\newblock In \emph{Proceedings of the IEEE/CVF Conference on Computer Vision and Pattern Recognition}, pages 19358--19369, 2023.

\bibitem[He et~al.(2022)He, Chen, Xie, Li, Doll{\'a}r, and Girshick]{mae}
Kaiming He, Xinlei Chen, Saining Xie, Yanghao Li, Piotr Doll{\'a}r, and Ross Girshick.
\newblock Masked autoencoders are scalable vision learners.
\newblock In \emph{Proceedings of the IEEE/CVF Conference on Computer Vision and Pattern Recognition}, pages 16000--16009, 2022.

\bibitem[Hendrycks et~al.(2021{\natexlab{a}})Hendrycks, Basart, Mu, Kadavath, Wang, Dorundo, Desai, Zhu, Parajuli, Guo, et~al.]{imagenet_r}
Dan Hendrycks, Steven Basart, Norman Mu, Saurav Kadavath, Frank Wang, Evan Dorundo, Rahul Desai, Tyler Zhu, Samyak Parajuli, Mike Guo, et~al.
\newblock The many faces of robustness: A critical analysis of out-of-distribution generalization.
\newblock In \emph{Proceedings of the IEEE/CVF International Conference on Computer Vision}, pages 8340--8349, 2021{\natexlab{a}}.

\bibitem[Hendrycks et~al.(2021{\natexlab{b}})Hendrycks, Zhao, Basart, Steinhardt, and Song]{imagenet_a}
Dan Hendrycks, Kevin Zhao, Steven Basart, Jacob Steinhardt, and Dawn Song.
\newblock Natural adversarial examples.
\newblock In \emph{Proceedings of the IEEE/CVF Conference on Computer Vision and Pattern Recognition}, pages 15262--15271, 2021{\natexlab{b}}.

\bibitem[Hesse et~al.(2023)Hesse, Schaub-Meyer, and Roth]{funnybirds}
Robin Hesse, Simone Schaub-Meyer, and Stefan Roth.
\newblock Funnybirds: A synthetic vision dataset for a part-based analysis of explainable ai methods.
\newblock In \emph{Proceedings of the IEEE/CVF International Conference on Computer Vision}, pages 3981--3991, 2023.

\bibitem[Hila~Chefer and Wolf(2022)]{robust}
Idan~Schwartz Hila~Chefer and Lior Wolf.
\newblock Optimizing relevance maps of vision transformers improves robustness.
\newblock In \emph{Advances in Neural Information Processing Systems}, 2022.

\bibitem[Hooker et~al.(2019)Hooker, Erhan, Kindermans, and Kim]{roar}
Sara Hooker, Dumitru Erhan, Pieter-Jan Kindermans, and Been Kim.
\newblock A benchmark for interpretability methods in deep neural networks.
\newblock \emph{Advances in Neural Information Processing Systems}, 32, 2019.

\bibitem[Kingma and Ba(2014)]{adam}
Diederik~P Kingma and Jimmy Ba.
\newblock Adam: A method for stochastic optimization.
\newblock \emph{arXiv preprint arXiv:1412.6980}, 2014.

\bibitem[Kingma and Welling(2013)]{vae}
Diederik~P Kingma and Max Welling.
\newblock Auto-encoding variational bayes.
\newblock \emph{arXiv preprint arXiv:1312.6114}, 2013.

\bibitem[Kornblith et~al.(2019)Kornblith, Norouzi, Lee, and Hinton]{cka}
Simon Kornblith, Mohammad Norouzi, Honglak Lee, and Geoffrey Hinton.
\newblock Similarity of neural network representations revisited.
\newblock In \emph{International Conference on Machine Learning}, pages 3519--3529. PMLR, 2019.

\bibitem[Liu et~al.(2021)Liu, Lin, Cao, Hu, Wei, Zhang, Lin, and Guo]{swin1}
Ze Liu, Yutong Lin, Yue Cao, Han Hu, Yixuan Wei, Zheng Zhang, Stephen Lin, and Baining Guo.
\newblock Swin transformer: Hierarchical vision transformer using shifted windows.
\newblock In \emph{Proceedings of the IEEE/CVF International Conference on Computer Vision}, pages 10012--10022, 2021.

\bibitem[Liu et~al.(2022)Liu, Hu, Lin, Yao, Xie, Wei, Ning, Cao, Zhang, Dong, et~al.]{swin2}
Ze Liu, Han Hu, Yutong Lin, Zhuliang Yao, Zhenda Xie, Yixuan Wei, Jia Ning, Yue Cao, Zheng Zhang, Li Dong, et~al.
\newblock Swin transformer v2: Scaling up capacity and resolution.
\newblock In \emph{Proceedings of the IEEE/CVF Conference on Computer Vision and Pattern Recognition}, pages 12009--12019, 2022.

\bibitem[Otte(2013)]{whyneed3}
Clemens Otte.
\newblock Safe and interpretable machine learning: a methodological review.
\newblock \emph{Computational Intelligence in Intelligent Data Analysis}, pages 111--122, 2013.

\bibitem[Petsiuk et~al.(2018)Petsiuk, Das, and Saenko]{rise}
Vitali Petsiuk, Abir Das, and Kate Saenko.
\newblock Rise: Randomized input sampling for explanation of black-box models.
\newblock \emph{British Machine Vision Conference}, 2018.

\bibitem[Radford et~al.(2021)Radford, Kim, Hallacy, Ramesh, Goh, Agarwal, Sastry, Askell, Mishkin, Clark, et~al.]{clip}
Alec Radford, Jong~Wook Kim, Chris Hallacy, Aditya Ramesh, Gabriel Goh, Sandhini Agarwal, Girish Sastry, Amanda Askell, Pamela Mishkin, Jack Clark, et~al.
\newblock Learning transferable visual models from natural language supervision.
\newblock In \emph{International Conference on Machine Learning}, pages 8748--8763. PMLR, 2021.

\bibitem[Rong et~al.(2022)Rong, Leemann, Borisov, Kasneci, and Kasneci]{road}
Yao Rong, Tobias Leemann, Vadim Borisov, Gjergji Kasneci, and Enkelejda Kasneci.
\newblock A consistent and efficient evaluation strategy for attribution methods.
\newblock In \emph{International Conference on Machine Learning}, pages 18770--18795. PMLR, 2022.

\bibitem[Samek and M{\"u}ller(2019)]{whyneed2}
Wojciech Samek and Klaus-Robert M{\"u}ller.
\newblock Towards explainable artificial intelligence.
\newblock \emph{Explainable AI: interpreting, explaining and visualizing deep learning}, pages 5--22, 2019.

\bibitem[Schulz et~al.(2020)Schulz, Sixt, Tombari, and Landgraf]{iba}
Karl Schulz, Leon Sixt, Federico Tombari, and Tim Landgraf.
\newblock Restricting the flow: Information bottlenecks for attribution.
\newblock In \emph{International Conference on Learning Representations}, 2020.

\bibitem[Selvaraju et~al.(2017)Selvaraju, Cogswell, Das, Vedantam, Parikh, and Batra]{gradcam}
Ramprasaath~R Selvaraju, Michael Cogswell, Abhishek Das, Ramakrishna Vedantam, Devi Parikh, and Dhruv Batra.
\newblock Grad-cam: Visual explanations from deep networks via gradient-based localization.
\newblock In \emph{Proceedings of the IEEE/CVF International Conference on Computer Vision}, pages 618--626, 2017.

\bibitem[Steiner et~al.(2024)Steiner, Kolesnikov, Zhai, Wightman, Uszkoreit, and Beyer]{vit}
Andreas~Peter Steiner, Alexander Kolesnikov, Xiaohua Zhai, Ross Wightman, Jakob Uszkoreit, and Lucas Beyer.
\newblock How to train your vit? data, augmentation, and regularization in vision transformers.
\newblock \emph{Transactions on Machine Learning Research}, 2024.

\bibitem[Sundararajan et~al.(2017)Sundararajan, Taly, and Yan]{integrated}
Mukund Sundararajan, Ankur Taly, and Qiqi Yan.
\newblock Axiomatic attribution for deep networks.
\newblock In \emph{International Conference on Machine Learning}, pages 3319--3328. PMLR, 2017.

\bibitem[Tishby et~al.(2000)Tishby, Pereira, and Bialek]{ib_principle1}
Naftali Tishby, Fernando~C Pereira, and William Bialek.
\newblock The information bottleneck method.
\newblock \emph{arXiv preprint physics/0004057}, 2000.

\bibitem[Touvron et~al.(2021)Touvron, Cord, Douze, Massa, Sablayrolles, and J{\'e}gou]{deit}
Hugo Touvron, Matthieu Cord, Matthijs Douze, Francisco Massa, Alexandre Sablayrolles, and Herv{\'e} J{\'e}gou.
\newblock Training data-efficient image transformers \& distillation through attention.
\newblock In \emph{International Conference on Machine Learning}, pages 10347--10357. PMLR, 2021.

\bibitem[Touvron et~al.(2022)Touvron, Cord, and J{\'e}gou]{deit3}
Hugo Touvron, Matthieu Cord, and Herv{\'e} J{\'e}gou.
\newblock Deit iii: Revenge of the vit.
\newblock In \emph{Proceedings of the European Conference on Computer Vision}, pages 516--533. Springer, 2022.

\bibitem[Wang et~al.(2004)Wang, Bovik, Sheikh, and Simoncelli]{ssim}
Zhou Wang, Alan~C Bovik, Hamid~R Sheikh, and Eero~P Simoncelli.
\newblock Image quality assessment: from error visibility to structural similarity.
\newblock \emph{IEEE Transactions on Image Processing}, 13\penalty0 (4):\penalty0 600--612, 2004.

\bibitem[Xie et~al.(2023)Xie, Li, Cao, and Zhang]{vit-cx}
Weiyan Xie, Xiao-Hui Li, Caleb~Chen Cao, and Nevin~L Zhang.
\newblock Vit-cx: causal explanation of vision transformers.
\newblock In \emph{Proceedings of the Thirty-Second International Joint Conference on Artificial Intelligence}, pages 1569--1577, 2023.

\bibitem[Zhai et~al.(2023)Zhai, Mustafa, Kolesnikov, and Beyer]{siglip}
Xiaohua Zhai, Basil Mustafa, Alexander Kolesnikov, and Lucas Beyer.
\newblock Sigmoid loss for language image pre-training.
\newblock In \emph{Proceedings of the IEEE/CVF International Conference on Computer Vision}, pages 11975--11986, 2023.

\bibitem[Zhang et~al.(2021)Zhang, Khakzar, Li, Farshad, Kim, and Navab]{inputiba}
Yang Zhang, Ashkan Khakzar, Yawei Li, Azade Farshad, Seong~Tae Kim, and Nassir Navab.
\newblock Fine-grained neural network explanation by identifying input features with predictive information.
\newblock \emph{Advances in Neural Information Processing Systems}, 34:\penalty0 20040--20051, 2021.

\end{thebibliography}
}

% Uncomment below if compile appendix
% \appendix
\clearpage
\renewcommand{\thesection}{\Alph{section}}
\renewcommand{\thefigure}{\Alph{figure}}
\renewcommand{\theequation}{\Alph{equation}}
\renewcommand{\thetable}{\Alph{table}}

\setcounter{page}{1}
\setcounter{table}{0}
\setcounter{figure}{0}
\setcounter{equation}{0}
\maketitlesupplementary

\setcounter{section}{0}

% \section{Rationale}
% \label{sec:rationale}
% % 
% Having the supplementary compiled together with the main paper means that:
% % 
% \begin{itemize}
% \item The supplementary can back-reference sections of the main paper, for example, we can refer to \cref{sec:intro};
% \item The main paper can forward reference sub-sections within the supplementary explicitly (e.g. referring to a particular experiment); 
% \item When submitted to arXiv, the supplementary will already included at the end of the paper.
% \end{itemize}
% % 
% To split the supplementary pages from the main paper, you can use \href{https://support.apple.com/en-ca/guide/preview/prvw11793/mac#:~:text=Delete%20a%20page%20from%20a,or%20choose%20Edit%20%3E%20Delete).}{Preview (on macOS)}, \href{https://www.adobe.com/acrobat/how-to/delete-pages-from-pdf.html#:~:text=Choose%20%E2%80%9CTools%E2%80%9D%20%3E%20%E2%80%9COrganize,or%20pages%20from%20the%20file.}{Adobe Acrobat} (on all OSs), as well as \href{https://superuser.com/questions/517986/is-it-possible-to-delete-some-pages-of-a-pdf-document}{command line tools}.

\section{Derivative of Eq.~\ref{eq:main1}} % Variational upper bound

\begin{equation}
    \begin{split}
    % \begin{multlined}
        & I[Z_l; Z_{l-1}] \\
        &= E_{Z_{l-1}}[D_{KL}[P(Z_l | Z_{l-1}) || P(Z_l)]] \\
        &= \int_{Z_{l-1}} p(z_{l-1}) \left( \int_{Z_l} p(z_l | z_{l-1}) \text{log} \frac{p(z_l|z_{l-1})}{p(z_l)} dZ_l \right) dZ_{l-1} \\
        &= \int_{Z_l} \int_{Z_{l-1}} p(z_l, z_{l-1}) \text{log} \frac{p(z_l|z_{l-1})}{q(z_l)} dZ_l dZ_{l-1} \\ & \qquad - \int_{Z_l} \int_{Z_{l-1}} p(z_l, z_{l-1})  \text{log} \frac{p(z_l)}{q(z_l)} \\
        &= \int_{Z_l} \int_{Z_{l-1}} p(z_l, z_{l-1}) \text{log} \frac{p(z_l|z_{l-1})}{q(z_l)} dZ_l dZ_{l-1} \\ 
        & \qquad - \int_{Z_l} \left( \int_{Z_{l-1}} p(z_{l-1} | z_l) \right) p(z_l) \text{log} \frac{p(z_l)}{q(z_l)} \\
        &= E_{Z_{l-1}}[D_{KL}[P(Z_l | Z_{l-1}) || Q(Z_l)]] \\
        & \qquad - D_{KL}[P(Z_l)|| Q(Z_l)] \\
        & \leq E_{Z_{l-1}}[D_{KL}[P(Z_l | Z_{l-1}) || Q(Z_l)]] \ \text{.}
    % \end{multlined}
    \end{split}
\end{equation}

\begin{table}[t!]
\centering
\begin{adjustbox}{width=0.44\textwidth}
\begin{tabular}{p{3.3cm}ccc}
\specialrule{2.0pt}{1pt}{1pt}
Model   & Base model & Large model & Search space \\ \hline
Start Layer         & 4            & 8             & \{0,1,2,...,23 \} \\
End layer            & 12            & 24             & \{1,2,3,...,24 \} \\
Iteration      & \multicolumn{2}{c}{10}             & \{1,5,10,20\} \\ 
Optimizer       & \multicolumn{2}{c}{Adam}      & \{SGD, Adam\} \\
Batch size         & \multicolumn{2}{c}{10}    &  \{1,5,10,20\} \\
Learning rate       &   \multicolumn{2}{c}{1}  &  \{0.1,0.5,1,10\} \\
Trade-off parameter $\beta$         & \multicolumn{2}{c}{10}             & \{0.1,1,10,100\} \\
\specialrule{2.0pt}{1pt}{1pt}
\end{tabular}
\end{adjustbox}
% \end{centering}
\caption{\textbf{Hyperparameters selected in \ours{}}. Except for the start and end layer index, we unify the hyperparameters among the base and large models.}
\label{tab:sup_hyperparams}
\end{table}

\section{Relationship between $I[Y; Z_L]$ and Cross-Entropy Loss} 
% Cross Entropy를 조절하는 것이 I[Y; Z<=L] 조정하는 것.
Computation of the cross entropy $H(Y; \hat{Y})$ between the label $Y$ and predicted label $\hat{Y}$ can be considered as the conditional cross entropy $H(Y; \hat{Y}|Z_L)$ because $ Z_L$ determines the prediction.
Here, ViT utilizes the imputed representation $Z_L$ to predict the label $\hat{Y}$.
Such that, the conditional cross entropy can be divided into conditional entropy and KL divergence:
\begin{equation}
\begin{split}
    H(Y; \hat{Y}|Z_L) &= H(Y|Z_L) - H(Y) + D_{KL}[Y|Z_L || \hat{Y}|Z_L] \\
     &= H(Y|Z_L) + D_{KL}[Y || \hat{Y}|Z_L] \ \text{.}
\end{split}    
\end{equation}
Since mutual information between the bottleneck variable of $L$-th layer $Z_L$ and the label $Y$ can be expressed as:
\begin{equation}
    I[Y; Z_L] = H(Y) - H(Y|Z_L),
\end{equation}
where $H(Y)$ and $H(Y|Z_L)$ denote the entropy of $Y$ and conditional entropy of $Y$ conditioned on $Z_L$.
Thus, we can relate mutual information and conditional cross entropy as:
\begin{table}[t!]
\centering
\footnotesize
% \begin{centering}
% \renewcommand{\arraystretch}{1.0} 
% \begin{adjustbox}{width=.475\textwidth}
% \begin{tabular}{p{1.4cm}  p{1.4cm}  p{1.25 cm}  p{1.25 cm}  p{1.25 cm}  p{1.25 cm}}
\resizebox{\linewidth}{!}{
\begin{tabular}{cccccccc}
     \specialrule{2.0pt}{1pt}{1pt} % (all/diff)
     \noalign{\vspace{1.5pt}}
     % \noalign{\smallskip}
    % \multirow{2}{*}{Model}&  & \multicolumn{2}{c}{\underline{\hspace{2.5em}Difficult\hspace{2.5em}}} & \multicolumn{2}{c}{\underline{\hspace{3.5em}Easy\hspace{3.5em}}} \\
    \multirow{2.3}{*}{Model}&\multirow{2.3}{*}{Setting}& & \multicolumn{2}{c}{Low-confident} & & \multicolumn{2}{c}{High-confident} \\
    % \noalign{\smallskip}
    \noalign{\vspace{1.5pt}}
    \cline{4-5}
    \cline{7-8}
    \noalign{\vspace{1.5pt}}
    & & & \begin{tabular}[c]{c@{}}0-20\end{tabular} & 20-40 & & \begin{tabular}[c]{c@{}}60-80\end{tabular} & 80-100 \\
    \midrule 
\multirow{4}{*}{ViT-B ($s$)} & 2-12 & & 1.20/23.86 & 2.57/35.02 & & \hspace{4pt}8.63/56.98 & 17.30/75.66 \\
& \cellcolor{shadecolor}4-12 & \cellcolor{shadecolor} & \cellcolor{shadecolor}1.27/24.40 & \cellcolor{shadecolor}2.66/33.17 & \cellcolor{shadecolor} & \cellcolor{shadecolor}\hspace{4pt}8.70/55.48 & \cellcolor{shadecolor}16.83/75.64 \\
& 6-12 & & 1.11/21.45 & 2.83/31.99 & & \hspace{4pt}9.61/54.62 & 18.58/75.24 \\
& 8-12 & & 1.53/15.44 & 3.48/29.37 & & 10.58/51.42 & 22.37/72.65 \\ \midrule
\multirow{4}{*}{DeiT-B ($s$)} & 2-12 & & 0.67/22.08 & 2.44/33.16 & & \hspace{4pt}9.82/54.42 & 17.54/66.12 \\
& \cellcolor{shadecolor}4-12 &\cellcolor{shadecolor}& \cellcolor{shadecolor}0.62/19.90 & \cellcolor{shadecolor}2.33/34.34 & \cellcolor{shadecolor} & \cellcolor{shadecolor}\hspace{4pt}9.49/54.49 & \cellcolor{shadecolor}16.63/66.29 \\
& 6-12 & & 0.67/18.82 & 2.57/32.14 & & \hspace{4pt}9.14/54.00 & 15.62/66.02 \\
& 8-12 & & 0.79/17.47 & 3.23/30.26 & & 10.37/52.13 & 17.42/64.38 \\ \midrule
\multirow{4}{*}{ViT$^\dagger$-B ($s$)} & 2-12 & & 1.30/35.20 & 4.17/45.11 & & 12.41/64.28 & 22.70/75.36 \\
& \cellcolor{shadecolor}4-12 &\cellcolor{shadecolor}& \cellcolor{shadecolor}1.22/31.93 & \cellcolor{shadecolor}3.53/42.82 & \cellcolor{shadecolor} & \cellcolor{shadecolor}11.94/64.42 & \cellcolor{shadecolor}20.58/75.56 \\
& 6-12 & & 1.23/32.07 & 3.69/41.54 & & 12.20/62.82 & 21.37/74.93 \\
& 8-12 & & 1.50/27.70 & 4.30/36.59 & & 13.79/60.44 & 23.49/73.12 \\
\midrule
\multirow{4}{*}{ViT-B ($e$)} & 4-6 & & 4.17/13.04 & 3.92/25.56 & & 13.12/47.93 & 27.88/62.34 \\
& 4-8 & & 2.85/11.94 & 3.14/28.94 & & 10.09/52.79 & 19.18/74.24 \\
& 4-10 & & 2.81/17.34 & 2.85/29.77 & & \hspace{4pt}9.04/54.17 & 16.86/75.66 \\
& \cellcolor{shadecolor}4-12 &\cellcolor{shadecolor}& \cellcolor{shadecolor}1.27/24.40 & \cellcolor{shadecolor}2.66/33.17 & \cellcolor{shadecolor} & \cellcolor{shadecolor}\hspace{4pt}8.70/55.48 & \cellcolor{shadecolor}16.83/75.64 \\ \midrule
\multirow{4}{*}{DeiT-B ($e$)}& 4-6 & & 1.49/14.19 & 5.66/25.92 & & 18.60/44.53 & 36.85/50.18 \\
& 4-8 & & 1.30/18.34 & 3.76/18.34 & & 14.12/40.51 & 25.10/50.75 \\
& 4-10 & & 0.86/20.92 & 3.32/33.47 & & 11.49/52.65 & 19.40/64.56 \\
& \cellcolor{shadecolor}4-12 &\cellcolor{shadecolor}& \cellcolor{shadecolor}0.62/19.90 & \cellcolor{shadecolor}2.33/34.34 & \cellcolor{shadecolor} & \cellcolor{shadecolor}\hspace{4pt}9.49/54.49 & \cellcolor{shadecolor}16.32/66.29 \\   \midrule
\multirow{4}{*}{ViT$^\dagger$-B ($e$)}& 4-6 & & 2.89/17.21 & 7.27/34.18 & & 22.30/55.61 & 40.42/63.98 \\
& 4-8 & & 2.14/25.92 & 5.27/39.92 & & 15.11/60.99 & 25.95/73.52 \\
& 4-10 & & 1.21/29.02 & 4.17/41.50 & & 12.87/63.30 & 22.19/75.31 \\
& \cellcolor{shadecolor}4-12 &\cellcolor{shadecolor}& \cellcolor{shadecolor}1.22/31.93 & \cellcolor{shadecolor}3.53/42.82 & \cellcolor{shadecolor} & \cellcolor{shadecolor}11.94/64.42 & \cellcolor{shadecolor}20.58/75.56 \\  
    \specialrule{2.0pt}{1pt}{1pt}
\end{tabular}
}
% \end{adjustbox}
% \end{centering}
\vspace{-8pt}
\caption{\textbf{Ablation study on departure ($s$) and arrival ($e$) layers.}
% The ablation study of interpolating the departure ($s$) and arrival ($e$) layers from start to end.
The comparisons on interpolating the hyper-parameters from start to end layers.
The hyper-parameters used in \ours{} are filled with a light gray color.
We compare the quantitative results of the discrepancy between insertion/deletion scores with different intervals of confidence scores yielded by the model.
The better-qualified attribution map yields a higher discrepancy in insertion/deletion.
}
\vspace{-10pt}
\label{tab:hyperparams1}
\end{table}

\begin{table}
\centering
% \subfloat[Variants of patch size and depth]{
% \begin{centering}
% \renewcommand{\arraystretch}{1.0} 
% \begin{adjustbox}{width=.475\textwidth}
% \begin{tabular}{p{1.4cm}  p{1.4cm}  p{1.25 cm}  p{1.25 cm}  p{1.25 cm}  p{1.25 cm}}
\resizebox{0.9\linewidth}{!}{
\begin{tabular}{ccccc}
     \specialrule{2.0pt}{1pt}{1pt} % (all/diff)
     \noalign{\vspace{1.5pt}}
     % \noalign{\smallskip}
    % \multirow{2}{*}{Model}&  & \multicolumn{2}{c}{\underline{\hspace{2.5em}Difficult\hspace{2.5em}}} & \multicolumn{2}{c}{\underline{\hspace{3.5em}Easy\hspace{3.5em}}} \\
    % \multirow{2.3}{*}{Model}& & \multicolumn{2}{c}{Patch size} & & \multicolumn{2}{c}{Depth} \\
    Model & & \textit{SA} & \textit{FFN} & \textit{Blocks} \\
    % \noalign{\smallskip}
    % \noalign{\vspace{1.5pt}}
    % \noalign{\vspace{1.5pt}}
    % & & \begin{tabular}[c]{c@{}}$8$\end{tabular} & $32$ & & \begin{tabular}[c]{c@{}}ViT-L\end{tabular} & DeiT3-L \\
    \midrule 
ViT-B-16/224 & & 12.81/62.47 & 15.71/57.56 & 17.53/56.31 \\
DeiT-B-16/224 & & 11.59/53.86 & 13.41/51.12 & 13.53/52.39 \\
ViT$^\dagger$-B-16/224 & & 15.90/64.87 & 17.74/62.07 & 19.52/61.97 \\
    \specialrule{2.0pt}{1pt}{1pt}
\end{tabular}
}
% }
%%%%%%%
% \end{adjustbox}
% \end{centering}
\vspace{-8pt}
\caption{\textbf{Quantitative comparison of the results produced by placing bottleneck into various operations.} 
This experiment shows the correctness of the performance when inserting bottlenecks into various operations, including self-attention (SA), feed-forward network (FFN), and block between SA and FFN.
We compare 6,000 images randomly sampled from the IN-1k validation dataset.
ViT$^\dagger$ denotes the model trained with CLIP.
}
\vspace{-10pt}
\label{tab:sub_ablation_op}
\end{table}

\begin{table*}[th!]
\centering
\footnotesize
% \begin{subtable}[t!]{0.48\linewidth} %%%%%%
% \renewcommand{\arraystretch}{1.0} 
% \begin{adjustbox}{width=.475\textwidth}
% \begin{tabular}{p{1.4cm}  p{1.4cm}  p{1.25 cm}  p{1.25 cm}  p{1.25 cm}  p{1.25 cm}}
\resizebox{0.85\linewidth}{!}{
\begin{tabular}{ccccccccccccccc}
     \specialrule{2.0pt}{1pt}{1pt} % (all/diff)
     \noalign{\vspace{1.5pt}}
     % \noalign{\smallskip}
    % \multirow{2}{*}{Model}&  & \multicolumn{2}{c}{\underline{\hspace{2.5em}Difficult\hspace{2.5em}}} & \multicolumn{2}{c}{\underline{\hspace{3.5em}Easy\hspace{3.5em}}} \\
    \multirow{2.3}{*}{Model}& & \multirow{2.3}{*}{Accuracy} & & \multicolumn{4}{c}{Com.} & & \multirow{2.3}{*}{BI} & & \multicolumn{1}{c}{Cor.} & & \multicolumn{1}{c}{Con.} & \multirow{2.3}{*}{mX} \\
    % \noalign{\smallskip}
    \noalign{\vspace{1.5pt}}
    \cline{5-8}
    \cline{12-12}
    \cline{14-14}
    \noalign{\vspace{1.5pt}}
    & &  & & CSDC & PC & DC & D & &  & & SD & & TS & \\
    \midrule 
Chefer-LRP & & 97.6 & & 91.1 & 91.2 & 89.4 & 89.7 & & 99.8 & & 73.9 & & 95.8 & 86.6 \\
Generic & & 97.6 & & 91.0 & 90.8 & 89.6 & 89.6 & & 99.8 & & 74.2 & & 98.5 & 87.6  \\
IIA & & 97.6 & & 89.2 & 87.6 & 88.0 & 90.7 & & 99.8 & & 76.4 & & 98.6 & 84.1 \\
ViT-CX & & 97.6 & & 56.9 & 36.2 & 41.6 & 83.8 & & 99.8 & & 78.3 & & 57.7 & 66.8 \\
IBA & & 97.6 & & 96.0 & 97.8 & 94.4 & 91.8 & & 99.8 & & 76.9 & & 71.7 & 80.8 \\
Beyond & & 97.6 & & 87.8 & 84.8 & 84.8 & 84.1 & & 99.8 & & 75.8 & & 92.9 & 84.5 \\
\ours{} & & 97.6 & & 93.5 & 94.2 & 91.4 & 91.3 & & 99.8 & & 79.0 & & 98.2 & \textbf{89.8} \\
    \specialrule{2.0pt}{1pt}{1pt}
\end{tabular}
}
% }

%%%%%%%
% \end{adjustbox}
% \end{centering}
\vspace{-8pt}
\caption{\textbf{Numeric detailed results of FunnyBirds experiment.}
We provide the detailed numeric results for the reported FunnyBirds experiment.
The mean explainability score (mX) is obtained by averaging Com., Cor., and Con. scores.
The completeness score (Com.) is obtained by averaging CSDC, PC, DC, and D scores.
}
\vspace{-10pt}
\label{tab:funnybirds_full}
\end{table*}

\begin{equation}
\begin{split}
    I[Y; Z_L] &= H(Y) - H(Y|Z_L) \\
     &= H(Y) + D_{KL}[Y || \hat{Y}|Z_L] - H(Y; \hat{Y} | Z_L) \\
     & \geq -H(Y; \hat{Y} | Z_L)   \ \text{.}
\end{split}
\end{equation}
Here, we omit $H(Y)$ since it is constant.
As $D_{KL}[Y || \hat{Y}|Z_L] \geq 0$, minimization of the conditional cross entropy increases the mutual information.

\section{Experimental Settings}

\subsection{Model}
We detail the settings for the experiments.
We utilize the timm library, which is a publicly accessible open-source framework.
% The key for each model is illustrated in Tab~\ref{}.
We present all models by \{\textit{name}\}--\{\textit{depth}\}--\{\textit{patch size}\}/\{\textit{image resolution}\}, \textit{e.g.}, ViT-B-16/224.
All the included ViT models are pre-trained with ImageNet-21k.
The models belonging to the DeiT family are pre-trained with IN-1k, including DeiT3.
We leverage Swin-B with the settings of window size $7$ and patch size $4$.
For Swin2-B, we utilize the model with the settings including the input resolution of $256$ and window size $16$.
We denote the ViT$^*$ and ViT$^\dagger$ as the models trained with massive regression and CLIP, respectively.

\subsection{Hyperparameters Selected in \ours{}}
We provide quantitative comparisons against various hyperparameter settings.
\ours{} includes the departure $s$ and arrival $e$ layers and trade-off parameter $\beta$ as a hyperparameter to set.
The overall hyperparameter settings chosen in \ours{} are illustrated in Tab.~\ref{tab:sup_hyperparams}.
We discuss the setting of trade-off hyperparameter $\beta$ in Sec.~\ref{sec:beta_comparison} regarding the out-of-distribution problem.
We insert the bottleneck into the preceding operation of the self-attention layer \textit{i.e.}, normalization layer.

\noindent \textbf{Departure and Arrival Layers} We empirically select the hyperparameter, which broadly yields the best correctness performance.
Tab.~\ref{tab:hyperparams1} illustrates the quantitative comparisons against various hyperparameter settings.
As shown in the results, including earlier layers for ViT yields an increased performance in insertion/deletion.
However, our chosen hyperparameter shows enhanced performance in ViT pre-trained with CLIP and DeiT models including DeiT3.
Referring to these results, we set $4$ and $12$ as departure and arrival layers, respectively, for base models.
For large models, we set $8$ and $24$ as departure and arrival layers, respectively.

\noindent \textbf{Operation}
We compare the correctness scores by inserting bottlenecks into the three types of operations: self-attention (SA), feed-forward network (FFN), and intermediate blocks between SA and FFN (Block).
Tab.~\ref{tab:sub_ablation_op} shows the quantitative results for these settings.
As shown in the results, inserting the bottleneck before the SA layer yields the best performance compared to inserting it before other operations.

% \noindent \textbf{Departure Layer}
% The departure layer is the beginning position of the layers to insert the bottleneck.
% We compare the quantitative results of different departure layer settings with various model architectures.

\begin{table*}[!ht]
\centering
\begin{centering}

\resizebox{0.9\linewidth}{!}{
\begin{tabular}{lcccccccc}
\specialrule{2.0pt}{1pt}{1pt}
Variant                   & Model                                                        & Chefer-LRP & Generic & IIA & ViT-CX & IBA & Beyond & \ours{} \\ \hline
                        % & ViT-B-8/224$^*$ & - & 19.57/49.86 & 18.43/51.32 & 20.97/46.75 & 16.68/51.91 & 17.10/51.82 & \textbf{13.44}/\textbf{54.18} \\
                        % & ViT-B-32/224$^*$ & - & 17.54/55.25 & 16.11/56.35 & 16.04/58.09 & 15.51/58.84 & 15.15/58.51 & \textbf{13.73}/\textbf{62.02} \\ 
\multirow{3}{*}{SS*}    & ViT-B-16/224 (MAE) & - & 24.44/42.80 & 24.53/43.63 & 20.01/45.80 & \underline{15.66}/\underline{48.90} & 16.23/48.09 & \textbf{13.77}/\textbf{53.42} \\ 
                        & ViT-B-16/224 (Dino) & - & \underline{7.62}/50.42 & 7.46/\underline{50.52} & 14.09/45.55 & 8.91/49.45 & 8.12/50.23 & \textbf{6.83}/\textbf{53.68} \\
                        & BeiTv1-B-16/224 & - & 24.82/47.04 & 25.34/47.50 & 21.58/54.06 & \underline{13.96}/\underline{59.87} & 19.75/51.82 & \textbf{13.51}/\textbf{62.45} \\ \hline
\multirow{5}{*}{IN-A}    & ViT$^*$-B-16/224 & - & 2.46/25.99 & 2.68/25.37 & 
                         4.41/23.97 & 2.34/25.52 & \underline{2.27}/\underline{26.29} & \textbf{1.82/32.62} \\ 
                         & ViT$^*$-L-16/224 & - & 3.99/32.74 & 3.95/32.29 & 
                         4.40/33.64 & 3.19/34.86 & \underline{3.10}/\underline{35.36} & \textbf{2.53/41.07} \\ 
                         & ViT$^\dagger$-B-16/224 & - & 2.58/28.46 & \underline{2.44}/28.60 & 9.67/10.98 & 2.49/27.33 & 2.53/\underline{28.72} & \textbf{1.97/36.15} \\ 
                         & ViT$^\dagger$-L-16/224 & - & 4.19/37.82 & \underline{4.01}/\underline{38.13} & 7.49/34.45 & 4.27/35.77 & 4.16/38.12 & \textbf{3.25/42.74} \\  
                         & EVA-L-14/196 & - & 7.58/39.96 & 8.79/38.48 & 6.60/41.71 & \underline{4.50}/\underline{44.68} & 4.54/43.57 & \textbf{3.65/50.41} \\  \hline
\multirow{5}{*}{IN-R}    
                         & ViT$^*$-B-16/224 & - & 7.72/36.70 & 8.41/36.13 & 
                         9.18/34.34 & \underline{6.56}/\underline{37.85} & 6.79/37.58 & \textbf{5.14/43.56} \\ 
                         & ViT$^*$-L-16/224 & - &  13.21/38.11 & 13.44/37.32 & 
                         12.20/40.84 & \underline{8.66}/43.91 & 9.23/\underline{44.66} & \textbf{7.04/49.77} \\ 
                         & DeiT3-B-16/224 & - & 5.91/37.55 & 5.92/37.57 & 9.06/33.85 & \underline{5.41}/37.53 & 5.55/\underline{38.81} & \textbf{4.36/43.87} \\ 
                         & DeiT3-L-16/224 & - & 7.24/40.83 & 7.25/40.74 & 10.95/36.54 & \underline{6.46}/41.27 & 6.92/\underline{42.56} & \textbf{5.13}/\textbf{46.82} \\ 
                         & EVA-L-14/196 & - & 19.49/46.54 & 19.73/45.96 & 14.01/51.17 & \underline{11.36}/\underline{54.73} & 12.73/53.00 & \textbf{9.29}/\textbf{59.97} \\

\specialrule{2.0pt}{1pt}{1pt}
\end{tabular}
}
\vspace{-8pt}
\caption{\textbf{Quantitative feature importance assessment on insertion $\uparrow$ / deletion $\downarrow$.} 
We denote SS$^*$ as ViT trained with self-supervised learning~\cite{mae,dino,beit} and ViT$^\dagger$ as ViT trained with CLIP.
ViT$^\dagger$ and ViT$^*$ denote the ViTs trained with CLIP and massive regularization methods.
We additionally include EVA~\cite{eva} for the comparison.
We underline the state-of-the-art performance among the baselines.
}
\vspace{-12pt}
% \end{adjustbox}
\label{tab:sup_insdel}
\end{centering}
\end{table*}

% \noindent \textbf{Arrival Layer}

% \subsection{Trade-off Parameter}

% \subsection{Optimizer}

% \subsection{Overall Settings}

% \subsection{Layer selection for IBA}
% For the rigorous comparison, we evaluate correctness scores against various IBA settings.
% To measure the correctness scores, we utilize insertion/deletion as a metric for comparison.
% Table~\ref{} shows the insertion/deletion performance by interpolating a specific layer to insert bottleneck, from earlier to deeper layers.
% We insert IBA into the path forwarding all the features including shortcut features.
% This path is positioned between the feed-forward network (FFN) and the self-attention layer (SA), called Blocks.
% We compare the correctness scores against inserting bottleneck into FFN, SA, and Blocks in Tab.~\ref{}.
% As shown in the results, inserting IBA into Blocks results in the best correctness score.

\section{Additional Quantitative Results}
% Appendix 후보 (2)

\subsection{FunnyBirds Experiment}
FunnyBirds assessment measures the faithfulness of an attribution map with a comprehensive metric, including three metrics: completeness (Com.), correctness (Cor.), and contrastivity (Con.).
First, to assess Com., FunnyBirds assesses the controlled synthetic data check (CSDC), preservation check (PC), deletion check (DC), and distractibility (D).
These four metrics are averaged to compute the Com. score.
Second, evaluating Cor. includes a single deletion check (SD) which measures the correlation between part importance and the predicted scores of the targeted class.
Finally, Con. measures target sensitivity to directly measure the sensitivity to a target class by assessing whether parts of different classes are correctly identified as their respective class from a single image.
The overall score is obtained by averaging Com., Cor. and Con. scores.
% We detail the numeric results in this section.
Tab.~\ref{tab:funnybirds_full} shows the full numeric results of the FunnyBirds experiment.
As shown in the results, \ours{} 
In particular, the outperforming of \ours{} in terms of target sensitivity compared to IBA demonstrates the ability of \ours{} in class-discriminative ability.

\begin{figure*}[t!]
\centering

% \subfloat[]{
% \includegraphics[width=.182\linewidth]{Assets/road_result2_vit-b.pdf}}
% \subfloat[]{
% \includegraphics[width=.18\linewidth]{Assets/road_result2_deit-b.pdf}}
% \subfloat[]{
% \includegraphics[width=.18\linewidth]{Assets/road_result2_vit-l.pdf}}
% \subfloat[]{
% \includegraphics[width=.18\linewidth]{Assets/road_result2_deit3-l.pdf}}
% \subfloat[]{
% \includegraphics[width=.18\linewidth]{Assets/road_result2_vit-h.pdf}}

\subfloat[IN-1k]{
\includegraphics[width=.25\linewidth]{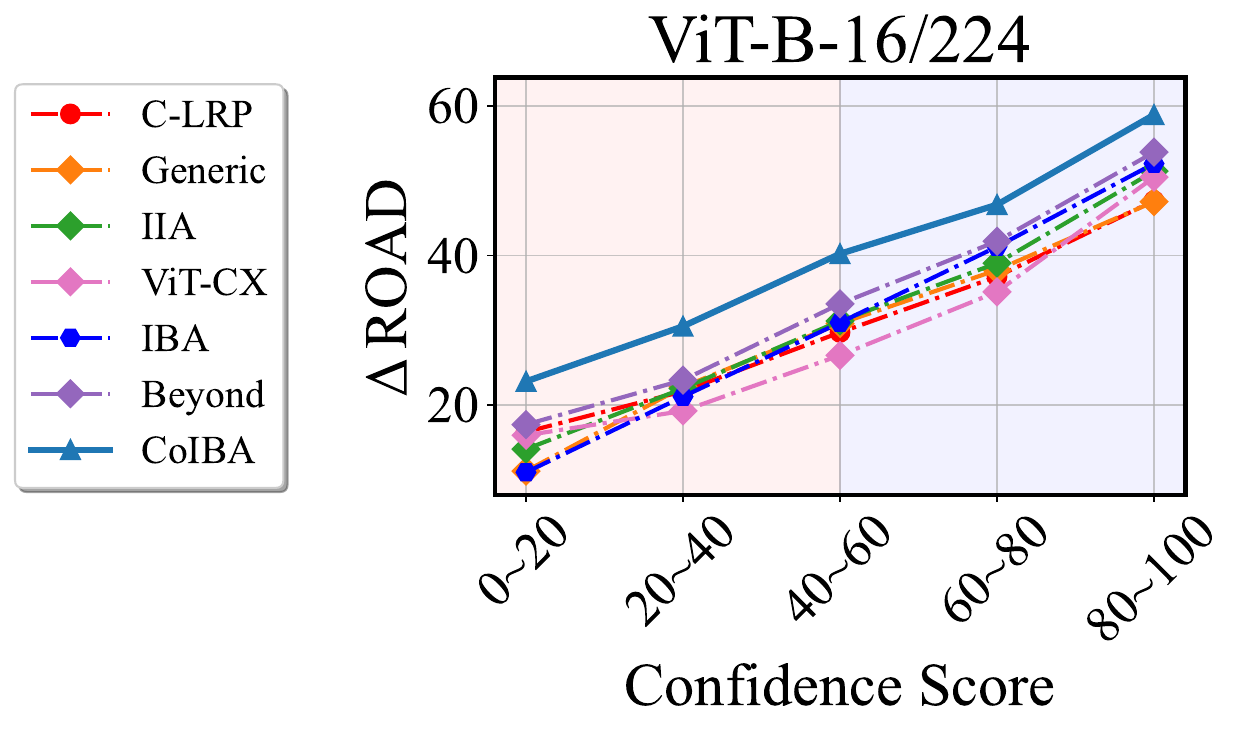}
\includegraphics[width=.18\linewidth]{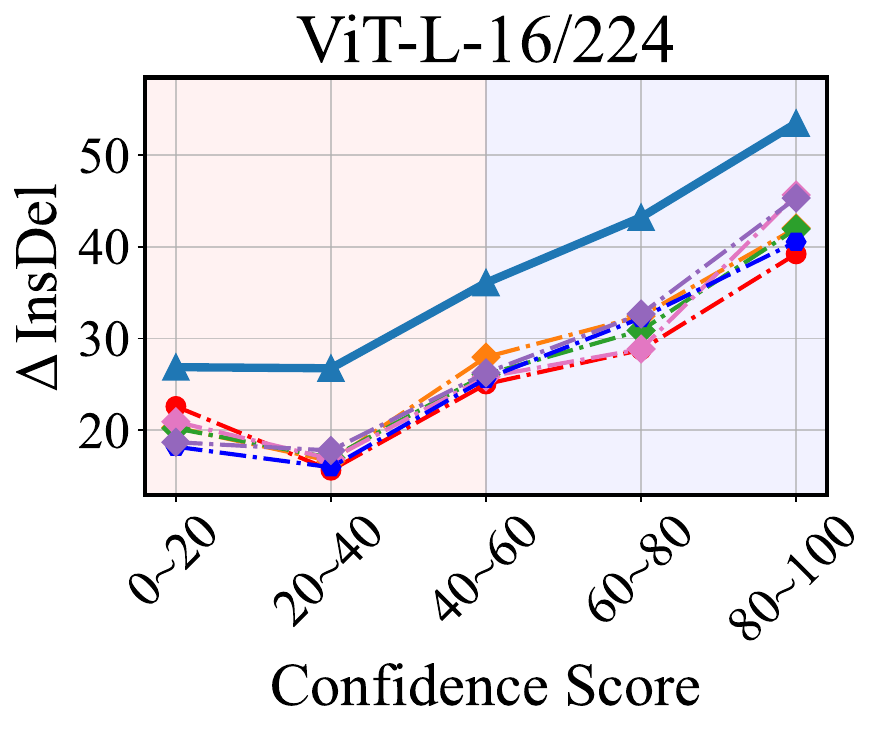}
\includegraphics[width=.18\linewidth]{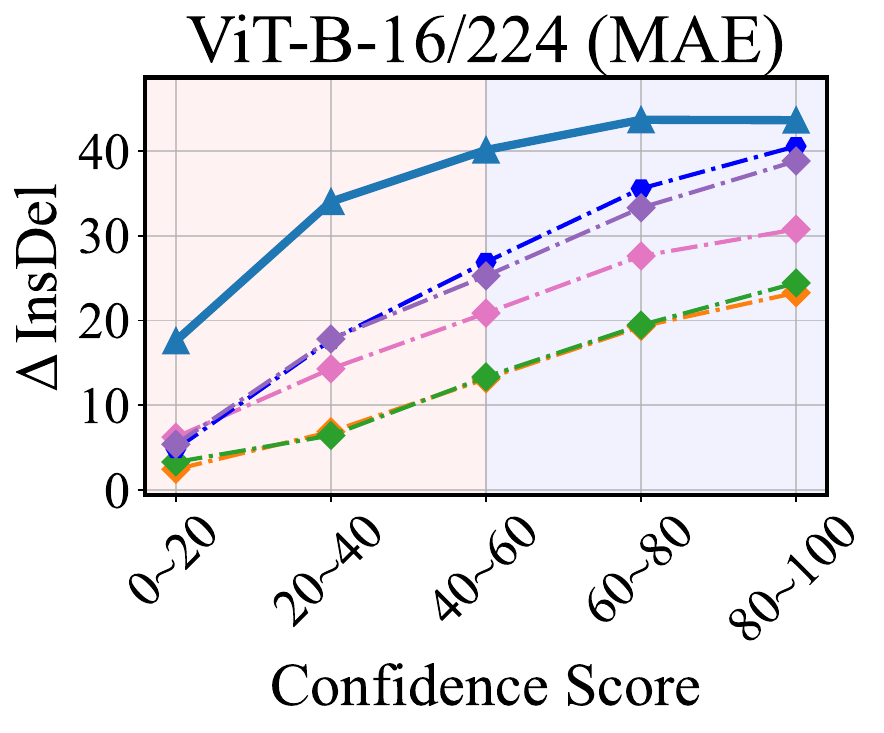}
\includegraphics[width=.18\linewidth]{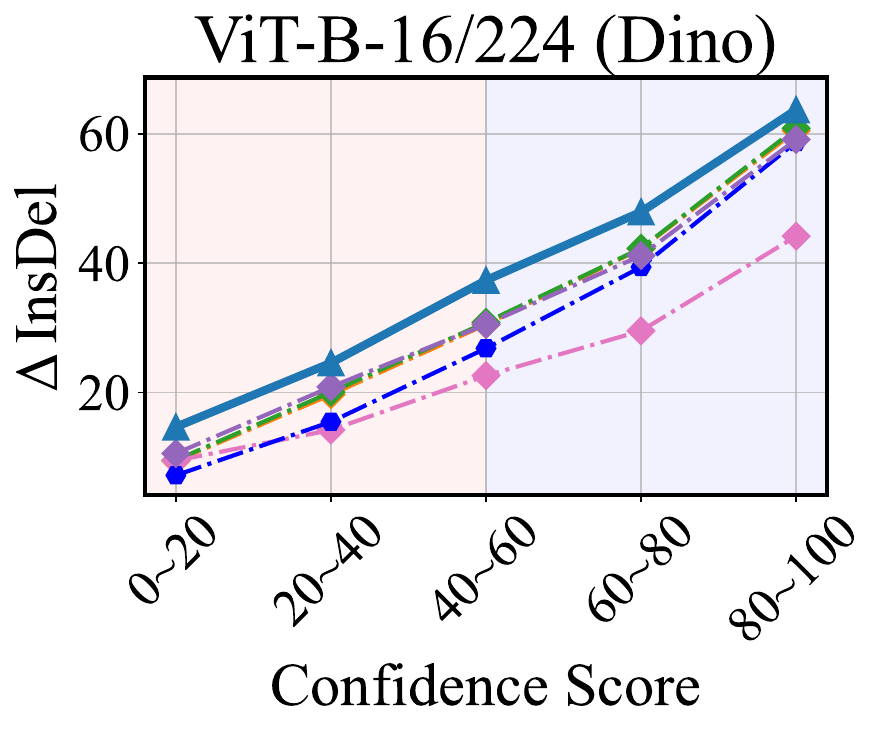}
\includegraphics[width=.18\linewidth]{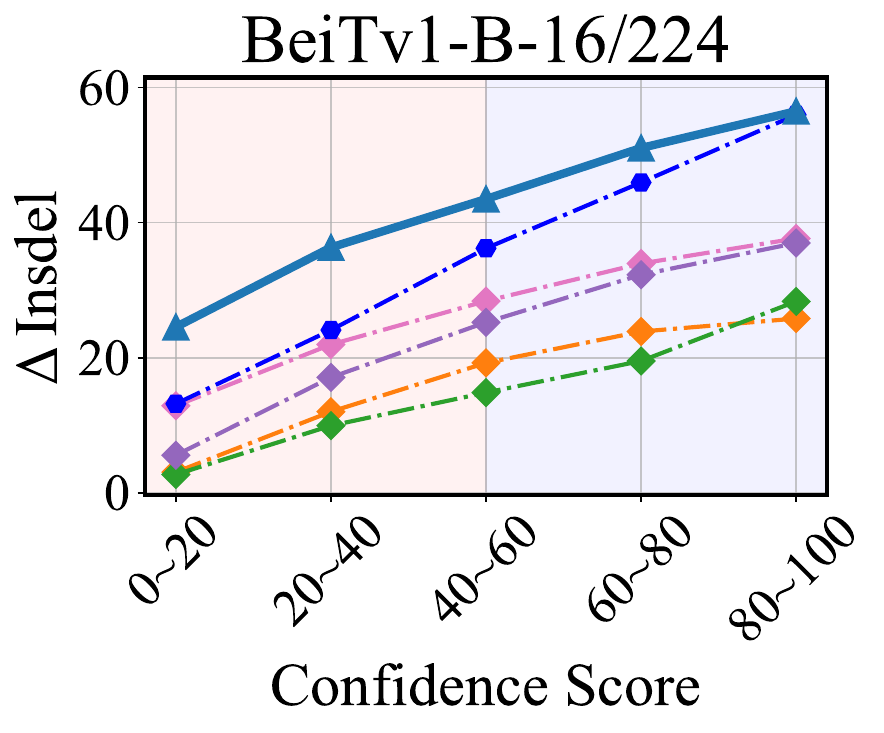}
\vspace{-2pt}
}

\hspace{0.06\linewidth}
\subfloat[IN-A]{
\includegraphics[width=.18\linewidth]{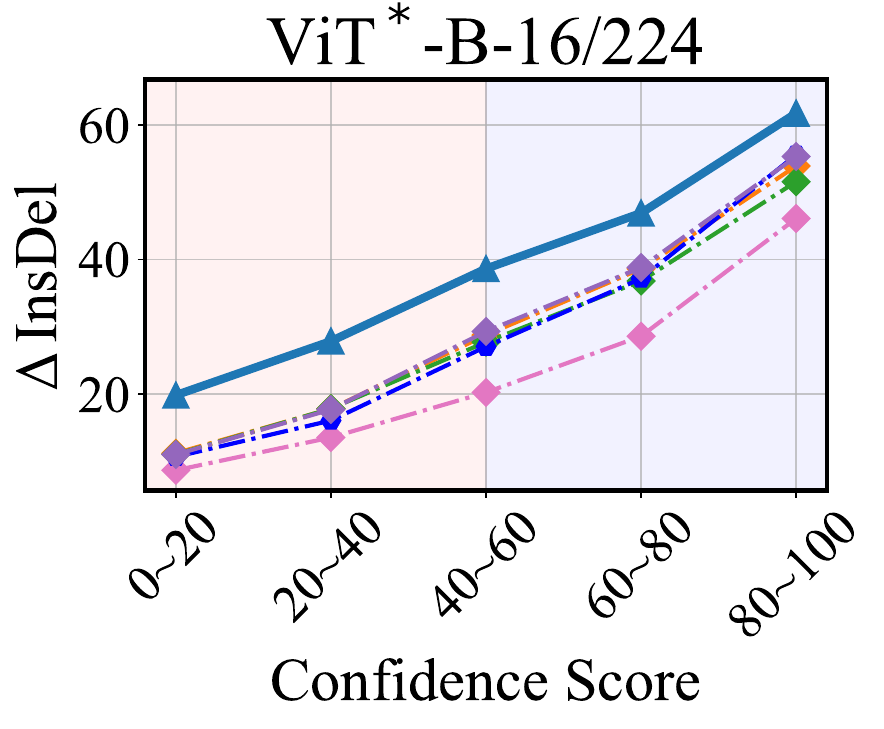}
\includegraphics[width=.18\linewidth]{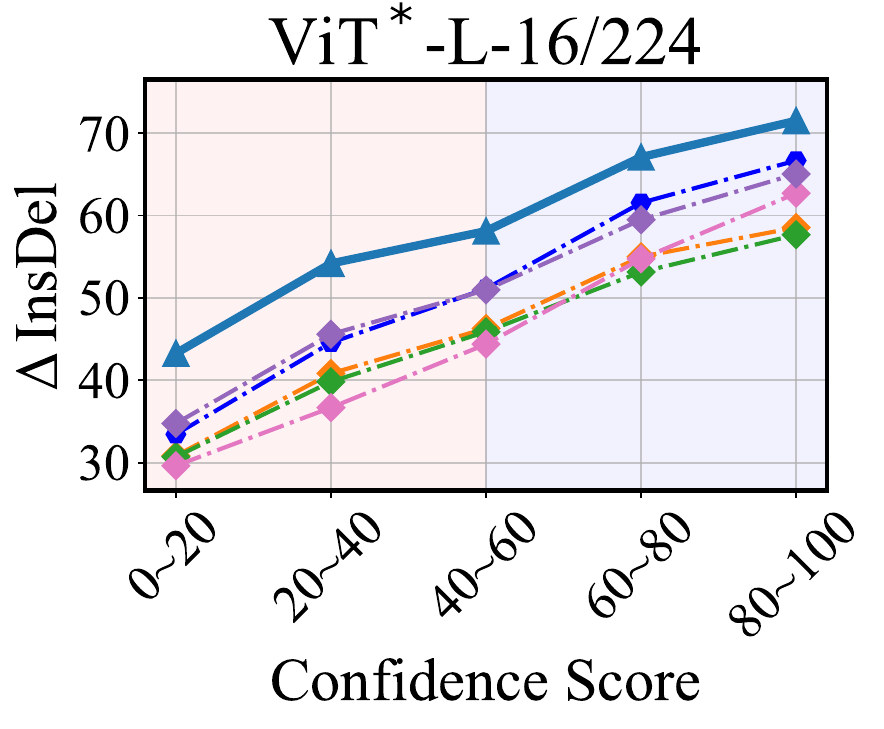}
\includegraphics[width=.18\linewidth]{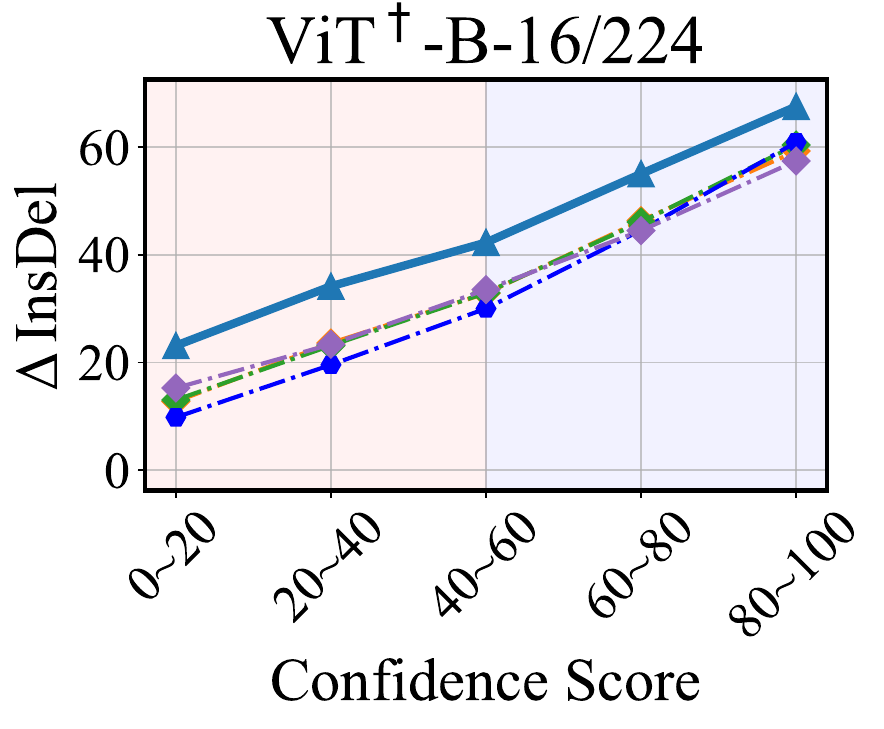}
\includegraphics[width=.18\linewidth]{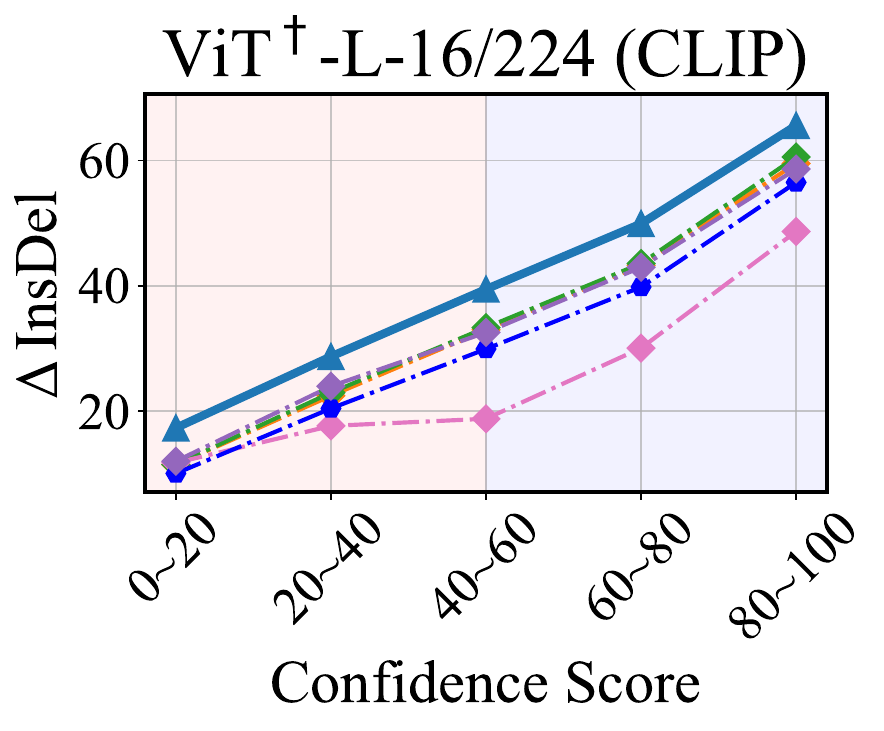}
\includegraphics[width=.18\linewidth]{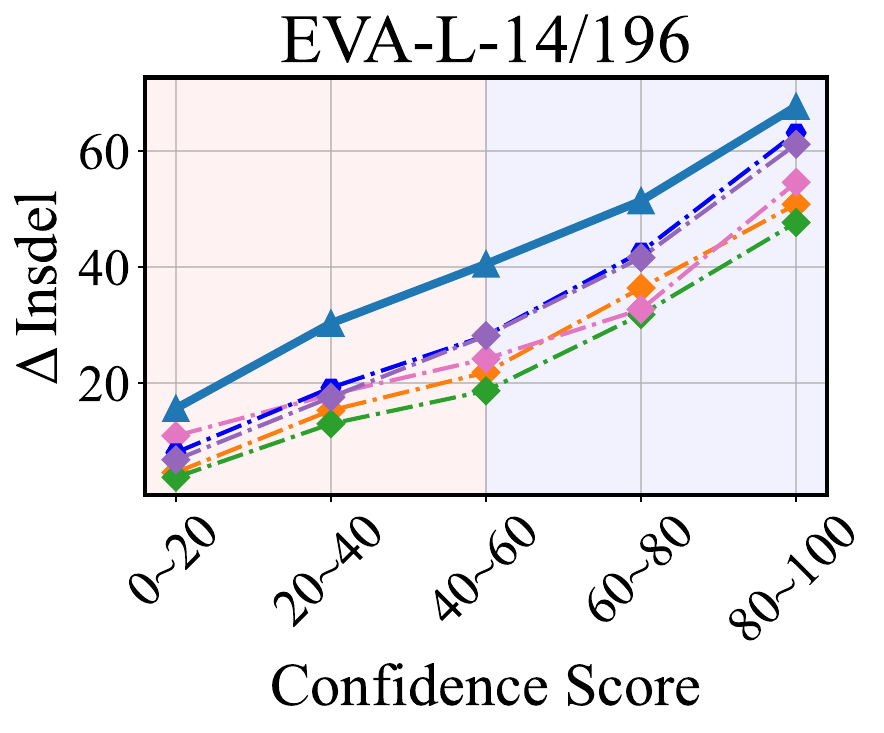}
\vspace{-2pt}
}

\hspace{0.06\linewidth}
\subfloat[IN-R]{
\includegraphics[width=.18\linewidth]{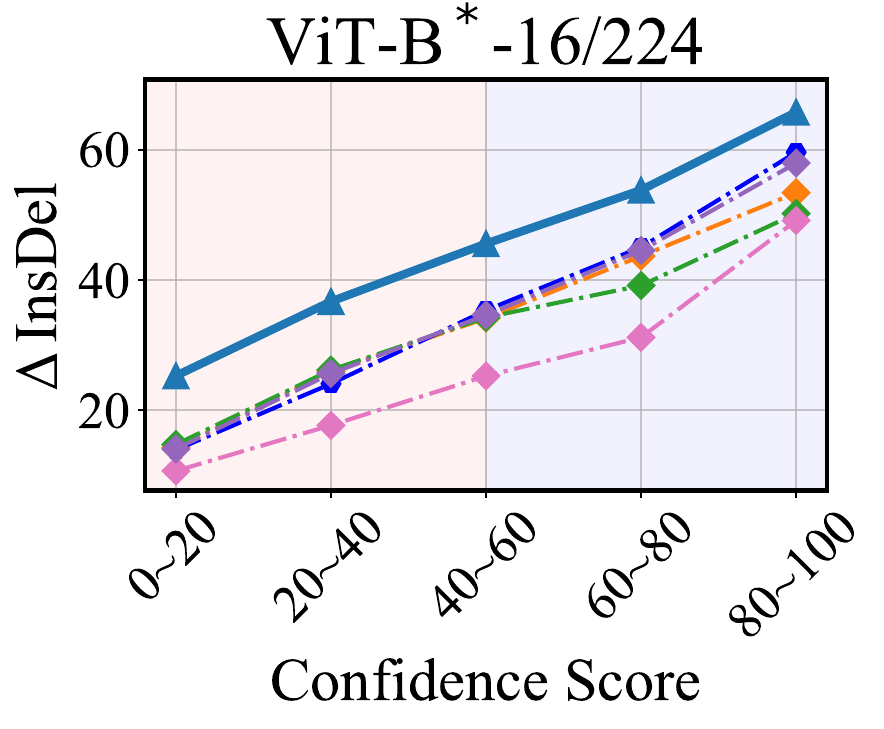}
\includegraphics[width=.18\linewidth]{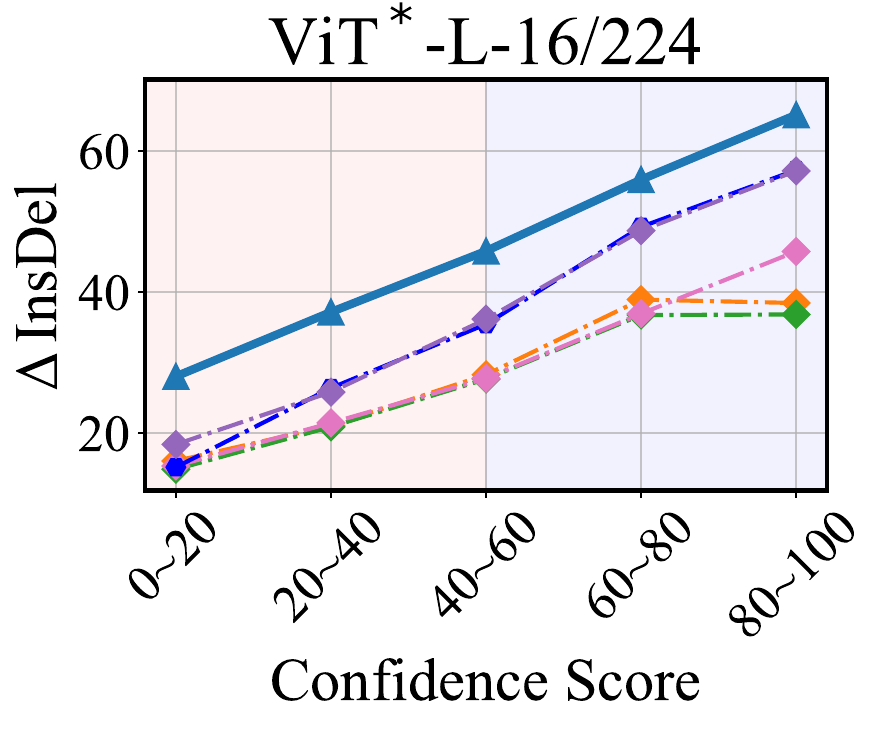}
\includegraphics[width=.18\linewidth]{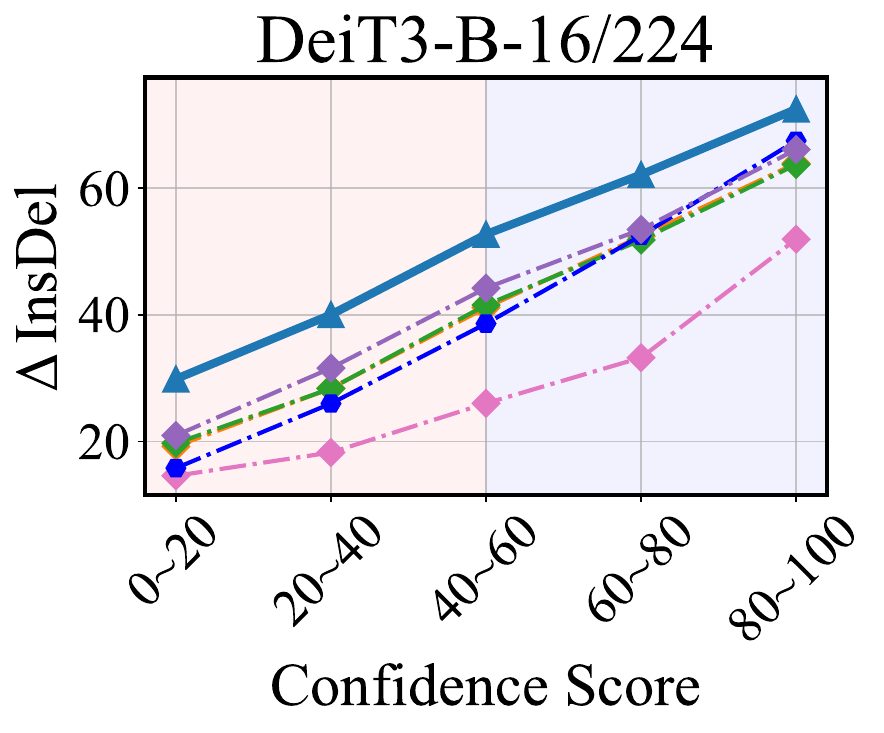}
\includegraphics[width=.18\linewidth]{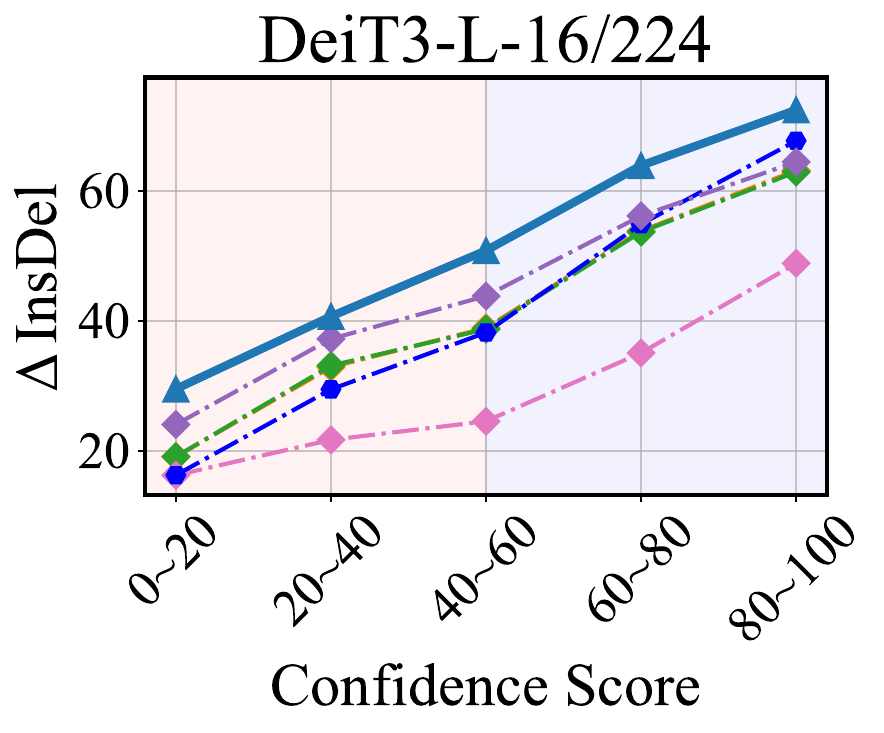}
\includegraphics[width=.18\linewidth]{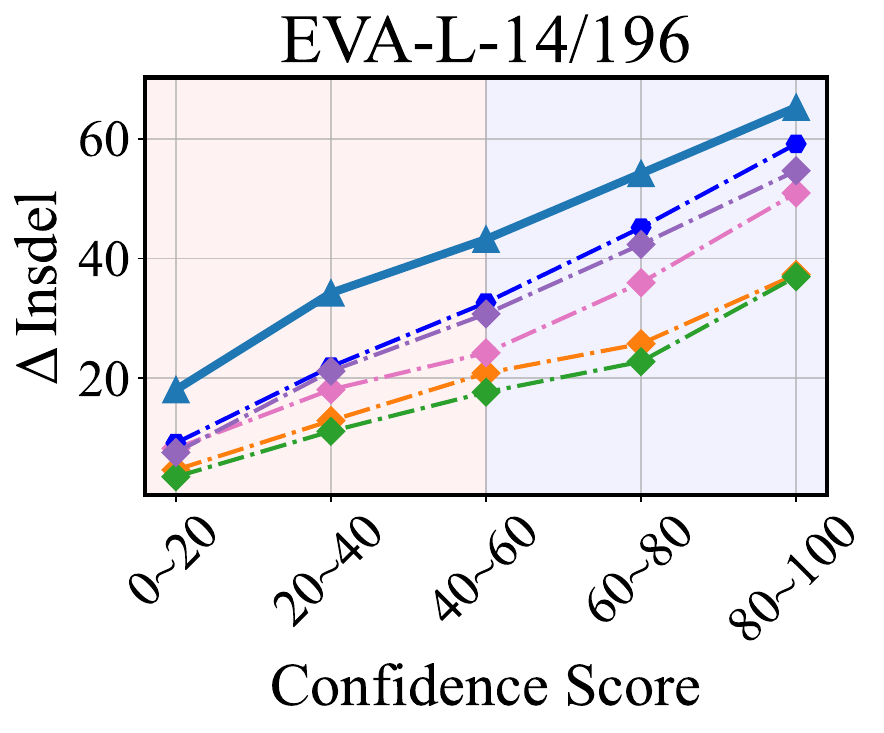}
\vspace{-2pt}
}
% \vspace{-7pt}
% \vspace{-7pt}
\vspace{-10pt}

\caption{\textbf{Difficulty-aware correctness assessment on insertion/deletion}
We measure differences in insertion/deletion ($\Delta$InsDel$\uparrow$) scores (higher is better). We fill the regions including low-confident samples and high-confident samples with red and blue, respectively, based on the prediction made by the model.
% We include additional results in the supplementary material.
}
\vspace{-10pt}
\label{fig:difficulty_aware_full1}
\end{figure*}

\subsection{Insertion/Deletion}
We provide additional results of insertion/deletion in a wide range of models and datasets.
For the deletion test, we leverage the image filled with zero pixels as a baseline, indicating a non-informative image.
For the insertion test, we blur the input image using the 2D Gaussian blurring method with kernel size $51$ and sigma $50$.
To demonstrate the generalizability of \ours{}, we include the model pre-trained with self-supervised learning and the results of IN-A and IN-R datasets in Tab.~\ref{tab:sup_insdel}.
As shown in the results, \ours{} provides attribution maps with remarkable correctness scores compared to the baselines.
In particular, the correctness performance in IN-A and IN-R indicate that \ours{} provides attribution maps regardless of the difficulty of input samples.

\subsection{Difficulty-aware Analysis}
We provide the additional quantitative results of difficulty-aware analysis.
First, we provide the confidence scores computed by the model per sample for different datasets including IN-1k, IN-A, and IN-R.
Second, we include the quantitative results to demonstrate that \ours{} consistently improves the correctness of resulting attribution maps for various confident samples.

\subsubsection{Distribution of Confidence Scores}
\begin{figure}[t!]
\centering

\subfloat[IN-1k]{
\includegraphics[width=.45\linewidth]{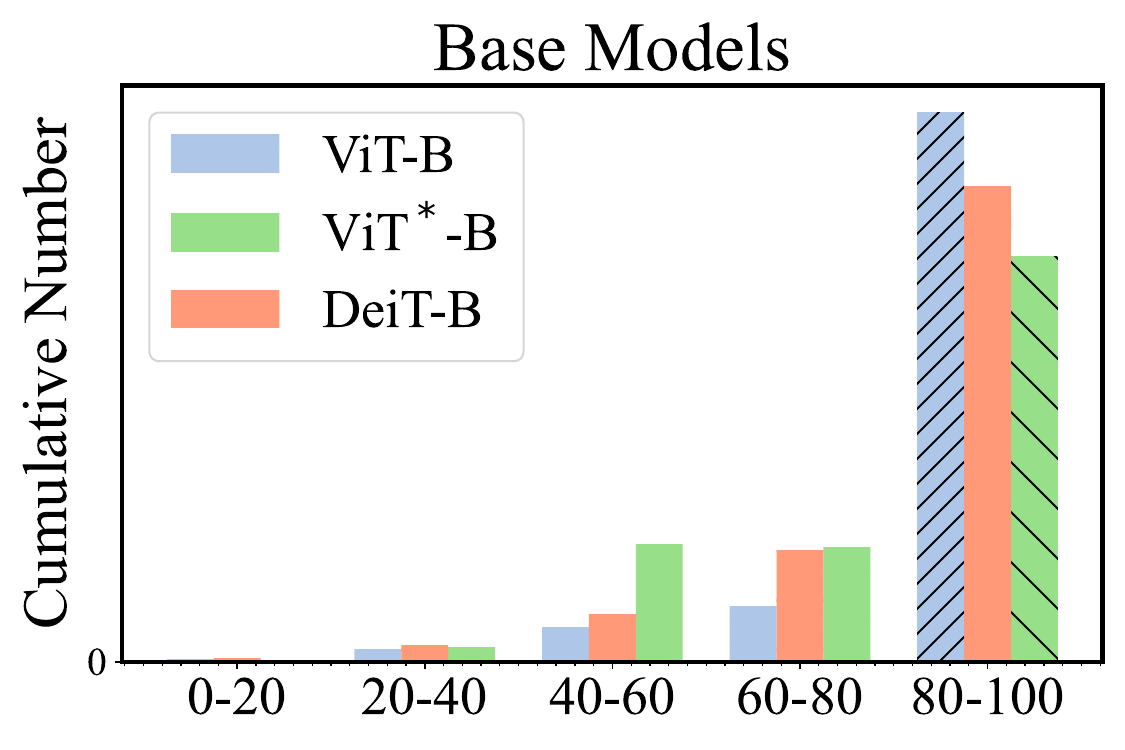}
\includegraphics[width=.45\linewidth]{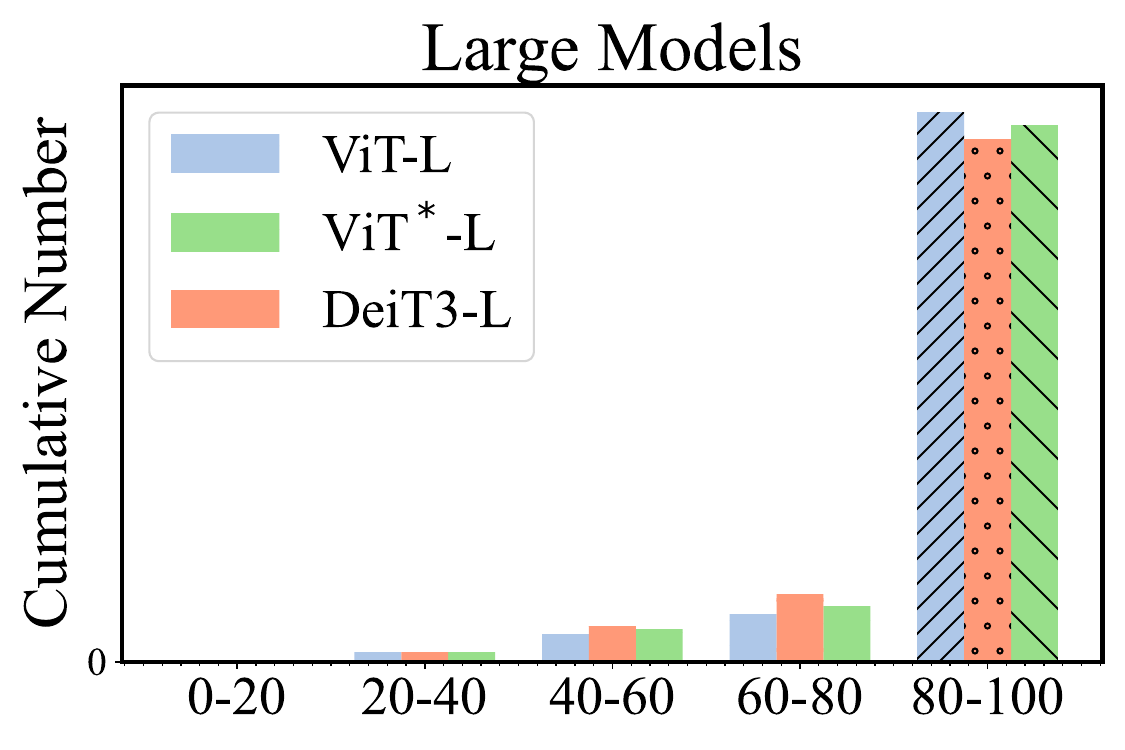}
}

\subfloat[IN-R]{
\includegraphics[width=.45\linewidth]{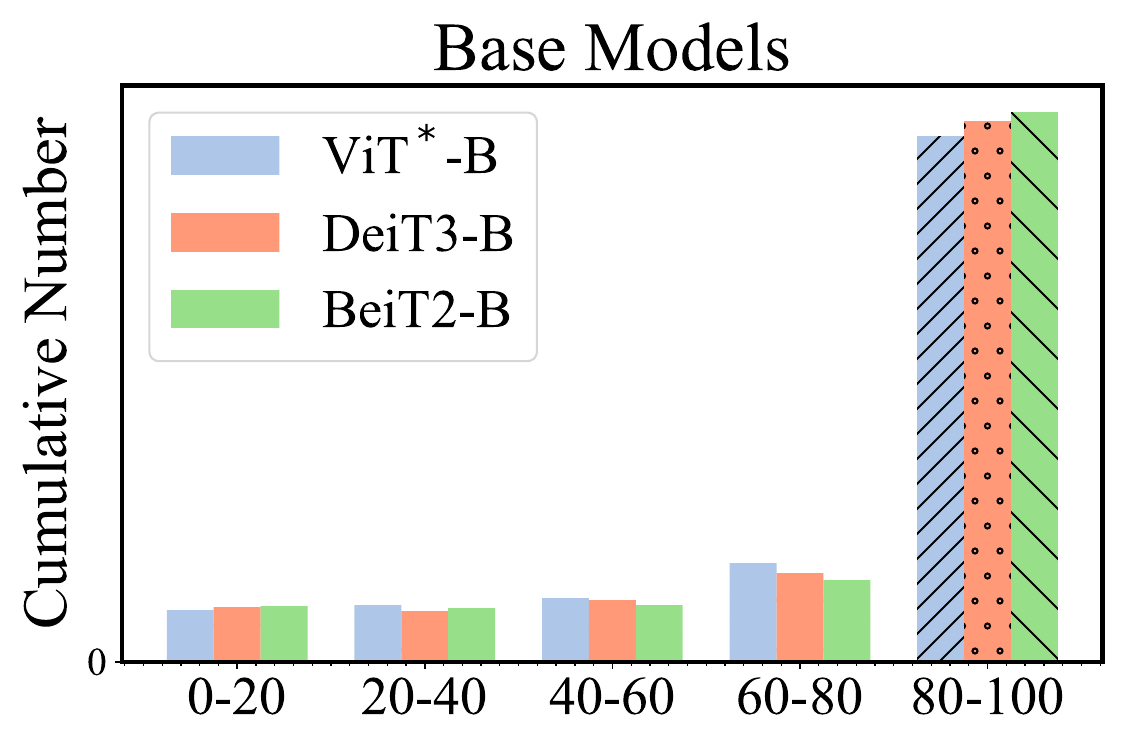}
\includegraphics[width=.45\linewidth]{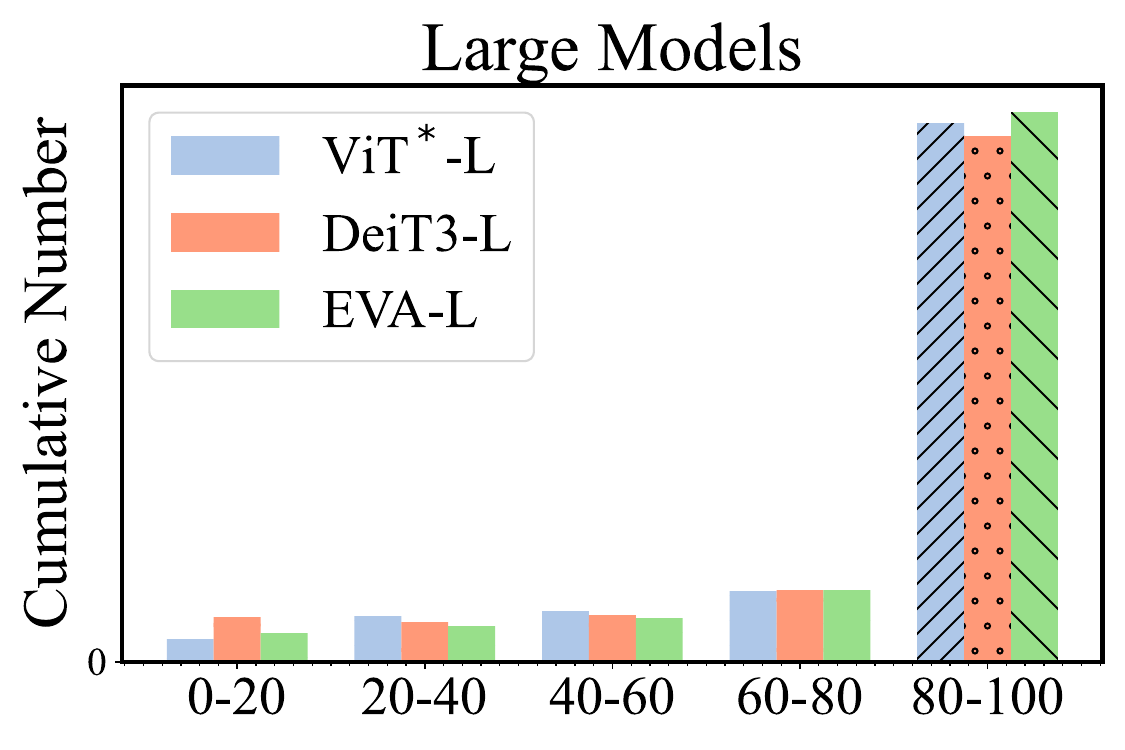}
}

\subfloat[IN-A]{
\includegraphics[width=.45\linewidth]{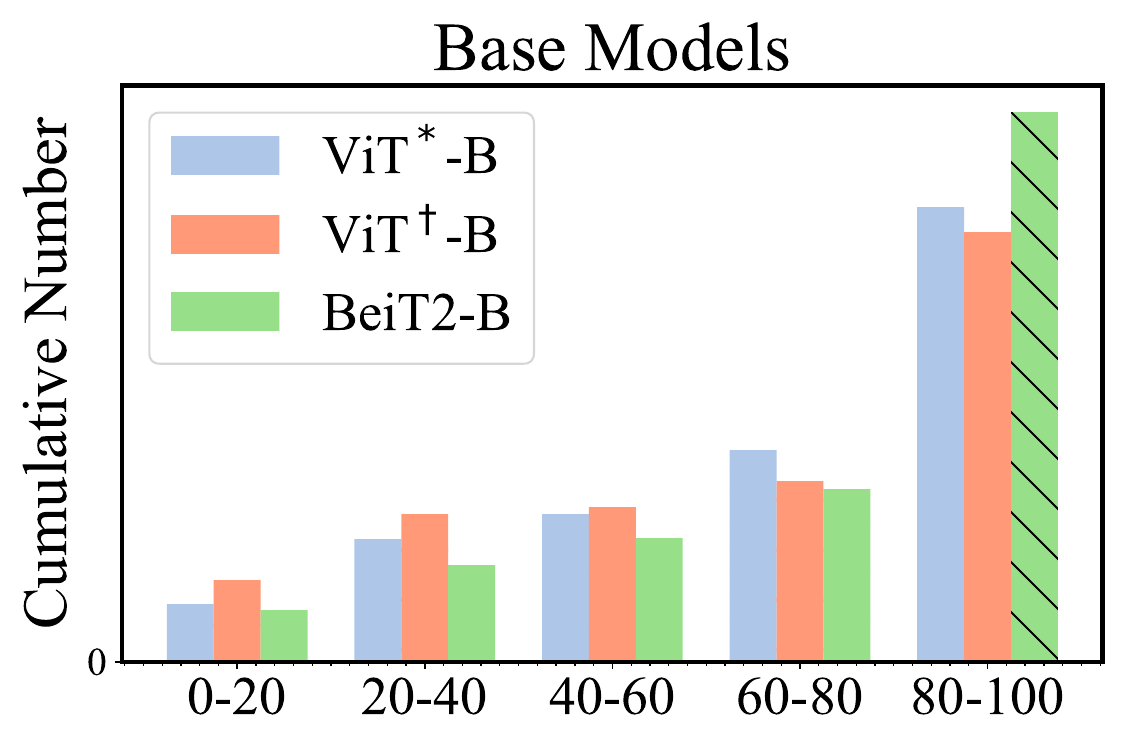}
\includegraphics[width=.45\linewidth]{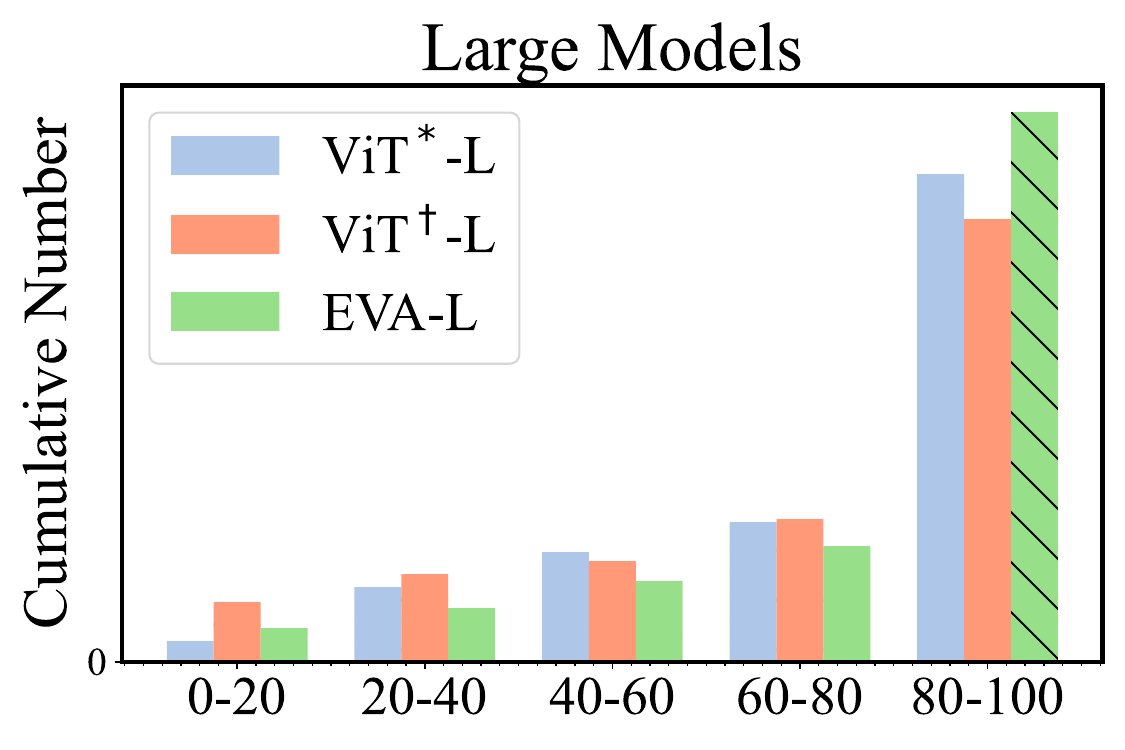}
}

\caption{\textbf{Cumulative number of confidence scores per sample.}
We compare the cumulative number of samples per different confidence scores computed by the model.
We include three datasets including IN-1k, IN-A, and IN-R for the experiment.
}
\label{fig:data_distr}

\end{figure}

Depending on the type of pre-trained parameters of the models and datasets, the resulting confidence score per sample for the model is diversified.
We provide the distribution per sample about confidence scores in Fig.~\ref{fig:data_distr}.
As shown in the figure, massive samples in IN-1k lead the model to output a high confidence score.
In contrast to this, IN-R and IN-A include numerous samples difficult to model.
Thus, amplified correctness scores in IN-A and IN-R demonstrate the capability of \ours{} in generating explanations with a high correctness score compared to the baselines.

\subsubsection{Quantitative Results}
% We provide the additional results of our difficulty-aware analysis, presented in Sec.~\ref{exp:difficulty-aware}.
In addition to the results presented in Sec.~\ref{exp:difficulty-aware}, we provide the additional results in Fig.~\ref{fig:difficulty_aware_full1}.
For IN-1k results, we include the ViT pre-trained with massive regression and CLIP and DeiT3 models.
For IN-A results, we include the same models to provide the results.
As shown in the results, 

\begin{figure}[t!]
\centering

\subfloat[ViT-B-16/224]{
\includegraphics[width=.45\linewidth]{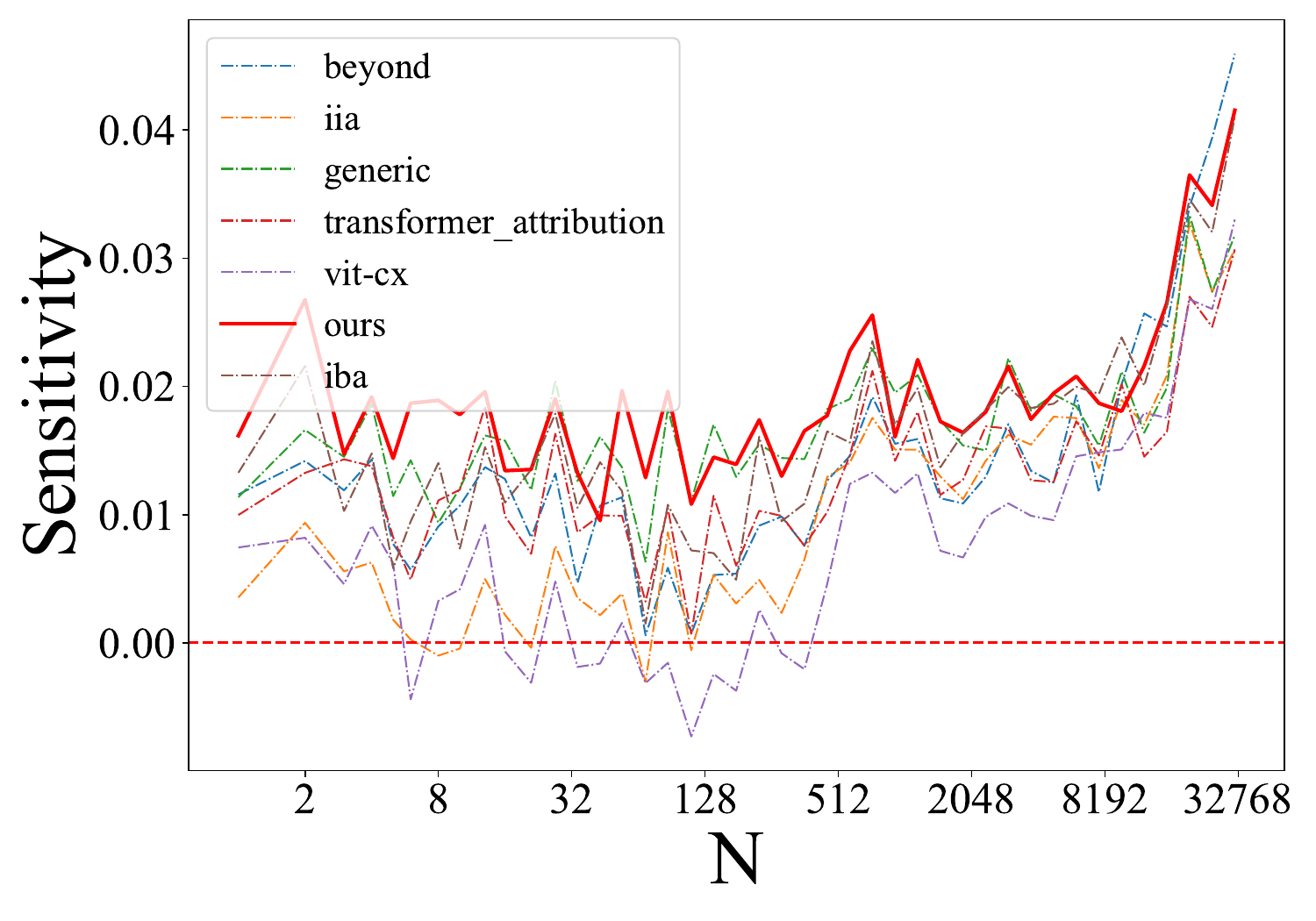}
}
\subfloat[ViT-L-16/224]{
\includegraphics[width=.45\linewidth]{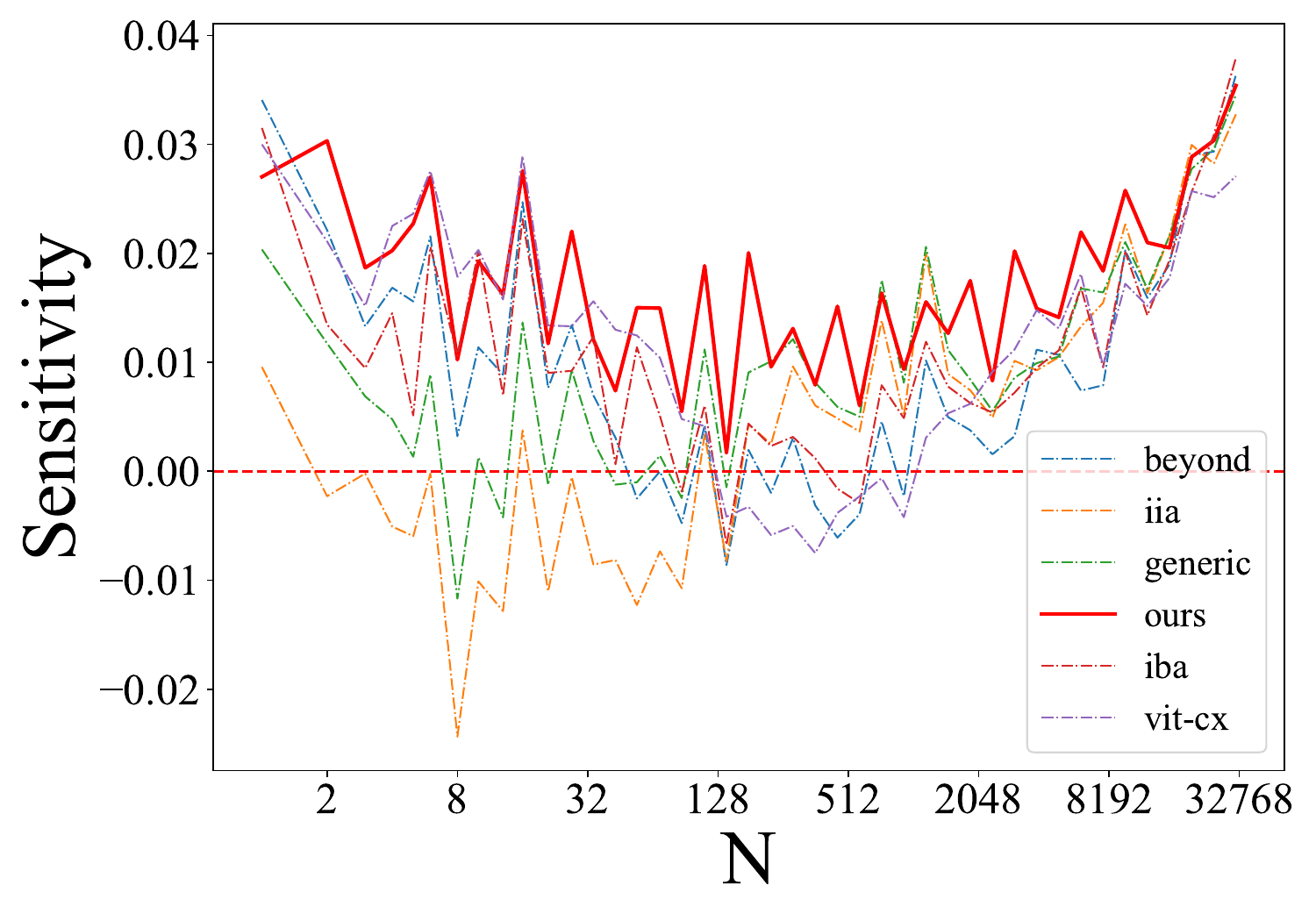}
}
% \subfloat[]{
% }

\subfloat[DeiT-B-16/224]{
\includegraphics[width=.45\linewidth]{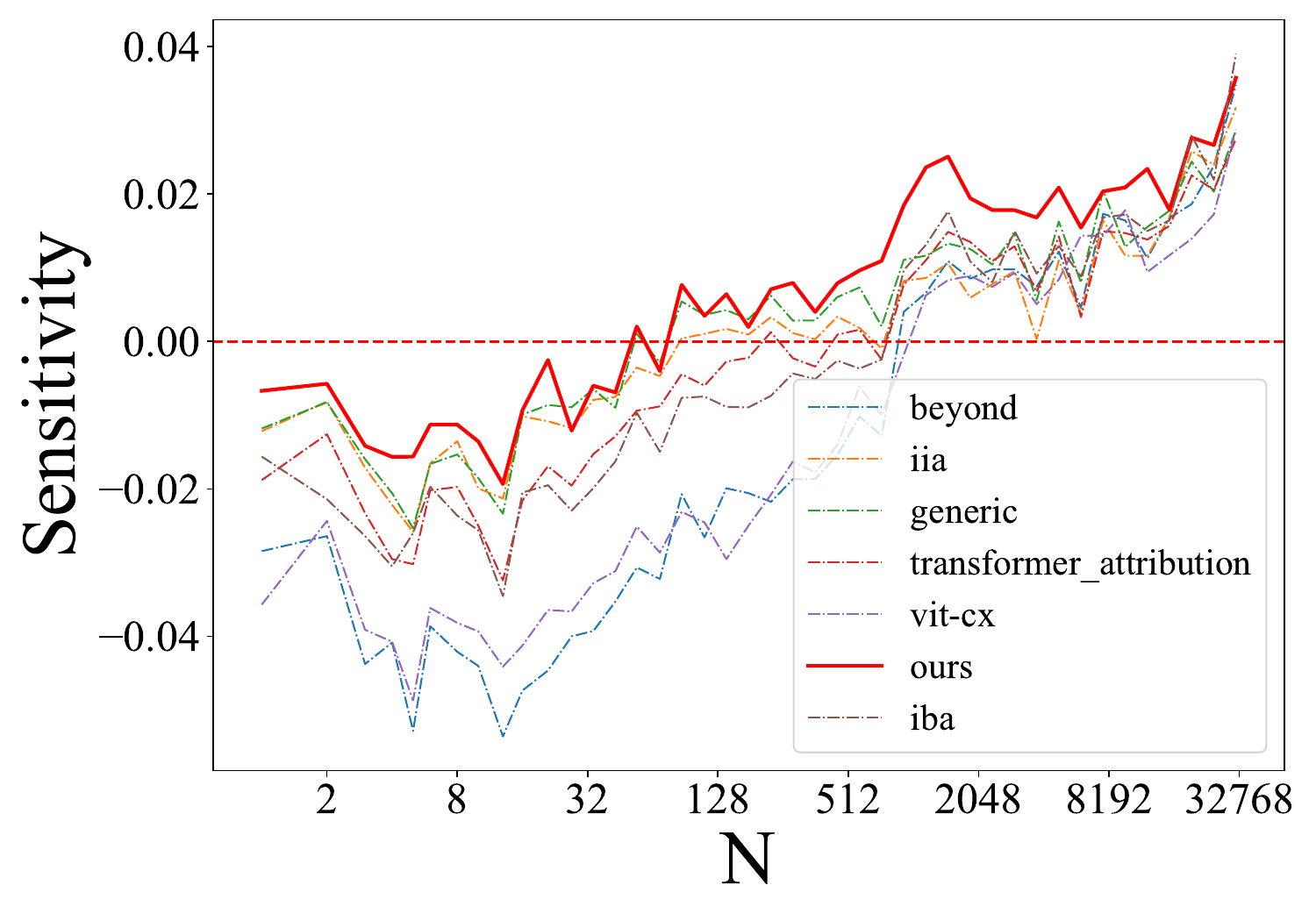}
}
\subfloat[DeiT3-L-16/224]{
\includegraphics[width=.45\linewidth]{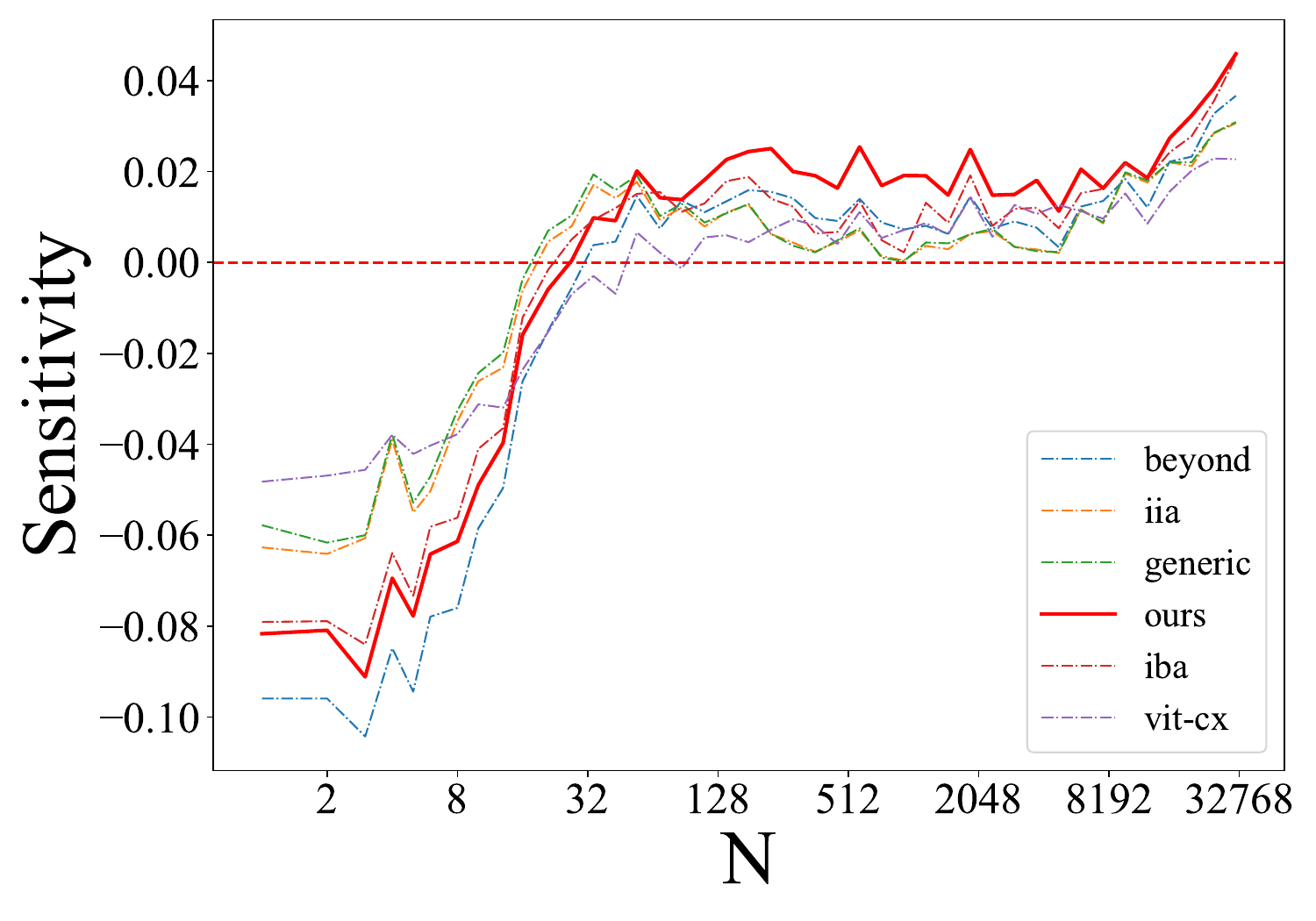}
}

\caption{\textbf{Quantitative results of Sensitivity-N.}
We compare the results of Sensitivity-N against the base and large models.
We include the ViT and DeiT models.
% To demonstrate the generalizability of \ours{}, we compare the results between the base and large models. 
We plot the performance of \ours{} with a solid line with red color. }
\label{fig:sensitivity_n}

\end{figure}

% \begin{figure}[t]
% \centering
% \begin{tabular}{c c}
% \hline
% Method & EHR \\
% \hline
% InputIBA & \textbf{0.476} $\pm$ 0.007 \\
% IBA & 0.356 $\pm$ 0.005\\
% GradCAM & 0.283 $\pm$ 0.005\\
% Guided BP & 0.421 $\pm$ 0.005\\
% Extremal Perturbation & 0.421 $\pm$ 0.007\\
% DeepSHAP & 0.183 $\pm$ 0.002\\
% Integrated Gradients & 0.155 $\pm$ 0.002\\
% \hline
% \end{tabular}
% \caption{\textbf{Quantitative Visual Evaluation - EHR}: This metric evaluates how precisely the attributions localize the features by comparing them with ground truth bounding boxes. InputIBA, extremal perturbations, and Guided Backpropagation score highest. IBA and GradCAM also perform well, but due to their lower resolution maps, they receive lower scores. The standard error is also presented in the table.}
% \label{tab:ehr}

% \end{figure}

% \setlength{\tabcolsep}{5pt}
\begin{table}[t!]
\centering
\footnotesize
% \begin{subtable}[t!]{0.48\linewidth} %%%%%%
\subfloat[Non-finetuned model vs finetuned model~\cite{robust}]{
% \begin{centering}
% \renewcommand{\arraystretch}{1.0} 
% \begin{adjustbox}{width=.475\textwidth}
% \begin{tabular}{p{1.4cm}  p{1.4cm}  p{1.25 cm}  p{1.25 cm}  p{1.25 cm}  p{1.25 cm}}
\resizebox{0.85\linewidth}{!}{
\begin{tabular}{ccccccc}
     \specialrule{2.0pt}{1pt}{1pt} % (all/diff)
     \noalign{\vspace{1.5pt}}
     % \noalign{\smallskip}
    % \multirow{2}{*}{Model}&  & \multicolumn{2}{c}{\underline{\hspace{2.5em}Difficult\hspace{2.5em}}} & \multicolumn{2}{c}{\underline{\hspace{3.5em}Easy\hspace{3.5em}}} \\
    \multirow{2.3}{*}{Model}& & \multicolumn{2}{c}{Non-finetuned} & & \multicolumn{2}{c}{Finetuned~\cite{robust}} \\
    % \noalign{\smallskip}
    \noalign{\vspace{1.5pt}}
    \cline{3-4}
    \cline{6-7}
    \noalign{\vspace{1.5pt}}
    & & ViT-B & DeiT-B & & ViT-B & DeiT-B \\
    \midrule 
Chefer-LRP & & 0.277 & 0.252 & & 0.297 & 0.274  \\
Generic & & 0.229 & 0.203 & & 0.285 & 0.232  \\
IIA & & 0.256 & 0.239 & & 0.293 & 0.271  \\
ViT-CX & & 0.240 & 0.212 & & 0.229 & 0.210  \\
IBA & & \textbf{0.297} & 0.235 & & \textbf{0.310} & 0.250  \\
Beyond & & 0.252 & 0.216 & & 0.263 & 0.208  \\
\ours{} & & 0.248 & \textbf{0.269} & & 0.253 & \textbf{0.281}  \\
    \specialrule{2.0pt}{1pt}{1pt}
\end{tabular}
}
}

\subfloat[Varaints of training strategy]{
% \begin{centering}
% \renewcommand{\arraystretch}{1.0} 
% \begin{adjustbox}{width=.475\textwidth}
% \begin{tabular}{p{1.4cm}  p{1.4cm}  p{1.25 cm}  p{1.25 cm}  p{1.25 cm}  p{1.25 cm}}
\resizebox{0.85\linewidth}{!}{
\begin{tabular}{ccccccc}
     \specialrule{2.0pt}{1pt}{1pt} % (all/diff)
     \noalign{\vspace{1.5pt}}
     % \noalign{\smallskip}
    % \multirow{2}{*}{Model}&  & \multicolumn{2}{c}{\underline{\hspace{2.5em}Difficult\hspace{2.5em}}} & \multicolumn{2}{c}{\underline{\hspace{3.5em}Easy\hspace{3.5em}}} \\
    \multirow{2.3}{*}{Model}& & \multicolumn{2}{c}{ViT$^*$} & & \multicolumn{2}{c}{ViT$^\dagger$} \\
    % \noalign{\smallskip}
    \noalign{\vspace{1.5pt}}
    \cline{3-4}
    \cline{6-7}
    \noalign{\vspace{1.5pt}}
    & & ViT-B & ViT-L & & ViT-B & ViT-L \\
    \midrule 
Chefer-LRP & & - & - & & - & -  \\
Generic & & 0.200 & 0.103 & & 0.247 & 0.226 \\
IIA & & 0.107 & 0.110 & & 0.237 & 0.234 \\
ViT-CX & & 0.232 & 0.221 & & 0.212 & 0.234 \\
IBA & & 0.226 & 0.191 & & 0.258 & 0.235 \\
Beyond & & 0.286 & 0.218 & & 0.225 & 0.205 \\
\ours{} & & \textbf{0.300} & \textbf{0.265} & & \textbf{0.305} & \textbf{0.304} \\
    \specialrule{2.0pt}{1pt}{1pt}
\end{tabular}
}
}

\subfloat[Variants of patch size and depth]{
% \begin{centering}
% \renewcommand{\arraystretch}{1.0} 
% \begin{adjustbox}{width=.475\textwidth}
% \begin{tabular}{p{1.4cm}  p{1.4cm}  p{1.25 cm}  p{1.25 cm}  p{1.25 cm}  p{1.25 cm}}
\resizebox{0.85\linewidth}{!}{
\begin{tabular}{ccccccc}
     \specialrule{2.0pt}{1pt}{1pt} % (all/diff)
     \noalign{\vspace{1.5pt}}
     % \noalign{\smallskip}
    % \multirow{2}{*}{Model}&  & \multicolumn{2}{c}{\underline{\hspace{2.5em}Difficult\hspace{2.5em}}} & \multicolumn{2}{c}{\underline{\hspace{3.5em}Easy\hspace{3.5em}}} \\
    \multirow{2.3}{*}{Model}& & \multicolumn{2}{c}{Patch size} & & \multicolumn{2}{c}{Depth} \\
    % \noalign{\smallskip}
    \noalign{\vspace{1.5pt}}
    \cline{3-4}
    \cline{6-7}
    \noalign{\vspace{1.5pt}}
    % & & \begin{tabular}[c]{c@{}}$8$\end{tabular} & $32$ & & \begin{tabular}[c]{c@{}}ViT-L\end{tabular} & DeiT3-L \\
    & & 8 & 32 & & ViT-L & DeiT3-L \\
    \midrule 
Chefer-LRP & & - & - & & 0.236 & -  \\
Generic & & 0.169 & 0.143 & & 0.173  & 0.297 \\
IIA & & 0.201 & 0.150 & & 0.176 & 0.304 \\
ViT-CX & & 0.187 & 0.236 & & 0.212 & 0.207 \\
IBA & & 0.211 & 0.246 & & 0.213 & 0.245 \\
Beyond & & 0.202 & 0.243 & & 0.201 & 0.207 \\
\ours{} & & \textbf{0.272} & \textbf{0.263} & & \textbf{0.247} & \textbf{0.334} \\
    \specialrule{2.0pt}{1pt}{1pt}
\end{tabular}
}
}
%%%%%%%
% \end{adjustbox}
% \end{centering}
\vspace{-8pt}
\caption{\textbf{Quantitative visual evaluation results of EHR.} 
This metric evaluates the localization capability of the feature attribution methods.
% The ViT-B and DeiT-B denote ViT-B-16/224 and DeiT-B-16/224, respectively.
We omit patch size and input resolution for simplicity.
For the variants of patch size, we utilize ViT-B-8/224 and ViT-B-32/224.
The robust~\cite{robust} indicates the model is further fine-tuned to focus on the foreground object.
We denote ViT pre-trained with massive regularization and CLIP as Reg and CLIP.
}
\vspace{-10pt}
\label{tab:ehr}
\end{table}

\subsection{Sensitivity-N} \label{sec:sensitivity_n} % Sanity check와 합쳐도 될듯. 
The sensitivity-N~\cite{sensitivity_n} evaluates feature attribution assessment by measuring the correlation.
The sensitivity is measured between the sum of attributions corresponding to the mask indices and the drop in model confidence caused by the rest of the feature subsets.
We leverage the Pearson correlation coefficient (PCC) to measure the correlation.
To compute the sensitivity, we compute the condition over 100 different indices and average them over 1,000 image samples.
We generate indices from 1 to 80\% of the number of pixels.
Loosely speaking, sensitivity-N extends the \textit{summation over delta} and \textit{completeness}.
Thus, the attribution maps yielding the high sensitivity provide a faithful explanation.
We compare the sensitivity-N of the original and enhanced ViT architecture.
The quantitative results of the sensitivity-N become visually distinguishable as shown in Fig.~\ref{fig:sensitivity_n}.
Compared to the existing approaches, \ours{} provides faithful attribution maps after the $10^3$ pixels are removed.

% % Appendix 후보 (1)
\subsection{Localization Assessment}
To assess the localization ability of \ours{} compared to the baselines, we compare the effective heat ratio (EHR)~\cite{inputiba} of the attribution maps yielded by different methods.
This EHR measures whether the explanation highlights object-focused attribution.
To this end, EHR computes the 
As shown in Tab.~\ref{tab:ehr}, the attribution maps provided by \ours{} correctly highlight the foreground object.
We include the ViT and DeiT models additionally fine-tuned to concentrate the object~\cite{robust}.
The results show that \ours{} provides the attribution maps that correctly highlight the human-labeled bounding box.
In particular, according to the results of ViT-B-8/224, when the granularity of the attributions is increased as the patch size of self-attention is decreased, the \ours{} the significant performance in localization.
These results demonstrate that the overall attribution maps produced by \ours{} correctly highlight the foreground object in addition to the increased correctness performance.

\begin{figure}[t!]
\centering

\subfloat[Independent]{
\label{fig:sanity_check_indep}
\includegraphics[width=.53\linewidth]{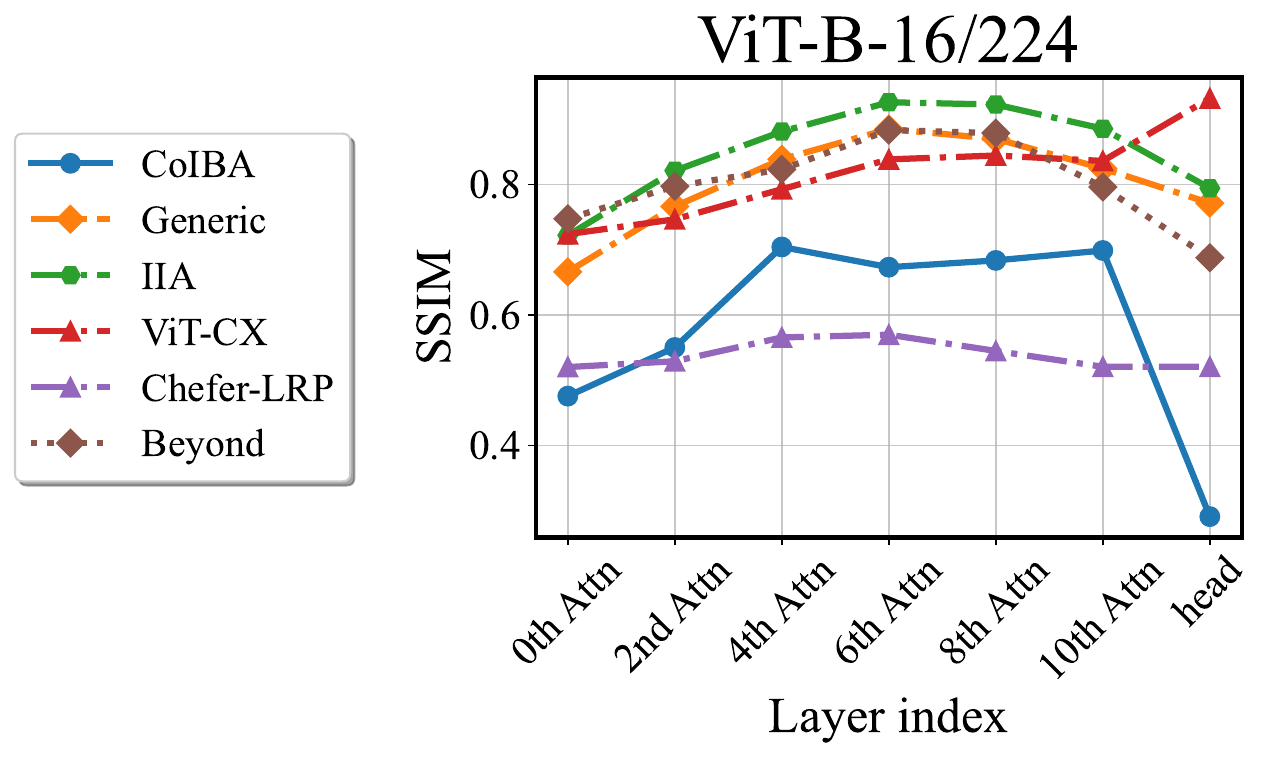}
}
\hspace{-5pt}
\subfloat[Cumulative]{
\label{fig:sanity_check_cumul}
\includegraphics[width=.37\linewidth]{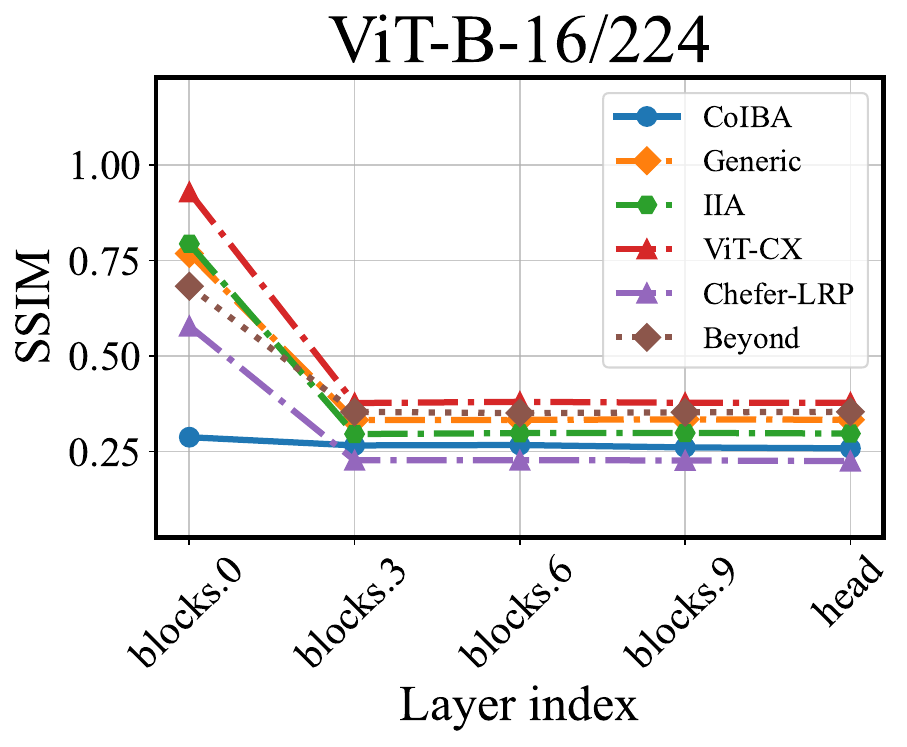}
\vspace{2pt}
}

\subfloat[Qualitative Results]{
\includegraphics[width=.9\linewidth]{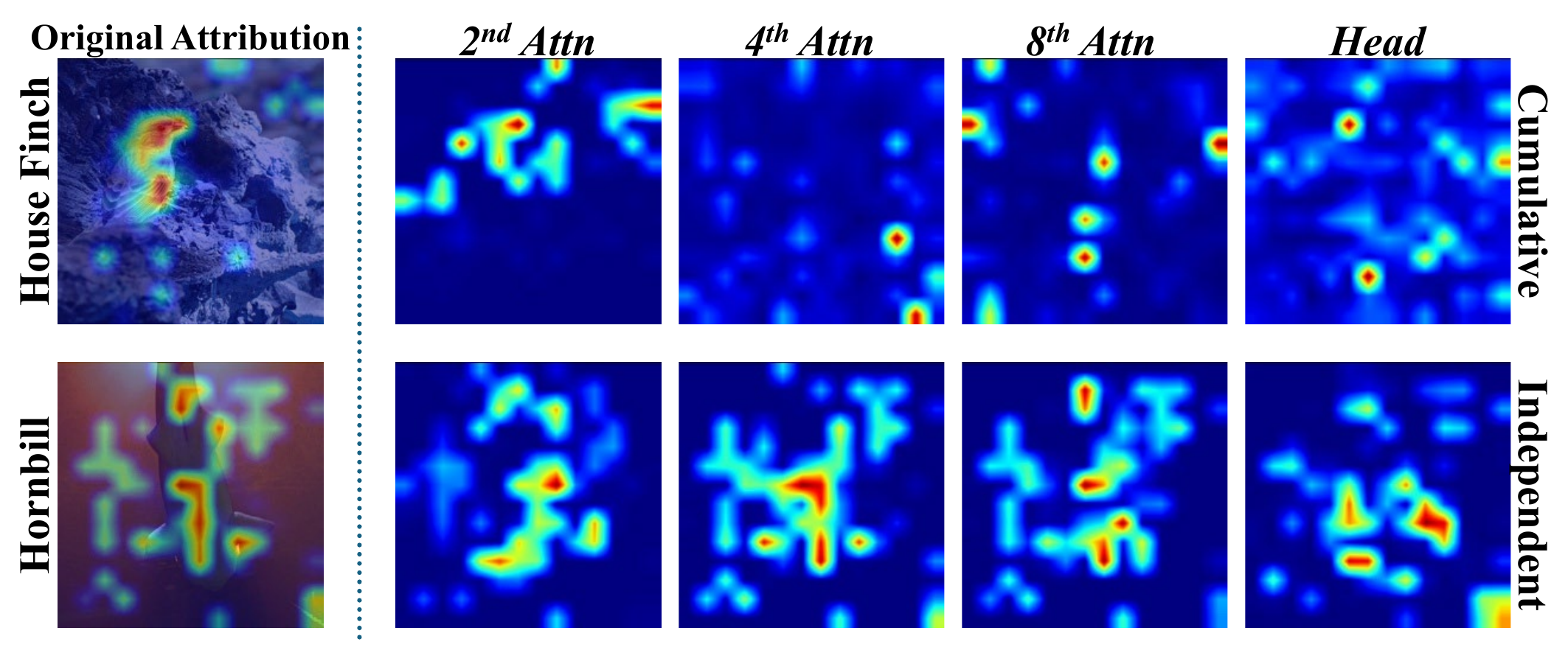}
}

\vspace{-5pt}
\caption{\textbf{An analysis of parameter randomization test for sanity check.} We utilize ViT-B-16/224 as the baseline architecture. We measure SSIM for the attribution maps produced from different indexed layers after randomization.
The Fig.~\ref{fig:sanity_check}~\subref{fig:sanity_check_indep} and~\ref{fig:sanity_check}~\subref{fig:sanity_check_cumul} represent the comparisons of independent and cumulative parameter randomization tests, respectively.
}
\vspace{-10pt}
\label{fig:sanity_check}

\end{figure}

\begin{table}[!t]
\centering
\begin{centering}

\resizebox{0.98\linewidth}{!}{
\begin{tabular}{lccccccc}
\specialrule{2.0pt}{1pt}{1pt}
Model                   & & Acc. & $\beta=0.01$ & $\beta=0.1$ & $\beta=1$ & $\beta=10$ & $\beta=100$ \\ \hline
   ViT-B &  & 81.8 & 100.0/86.6 & 100.0/90.2 & 98.8/88.8 & 15.2/13.4 & 1.4/0.6 \\ 
   DeiT-B &  & 82.0 & 100.0/90.1 & 100.0/93.0 & 99.8/94.5 & 36.6/43.3 & 0.1/1.0 \\ 
   % ViT-L &  & & 100.0/87.2 & 100.0/90.81 & 93.8/93.1 & 51.2/51.1 & 1.3/2.2 \\ 
   % DeiT3-B &  & & 100.0/84.3 & 100.0/90.2 & 100.0/92.7 & 84.5/65.4 & 3.1/2.7 \\ 

\specialrule{2.0pt}{1pt}{1pt}
\end{tabular}
}
\vspace{-8pt}
\caption{\textbf{Quantitative comparison on various $\beta$ settings.} 
We report top-1 accuracy per trade-off hyper-parameter setting $\beta$.
We divide correctly/incorrectly predicted samples.
}
\vspace{-10pt}
% \end{adjustbox}
\label{tab:beta_comparison}
\end{centering}
\end{table}

% Appendix 후보 (1)
\subsection{Sanity Check}
The sanity check~\cite{sanity_check} confirms whether the produced explanations are sensitive to the model parameter.
This experiment measures the similarity of the attributions produced with non-randomized and randomized model parameters.
We include two tests: cumulatively or independently randomizing the parameters of each layer.
For the cumulative parameter randomization, we randomize the model parameters after the 0, 3, 6, and 9 layers and measure the similarity of the attributions produced.
We select each attention layer with interval 2 in an independent parameter randomization test.
We report quantitative results in Fig.~\ref{fig:sanity_check} with exemplary results.
We utilize 1,000 images randomly sampled from the IN-1k validation dataset to measure the similarity between attributions utilizing the SSIM metric.
% As shown in the results, \ours{} provides obfuscated attributions with diminished similarity after the parameter randomization.
The diminished similarity in the results indicates that the attributions are consistently obfuscated as the parameters of each layer are randomized.
Therefore, \ours{} is sensitive to the model parameters, leading to yield faithful attributions to explain the decision-making process.
Furthermore, \ours{} is fairly sensitive to all the layers within the model parameters as the SSIM score results in uniformity across the layers.

\subsection{Discussion}
In this section, we provide further results included in Sec.~\ref{sec:discussion}.
We provide the quantitative results to show the effectiveness of the variational upper bound. After that, we provide the out-of-distribution caused by overly defined trade-off hyperparameter $\beta$.

% \subsubsection{Variational Upper Bound}
% In addition to results in Fig.~\ref{fig:discussion2}\subref{fig:discussion2_3_cka}, which includes ViT-B-16/224, we provide further results obtained by ViT-B-16/224 pre-trained with CLIP and DeiT-B-16/224 in Fig.~\ref{fig:sup_discussion2}.
% As illustrated in the results, optimizing the information bottleneck objective with our variational upper bound has high similarity compared to layer-wise attributions.
% This demonstrates our claim that the variational upper bound suggested in this paper is better at reflecting layer-wise attributions into the comprehensive attribution map provided by \ours{}, aligning with our observation as discussed in Sec.~\ref{sec:discussion_upperbound}.
% The reason for these results is that the variational upper bound setting is more sensitive to model parameters compared to a linearly combined mutual information objective.
% To demonstrate this, we independently randomize the trained parameters within the self-attention operation of each layer.
% Therefore, leveraging the variational upper bound suggested in \ours{} provides the attribution map with higher sensitivity than leveraging linearly combined mutual information.

\begin{table}[!t]
\centering
\begin{centering}

\resizebox{0.55\linewidth}{!}{
\begin{tabular}{lccc}
\specialrule{2.0pt}{1pt}{1pt}
Model                   & & IBA & \ours{} \\ \hline
   ViT-B-16/224 &  & 0.060 & 0.062 \\ 
   ViT-L-16/224 &  & 0.168 & 0.174 \\

\specialrule{2.0pt}{1pt}{1pt}
\end{tabular}
}
\vspace{-8pt}
\caption{\textbf{Comparisons of computational cost (sec) required to generate an attribution map.} 
}
\vspace{-10pt}
% \end{adjustbox}
\label{tab:computational_cost}
\end{centering}
\end{table}

\subsubsection{Out of Distribution and Over-compression}
\label{sec:beta_comparison}
The trade-off hyperparameter $\beta$ controls the degree of compression related to the relevancy term in Eq.~\eqref{eq:main2}
The relevant information is suppressed by setting excessive trade-off hyperparameter $\beta$.
In contrast to this, setting this hyperparameter overly small leads to leaving irrelevant information in the activations.
To empirically choose $\beta$, we analyze whether the decision made by the model is corrupted per compressing with different trade-off parameter settings.
As shown in Tab.~\ref{tab:beta_comparison}, setting $\beta$ larger than $1$ leads to over-compression, vanishing the relevant information as well.
In contrast to this, setting $\beta$ smaller than $1$ diminishes the correctness performance of a resulting attribution map.
These results demonstrate the reasonability of our choice $\beta=1$ in dealing with the trade-off between relevancy and compression term.

\section{Computational Cost}
We report the computational cost while generating a single attribution map for an input sample.
We utilize NVIDIA A6000 GPU to measure the computational time.
As shown in Tab.~\ref{tab:computational_cost}, \ours{} requires a similar computational cost compared to IBA.
For example, IBA and \ours{} consume 0.06 and 0.062 (sec) for computing the attribution map.
Therefore \ours{} requires a significantly small computational cost, compared to dealing with a specific layer, 
Therefore, a significantly small computational cost required by \ours{} demonstrates that \ours{} significantly amplifies the correctness of the resulting attribution map while requiring a small computational cost.
\newpage
% \clearpage
% \section{Qualitative Results}
% \begin{figure*}[!t]
% \centering
% \subfloat[ViT-B-16/224]{
% \includegraphics[width=0.86\textwidth]{Assets/Qualitative_vit-b.pdf}
% }

% \subfloat[ViT-L-16/224]{
% \includegraphics[width=0.86\textwidth]{Assets/Qualitative_vit-l.pdf}
% }
% % \includegraphics[width=1.0\textwidth]{./figs/methods-overview.png}
% % \includegraphics[width=1.0\textwidth]{./figs/methods.png}
% % \vspace{-10pt}
% \caption{\textbf{Visualized attribution maps for IN-k produced from ViT.}
% }
% % \vspace{-13pt}
% \label{fig:vis1}
% \end{figure*}

\begin{figure*}[!th]
\centering
\subfloat[ViT-B-16/224]{
\includegraphics[width=0.85\textwidth]{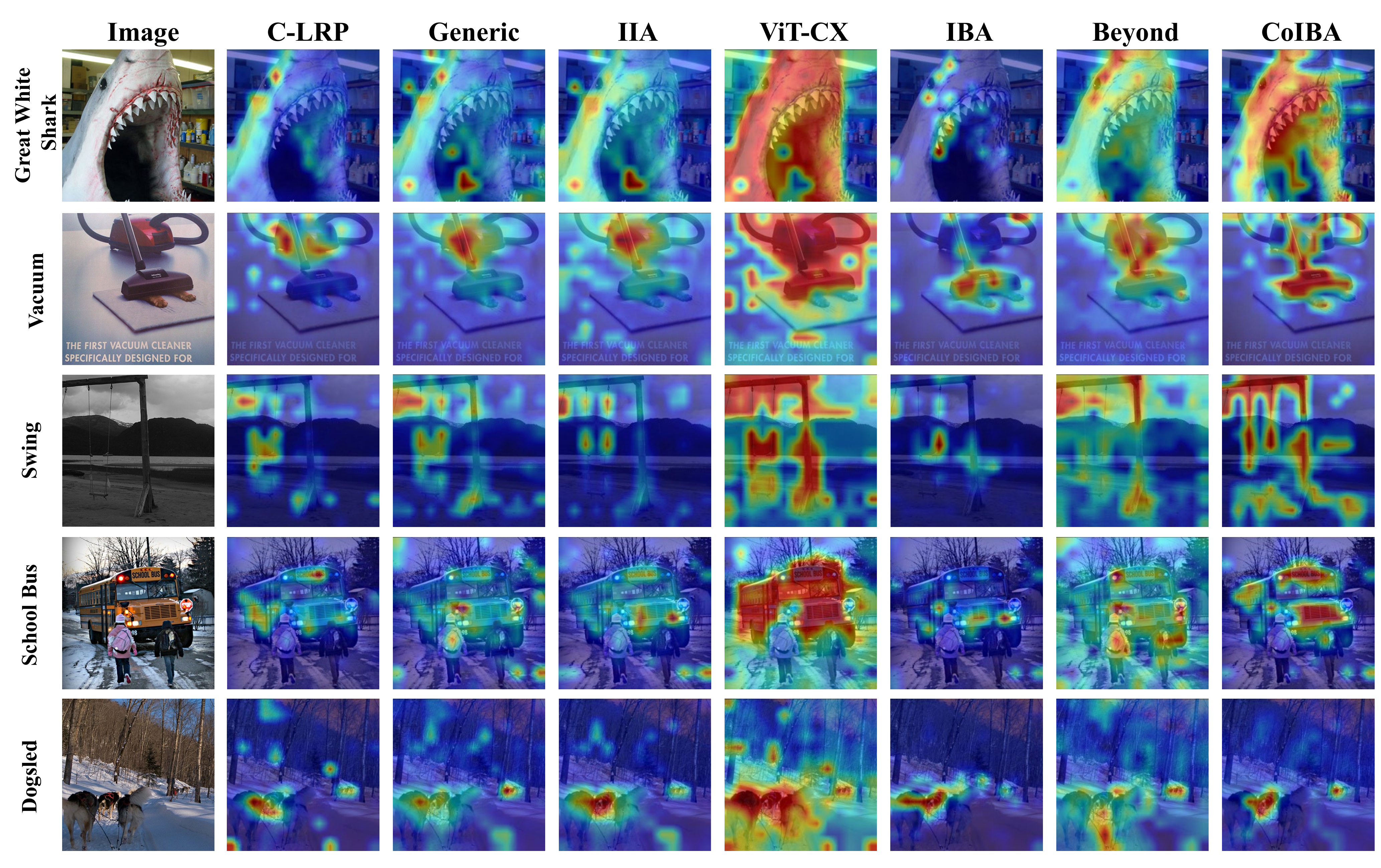}
}

\subfloat[ViT-L-16/224]{
\includegraphics[width=0.85\textwidth]{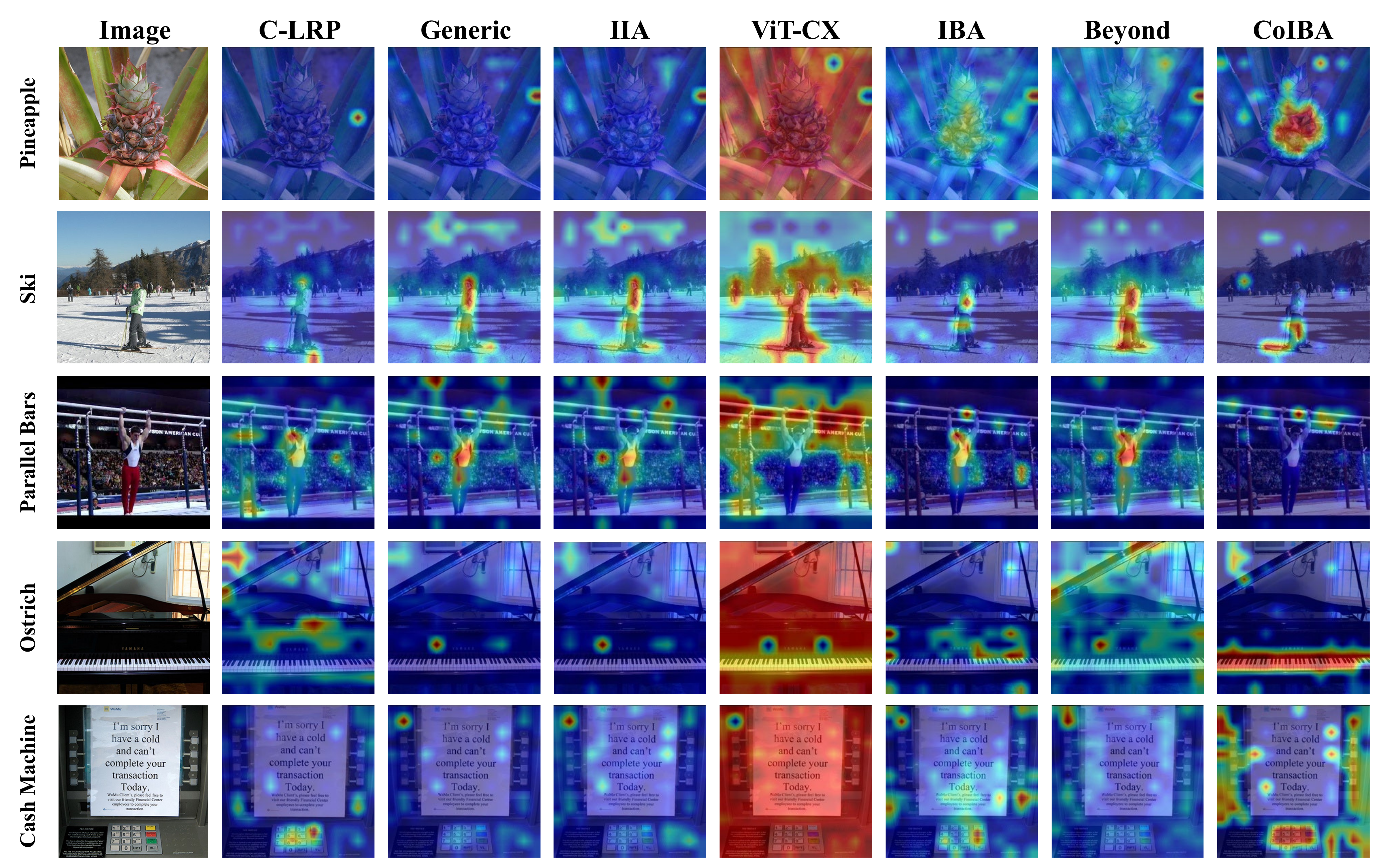}
}
% \includegraphics[width=1.0\textwidth]{./figs/methods-overview.png}
% \includegraphics[width=1.0\textwidth]{./figs/methods.png}
% \vspace{-10pt}
\caption{\textbf{Visualized attribution maps for IN-k produced from ViT.
C-LRP indicates the Chefer-LRP method.}
}
% \vspace{-13pt}

\label{fig:vis1}
\end{figure*}

\clearpage

\begin{figure*}[!th]
\centering
\subfloat[DeiT-B-16/224]{
\includegraphics[width=0.87\textwidth]{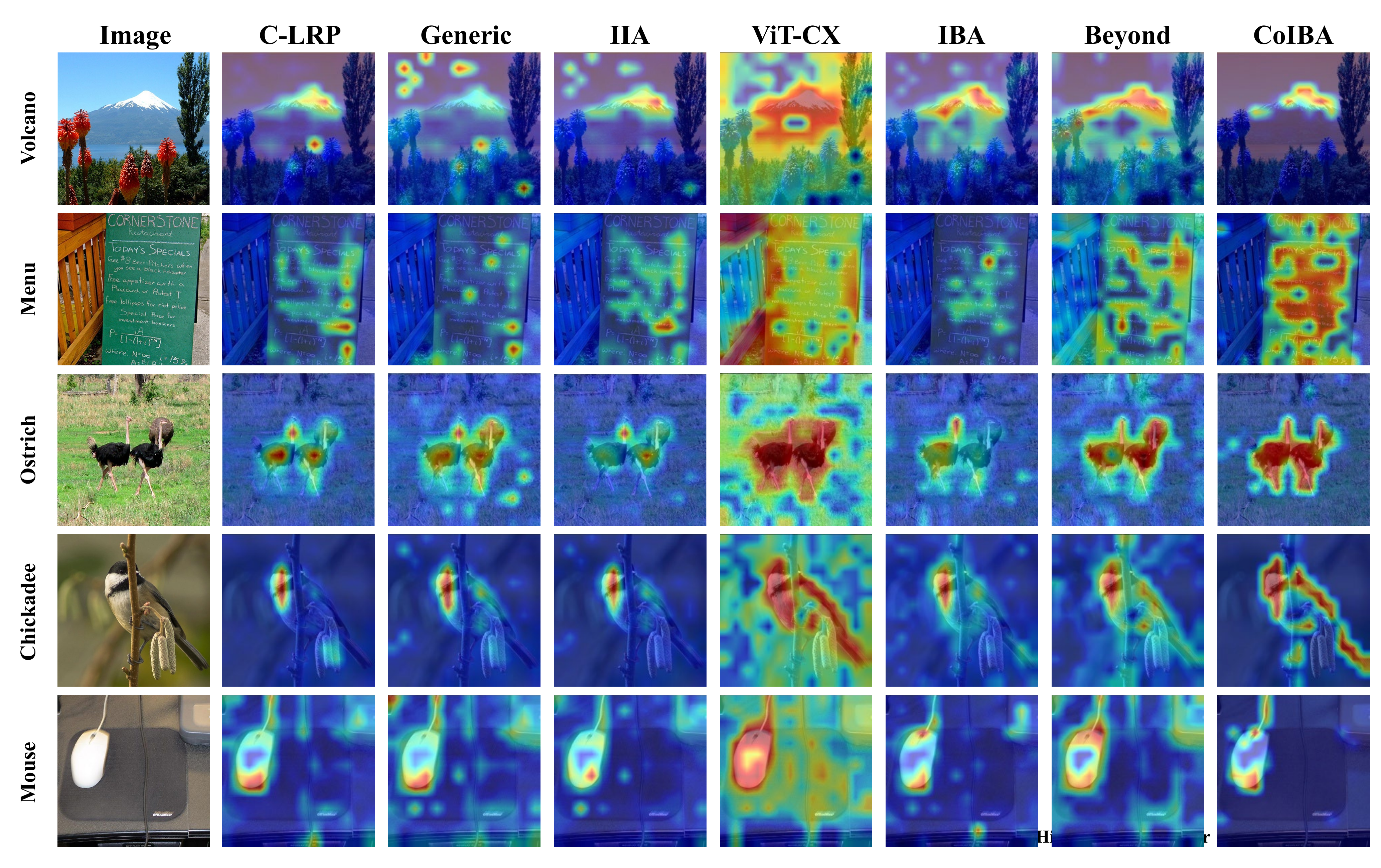}
}

\subfloat[DeiT3-B-16/224]{
\includegraphics[width=0.76\textwidth]{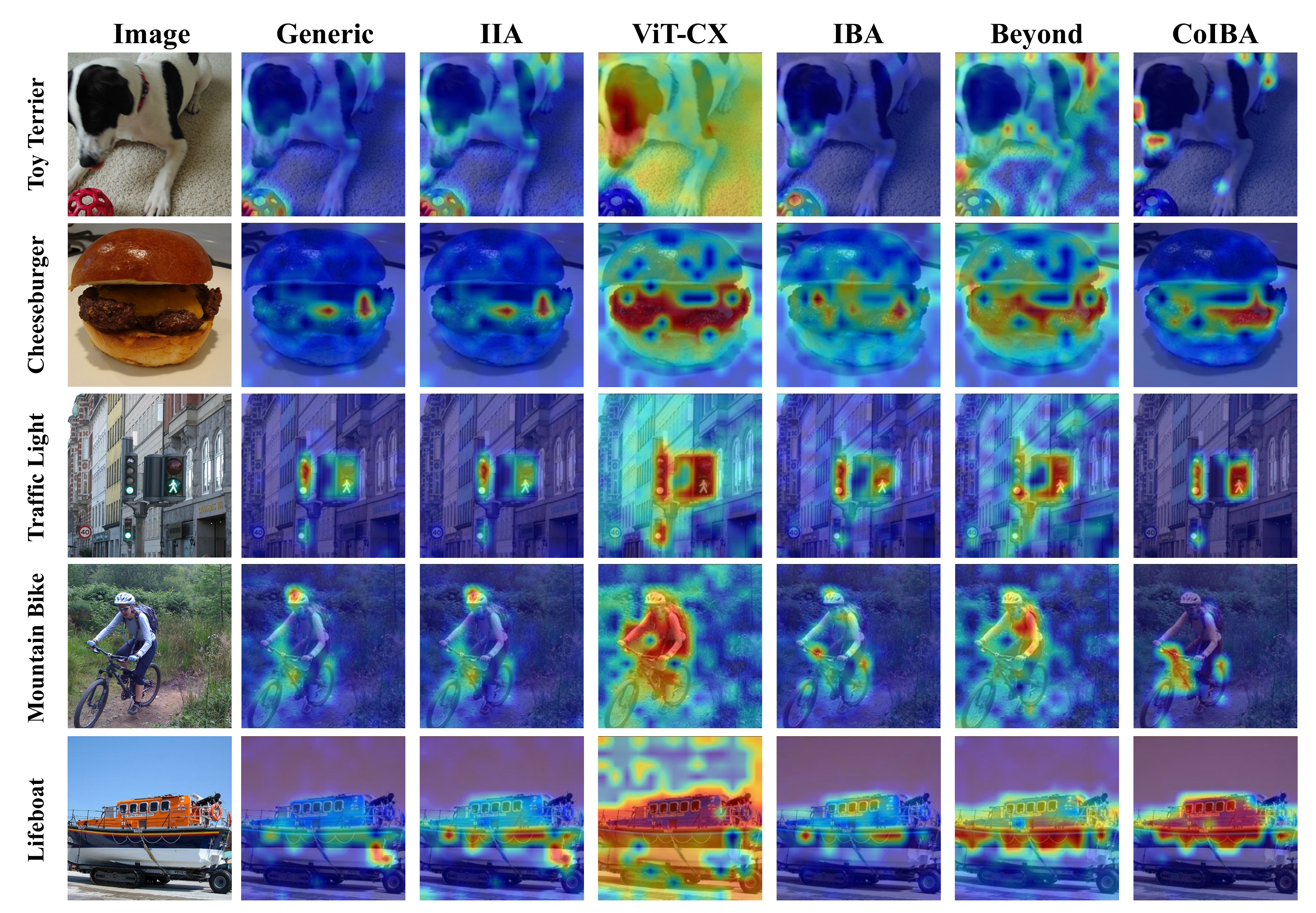}
}
% \includegraphics[width=1.0\textwidth]{./figs/methods-overview.png}
% \includegraphics[width=1.0\textwidth]{./figs/methods.png}
% \vspace{-10pt}
\caption{\textbf{Visualized attribution maps for IN-k produced from DeiT-B and DeiT3-L.}
}
% \vspace{-13pt}

\label{fig:vis2}
\end{figure*}

\clearpage
% \input{figs/sup_class_discriminative}
% \clearpage

\begin{figure*}[!th]
\centering

\subfloat[Class-discriminative ability -- ViT-B-16/224]{
\includegraphics[width=0.85\textwidth]{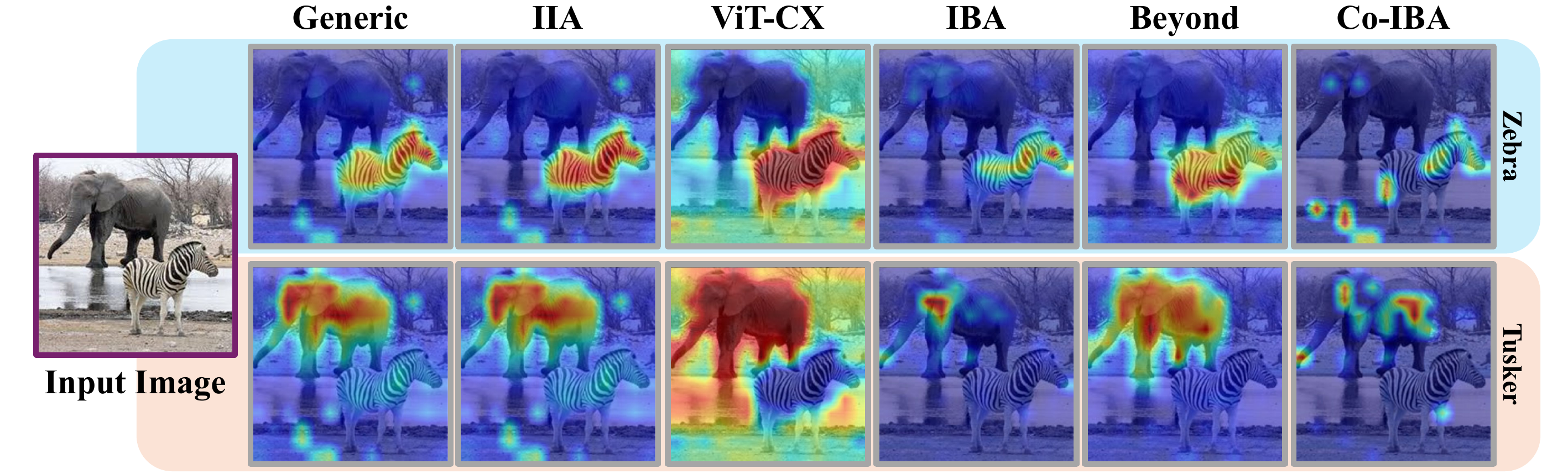}
}

\subfloat[Class-discriminative ability -- ViT$^*$-B-16/224]{
\includegraphics[width=0.85\textwidth]{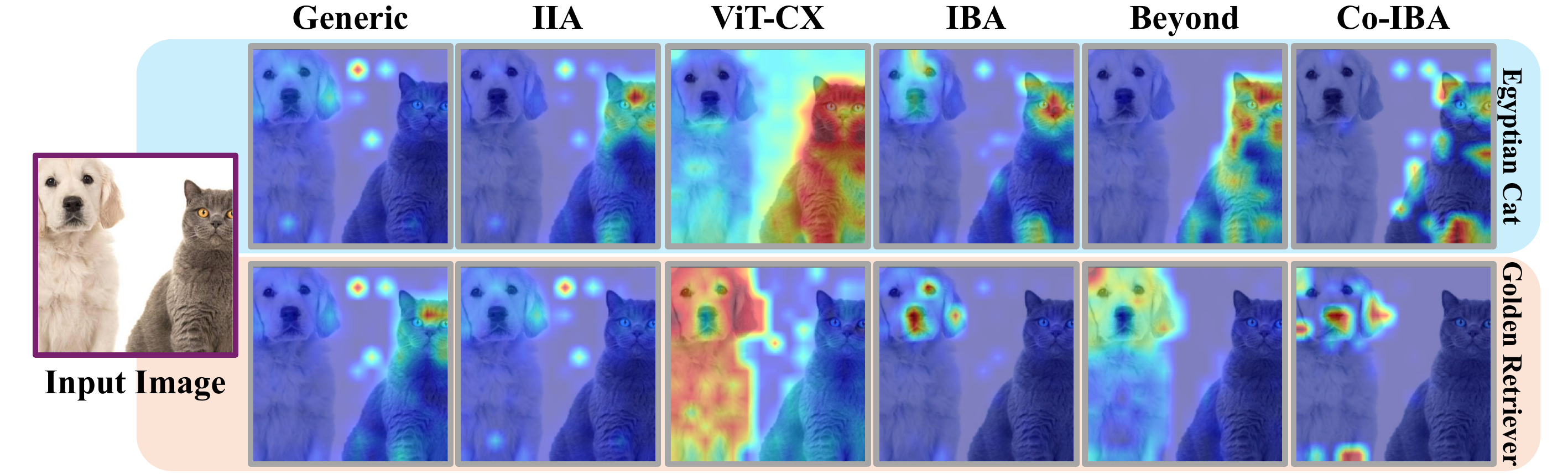}
}

% \subfloat[IN-R]{
% \includegraphics[width=0.85\textwidth]{Assets/Qualitative_vit-b_in-r.pdf}
% }

% \subfloat[IN-A]{
% \includegraphics[width=0.85\textwidth]{Assets/Qualitative_vit-b_in-a.pdf}
% }
% \includegraphics[width=1.0\textwidth]{./figs/methods-overview.png}
% \includegraphics[width=1.0\textwidth]{./figs/methods.png}
% \vspace{-10pt}
\caption{\textbf{
Comparisons of class-discriminative ability from choosing different targeted classes.
}
}
% \vspace{-13pt}

\label{fig:vis3}
\end{figure*}

\begin{figure*}[!th]
\centering

\subfloat[IN-R]{
\includegraphics[width=0.85\textwidth]{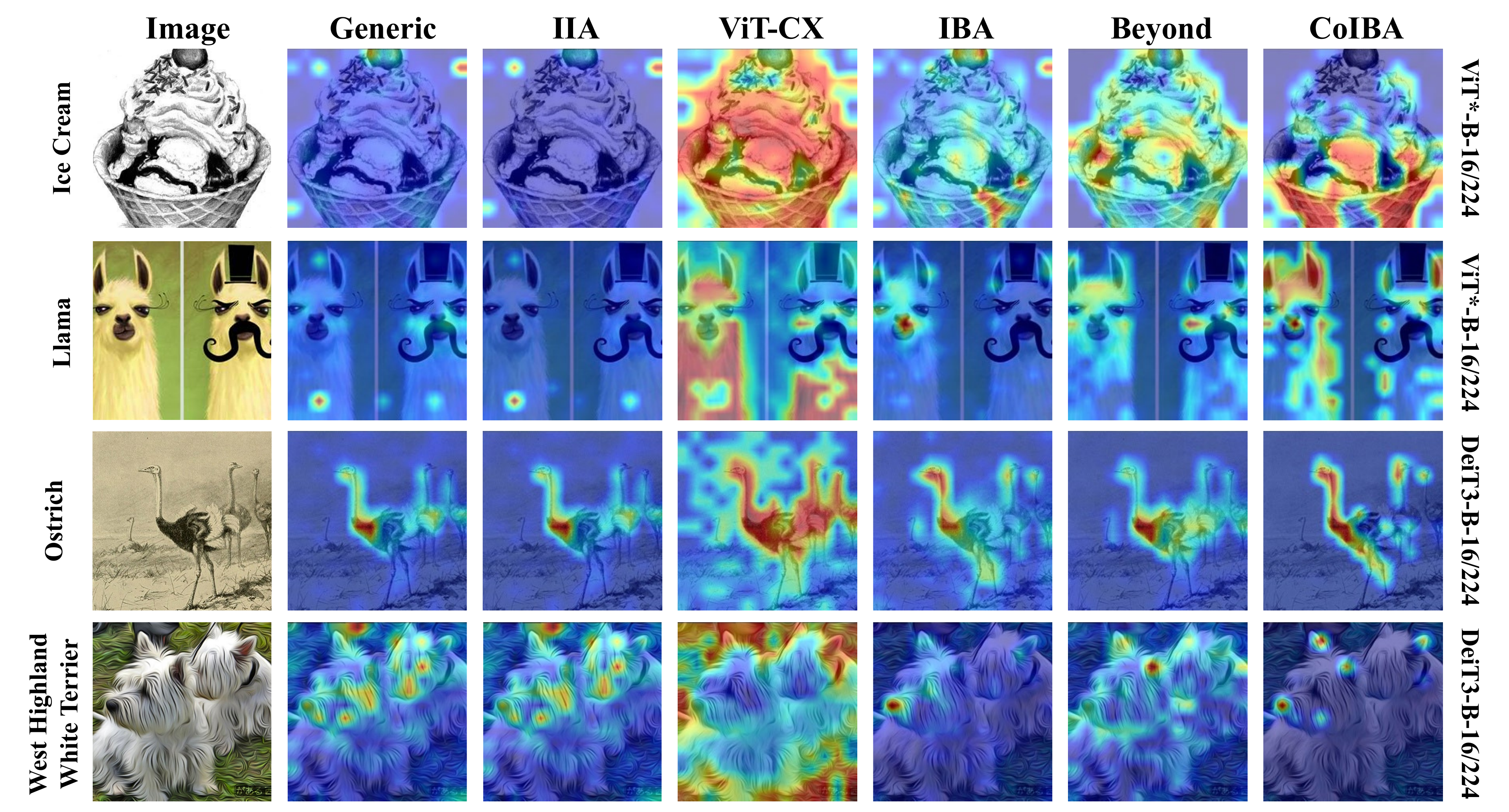}
}

\caption{\textbf{
Visualized attribution maps of IN-R.}
}
% \vspace{-13pt}

\label{fig:vis4}
\end{figure*}

\clearpage

\begin{figure*}[!th]
\centering

\subfloat[IN-A]{
\includegraphics[width=0.85\textwidth]{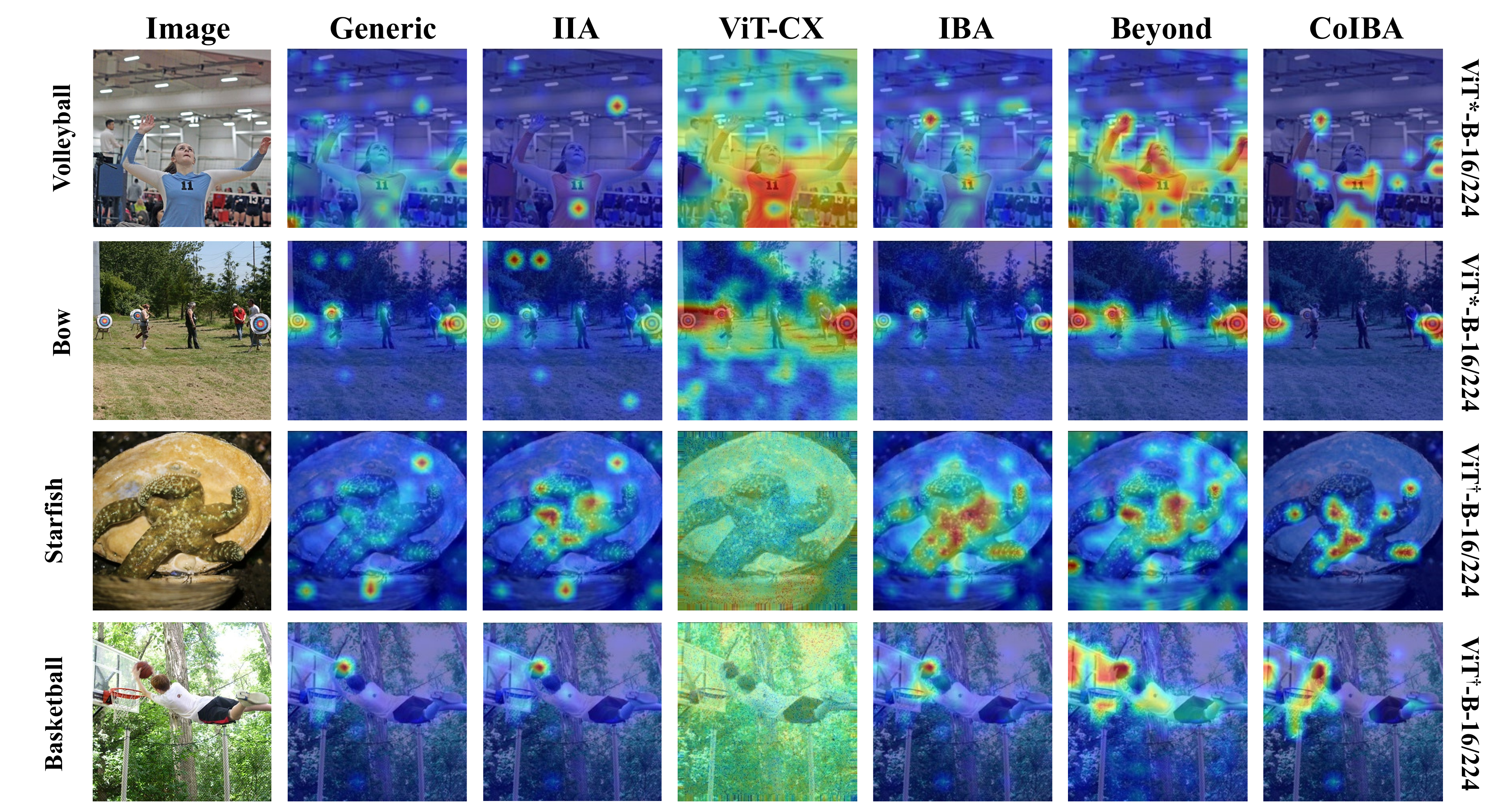}
}
% \includegraphics[width=1.0\textwidth]{./figs/methods-overview.png}
% \includegraphics[width=1.0\textwidth]{./figs/methods.png}
% \vspace{-10pt}
\caption{\textbf{Visualized attribution maps of IN-A.
}
}
% \vspace{-13pt}

\label{fig:vis5}
\end{figure*}

\clearpage

\newpage

% {
%     \small
%     \bibliographystyle{ieeenat_fullname}
%     \bibliography{supp}
% }
\end{document}